\documentclass[final,3p,times,authoryear]{elsarticle}

\usepackage{amssymb}
\usepackage{amsmath}

\usepackage{caption}
\usepackage{subcaption}
\usepackage{multirow}
\usepackage{hyperref}
\usepackage{float}
\usepackage[capitalize]{cleveref}
\usepackage{array,booktabs}
\newcommand {\otoprule}{\midrule [\heavyrulewidth]}
\newcolumntype {+}{ >{\global\let\currentrowstyle\relax}}
\newcolumntype {^}{ >{\currentrowstyle }}
  \newcommand {\rowstyle}[1]{\gdef\currentrowstyle{#1} %
  #1\ignorespaces
  }
\newcommand{\tabhead}{\rowstyle{\bfseries}}
\usepackage[para,online,flushleft]{threeparttable}
\usepackage[T1]{fontenc}
\usepackage{pifont}
\usepackage{tablefootnote}

\usepackage{xcolor}

\newcommand*{\lmss}{\fontfamily{lmss}\selectfont}

\definecolor{darkgray}{rgb}{0.5, 0.5, 0.5}
\definecolor{lightgray}{rgb}{0.8, 0.8, 0.8}
\definecolor{lightgreenbar}{RGB}{199, 233, 180}
\definecolor{darkgreenbar}{RGB}{72, 201, 176}
\definecolor{bluebar}{RGB}{29, 145, 192}

\usepackage[normalem]{ulem}

\usepackage{xspace}

\newcommand{\mgm}{CLaM\xspace}
\newcommand{\mgmvv}{CLaM-VV\xspace}
\newcommand{\mgmfv}{CLaM-FV\xspace}
\newcommand{\cbis}{CBIS\xspace} 
\newcommand{\cbisddsm}{CBIS-DDSM\xspace}

\newcommand{\cbisoff}{CBIS$_{\text{off}}$\xspace}
\newcommand{\cbiscus}{CBIS$_{\text{cus}}$\xspace}
\newcommand{\vindr}{VinDr\xspace}
\newcommand{\mgmannot}{CLaM-Annot\xspace}

\newcommand{\kimrn}{{\lmss DIB-MG}\xspace}

\newcommand{\rnIIIIV}{{\lmss Resnet34}\xspace}

\newcommand{\gmicresnet}{{\lmss GMIC-ResNet18}\xspace}
\newcommand{\gmic}{{\lmss GMIC}\xspace}
\newcommand{\densenet}{{\lmss DenseNet169}\xspace}

\newcommand{\ISMeanImg}{{\lmss IS-Mean$^{\text{img}}$}\xspace} 
\newcommand{\ISMaxImg}{{\lmss IS-Max$^{\text{img}}$}\xspace} 
\newcommand{\ISAttImg}{{\lmss IS-Att$^{\text{img}}$}\xspace}
\newcommand{\ISGattImg}{{\lmss IS-GAtt$^{\text{img}}$}\xspace} 
\newcommand{\ESMeanImg}{{\lmss ES-Mean$^{\text{img}}$}\xspace} 
\newcommand{\ESMaxImg}{{\lmss ES-Max$^{\text{img}}$}\xspace} 
\newcommand{\ESAttImg}{{\lmss ES-Att$^{\text{img}}$}\xspace} 
\newcommand{\ESGattImg}{{\lmss ES-GAtt$^{\text{img}}$}\xspace}
\newcommand{\ISAttSide}{{\lmss IS-Att$^{\text{side}}$}\xspace} 
\newcommand{\ESAttSide}{{\lmss ES-Att$^{\text{side}}$}\xspace}

\newcommand{\SILil}{{\lmss SIL$^{\text{IL}}$}\xspace} 
\newcommand{\SILcl}{{\lmss SIL$^{\text{CL}}$}\xspace}

\begin{document}

\makeatletter
\def\ps@pprintTitle{%
  \let\@oddhead\@empty
  \let\@evenhead\@empty
  \def\@oddfoot{\underreview}
  \let\@evenfoot\@oddfoot
}
\makeatother

\def\underreview{
{Under review\hfill}
\gdef\underreview{}
}

\begin{frontmatter}



\title{Case-level Breast Cancer Prediction for Real Hospital Settings} 


\author[ut,zgt]{Shreyasi Pathak}\corref{cor1}
\ead{shreyasi12dgp13@gmail.com}
\cortext[cor1]{Corresponding author}
\author[marburg,mannheim]{J\"org Schl\"otterer} 
\author[zgt]{Jeroen Geerdink}
\author[ut,zgt]{Jeroen Veltman}
\author[ut]{Maurice van Keulen}
\author[ut]{Nicola Strisciuglio}
\author[marburg]{Christin Seifert}
    \affiliation[ut]{organization={University of Twente},
        addressline={Drienerlolaan 5},
        city={Enschede},
        postcode={7522 NB},
        country={The Netherlands}}

    \affiliation[zgt]{organization={Hospital Group Twente},                 
        addressline={Geerdinksweg 141},
        city={Hengelo},
        postcode={7555 DL},
        country={The Netherlands}}
    \affiliation[marburg]{organization={University of Marburg},                 
        addressline={Biegenstraße 10},
        city={Marburg},
        postcode={35037},
        country={Germany}}
    \affiliation[mannheim]{organization={University of Mannheim},                 
        addressline={Schloss Ehrenhof Ost},
        city={Mannheim},
        postcode={68161},
        country={Germany}}

\begin{abstract}
Breast cancer prediction models for mammography assume that annotations are available for individual images or regions of interest (ROIs), and that there is a fixed number of images per patient. These assumptions do not hold in real hospital settings, where clinicians provide only a final diagnosis for the entire mammography exam (case).  Since data in real hospital settings scales with continuous patient intake, while manual annotation efforts do not, we develop a framework for case-level breast cancer prediction that does not require any manual annotation and can be trained with case labels readily available at the hospital. Specifically, we propose a two-level multi-instance learning (MIL) approach at patch and image level for case-level breast cancer prediction and evaluate it on two public and one private dataset. 
We propose a novel domain-specific MIL pooling observing that breast cancer may or may not occur in both sides, while images of both breasts are taken as a precaution during mammography.
We propose a dynamic training procedure for training our MIL framework on a variable number of images per case. 
We show that our two-level MIL model can be applied in real hospital settings where only case labels, and a variable number of images per case are available, without any loss in performance compared to models trained on image labels. 
Only trained with weak (case-level) labels, it has the capability to point out in which breast side, mammography view and view region the abnormality lies.
\end{abstract}


\end{frontmatter}



\section{Introduction}

\begin{figure}[thbp]
\captionsetup[subfigure]{labelformat=empty}
\centering
    \begin{subfigure}[b]{0.18\textwidth}
        \includegraphics[scale=0.24]{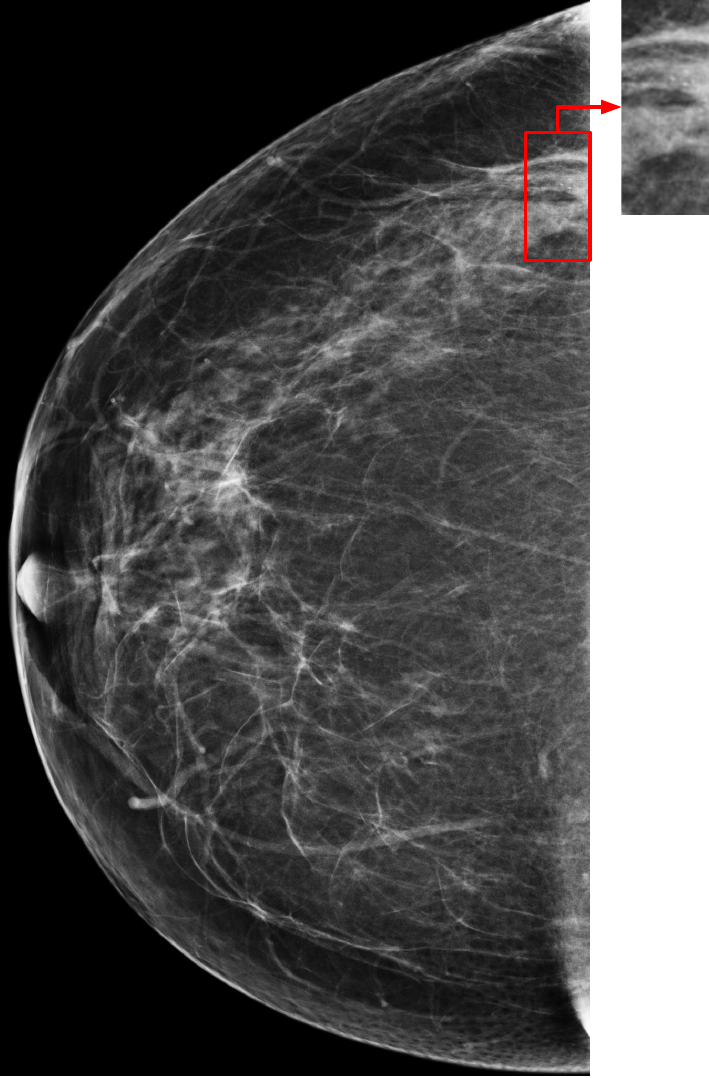}
        \caption{R-CC}
    \end{subfigure}%
    \begin{subfigure}[b]{0.23\textwidth}
        \includegraphics[scale=0.24]{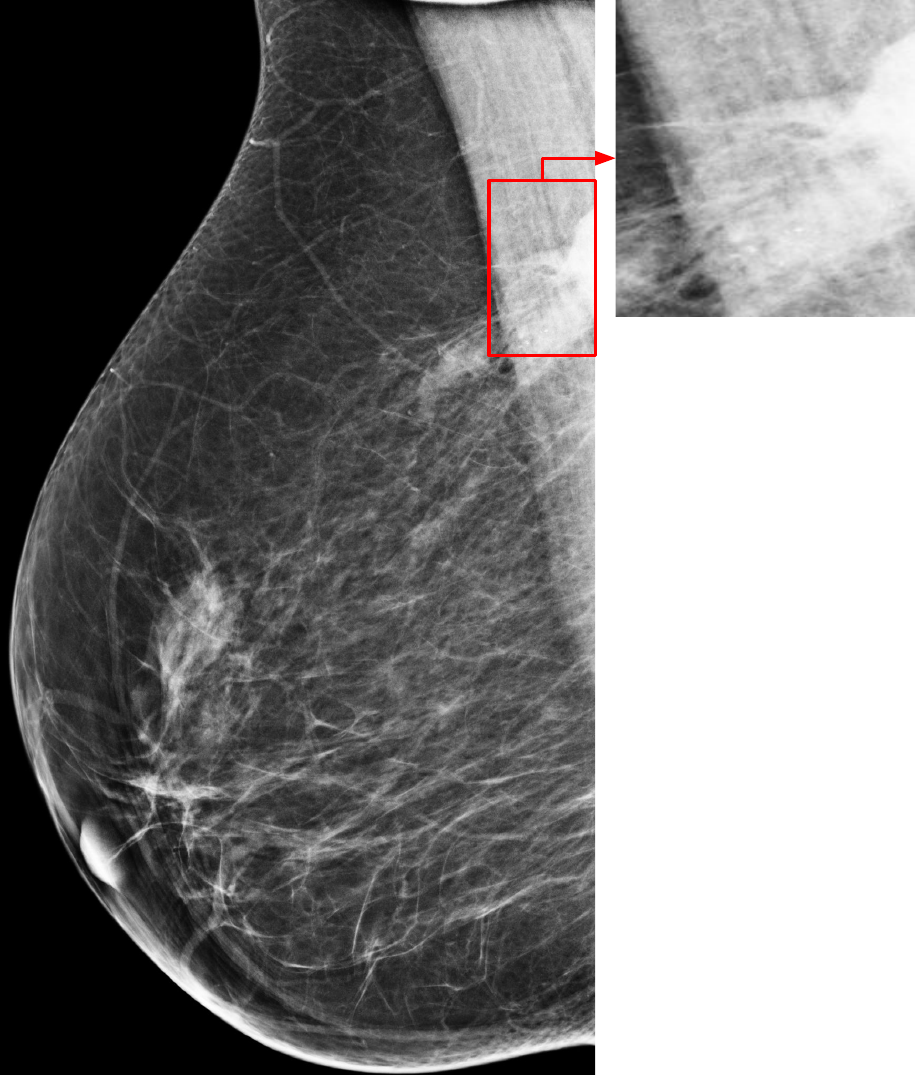}
        \caption{R-MLO}
    \end{subfigure}%
    \begin{subfigure}[b]{0.23\textwidth}
        \includegraphics[scale=0.24]{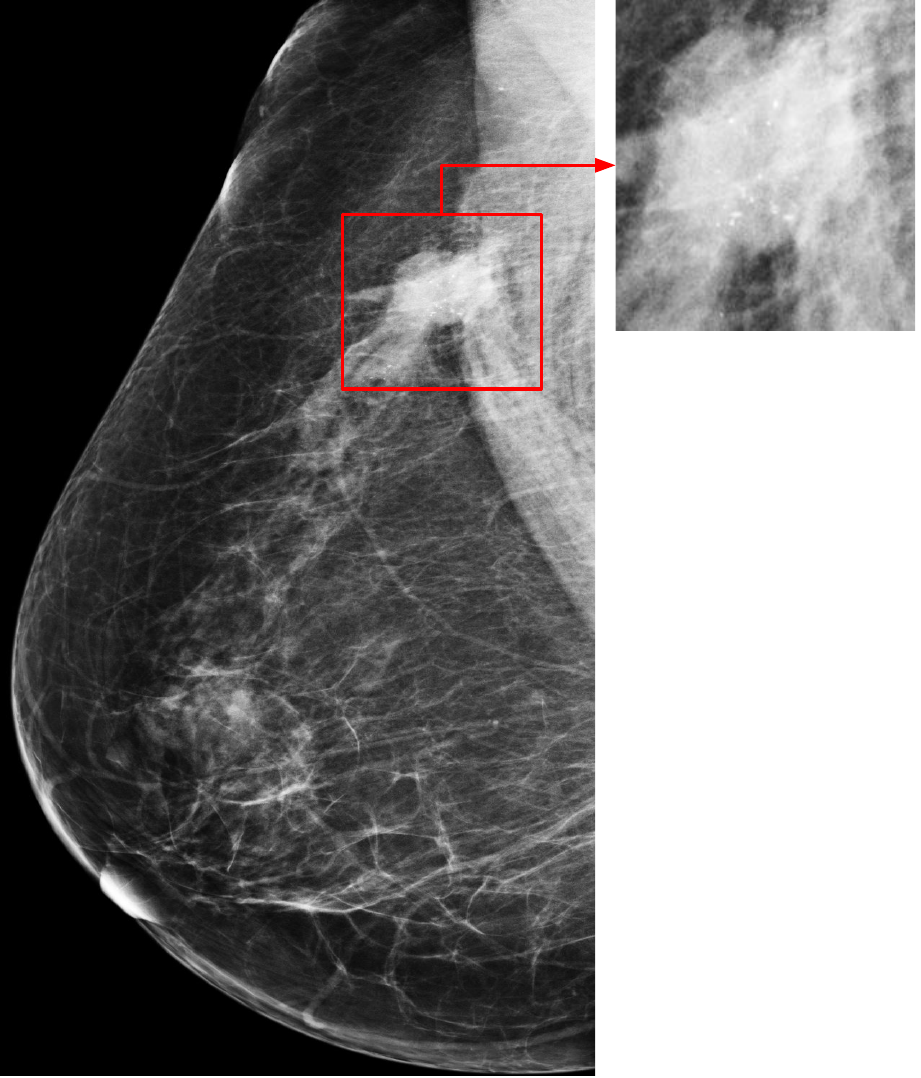}
        \caption{R-XCCL}
    \end{subfigure}%
    \begin{subfigure}[b]{0.18\textwidth}
        \includegraphics[scale=0.24]{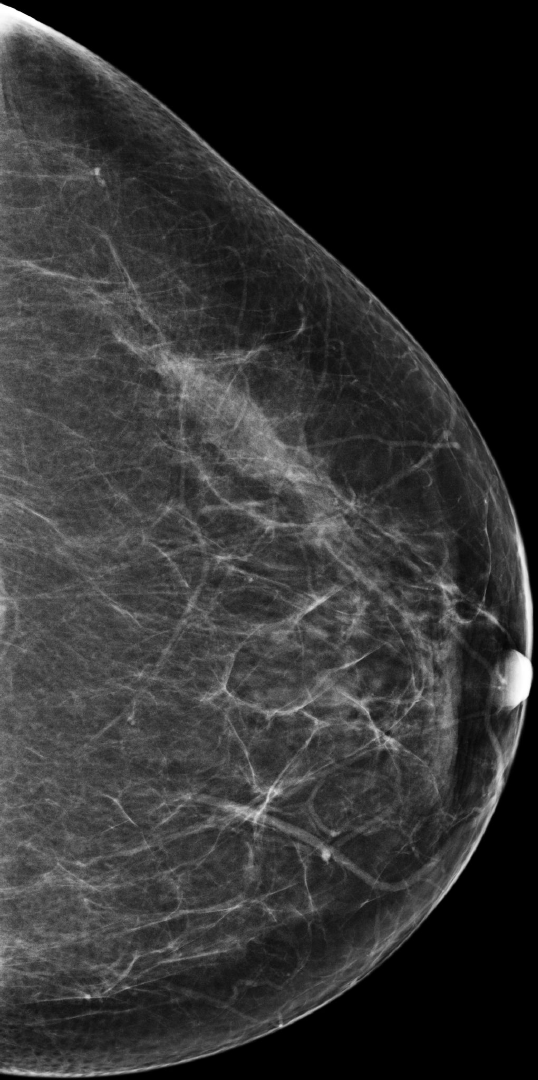}
        \caption{L-CC}
    \end{subfigure}%
    \begin{subfigure}[b]{0.18\textwidth}
        \includegraphics[scale=0.24]{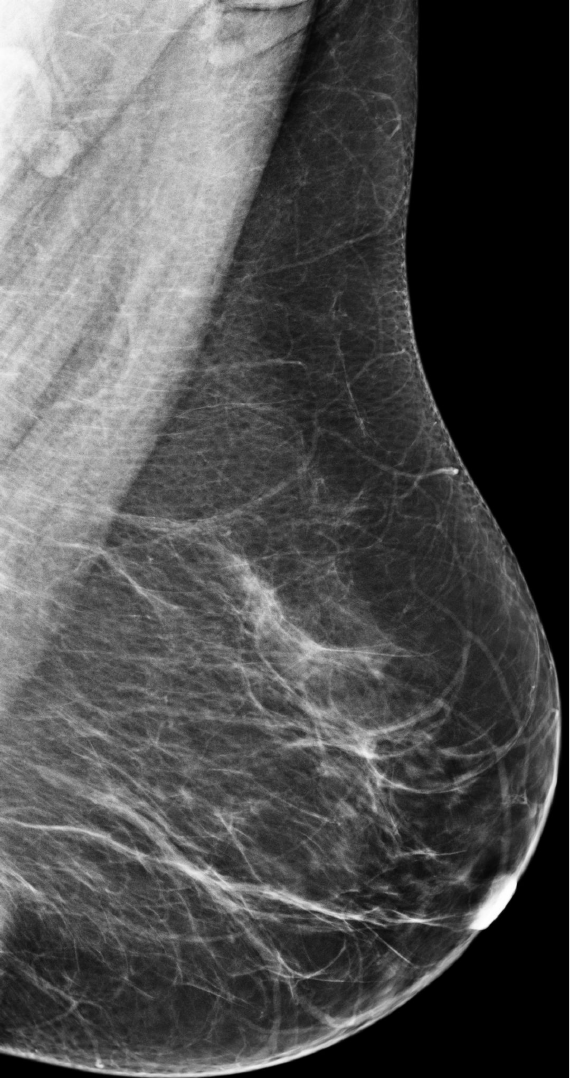}
        \caption{L-MLO}
    \end{subfigure}
\caption{A malignant case from the \mgm dataset showing standard craniocaudal (CC) and mediolateral oblique (MLO) views of the right (R-) and left (L-) breast, along with an additional exaggerated craniocaudal view of the right breast (R-XCCL). A pathologically proven malignant mass of irregular shape and indistinct margin with calcification is visible mainly in the R-MLO and R-XCCL views and no abnormality is visible in the left breast. The malignant mass is best visible in the additional view, R-XCCL, highlighting the importance of including additional views in the input to a predictive model. The case is labeled malignant in the hospital system due to the presence of malignant abnormalities in the right breast. This highlights the challenges \textbf{C1} (not all images contain abnormalities), \textbf{C2} (ROIs are small) and \textbf{C3} (variable number of images per case). 
}
\label{fig:intro:case-5image-mgm-example}
\end{figure}

Breast cancer is among the most common cancer types, with 2.26 million cases reported worldwide in 2020 causing 685k deaths~\citep{Cancer}. Early detection of breast cancer from mammography images has led to a significant decrease in mortality~\citep{tabar2001beyond,tabar2011swedish}. 
Computer-aided diagnostics for mammography images can improve clinical tools in cancer detection and reduce the workload of radiologists~\citep{shen2019deep,kyono2020improving,Wu_2020}. However, existing methods are usually trained using groundtruth labels at the image level~\citep{shu2020deep,shen2021interpretable} or even at the region of interest (ROI) level~\citep{rampun2018breast,shen2019deep}. Such labels are not available in a standard clinical workflow. The only groundtruth that is available is at the case level, where only a few out of several images of one case show signs of abnormality (cf. Fig.~\ref{fig:intro:case-5image-mgm-example}). The acquisition of image or ROI labels is not feasible in general hospitals due to high costs and time demands on clinicians. 

\begin{figure}[!thbp]
\centering
\includegraphics[scale=0.34]{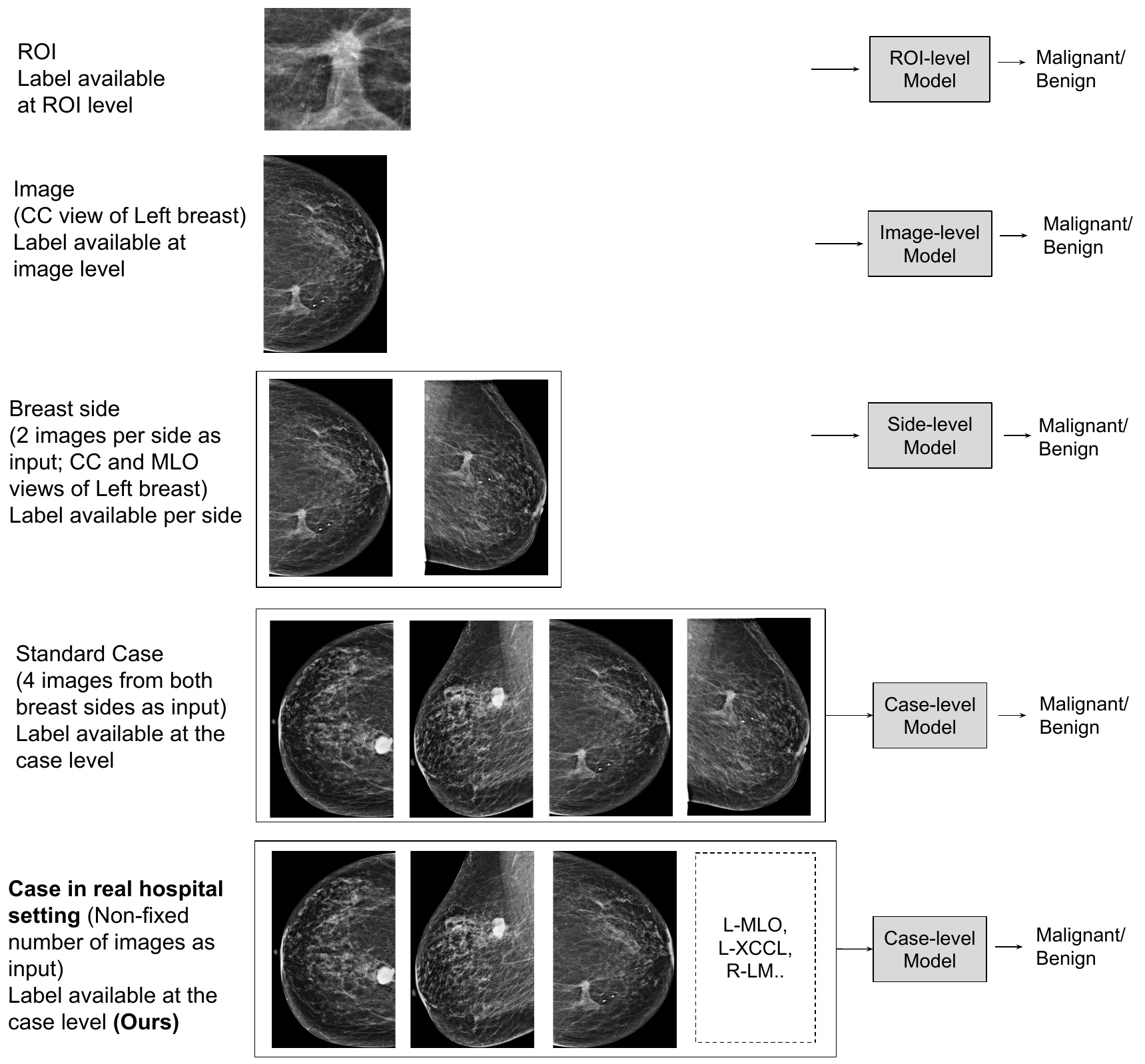}
\caption{Existing breast cancer prediction models differ at the input levels and are trained with labels at different granularities: ROI, image, side and case level (cf. Table~\ref{tab:relatedwork}). Here, we show how real hospital settings are different from the settings of existing work. In real hospital settings, labels are available at the case level and cases can contain a variable number of images (last row) and we propose a model for this setting.}
\label{fig:intro:overview-figure}
\end{figure}

In Fig.~\ref{fig:intro:overview-figure}, we show the settings of existing work on breast cancer prediction from mammograms. Deploying existing methods to real-world clinical workflows presents several challenges that we illustrate by an example in Fig.~\ref{fig:intro:case-5image-mgm-example}. Groundtruth labels are only available for an entire mammography exam, namely at the \emph{case level}, which contains multiple images from both sides of the breast, where abnormalities are not necessarily visible in each image (challenge \textbf{C1}). Thus, existing methods that require image or ROI labels are not suitable for training in real hospital settings as the case-level label can not be transferred to the image level.
Furthermore, regions of interest (ROIs) in mammography images are very small, covering only about 2\% of the image area~\citep{Wang_Zhang_Shu_Lv_Yi_2021}, and their manual annotation is not part of a standard clinical workflow. Case-level labels are thus defined by tiny regions of an image, making it difficult for a model to learn relevant features for effective cancer detection (challenge \textbf{C2}). While a standard mammography exam consists of four images, i.e., the two standard views craniocaudal (CC) and mediolateral oblique (MLO) per side,  a radiologist might decide to take more images (e.g., an exaggerated craniocaudal view, XCCL as in Fig.~\ref{fig:intro:case-5image-mgm-example}) to  reduce diagnostic uncertainty or fewer images when not needed. These cases complicate the training of cancer prediction models, since the number of images per case can vary (challenge \textbf{C3}).  
Training models for breast cancer prediction in a real-world clinical workflow can thus be formulated as a \textit{weakly-supervised learning} task, where only case-level labels are available for training. In summary, we highlight the three challenges (cf. Fig.~\ref{fig:intro:case-5image-mgm-example}) that are not addressed by current methods:

\begin{enumerate}
    \item[\textbf{C1}] Groundtruth labels are only available at the case level, while not all images may contain abnormalities.
    \item [\textbf{C2}] ROIs are not annotated and relevant regions cover only about 2\% of the image.
    \item[\textbf{C3}] A mammography exam can result in a variable number of images per case. 
\end{enumerate}

Since existing approaches do not address these challenges, they cannot be applied in real hospital settings.
Approaches that rely on annotated datasets with ROI~\citep{Levy_Jain_2016,shen2019deep}, image~\citep{zhu2017deep,shu2020deep}, or side labels~\citep{akselrod2019predicting,zhang2020new} are not applicable to clinical workflows where only case labels are available (\textbf{C1, C2}). 
Broadcasting case-level labels to all individual images~\citep{akselrod2019predicting,kyono2020improving} causes label noise, because cancer signs are not necessarily visible in all images (\textbf{C1}). For example, in Fig.~\ref{fig:intro:case-5image-mgm-example}, the malignant case-level label is not valid for all images, but only for R-CC, R-MLO and R-XCCL.
Similarly, models that average image predictions for case-level prediction~\citep{kim2018applying} do not account for cancer regions that are only visible in single images (\textbf{C1}) and result in poor performance.
Concatenation of image feature representations to create a case-level representation~\citep{carneiro2015unregistered,Wu_2020} or fixing the number of images in a case~\citep{kim2018applying,Wu_2020} can not handle a variable number of images per case (\textbf{C3}). Weakly-supervised learning at the image level to extract ROIs in an unsupervised manner~\citep{liu2021weakly,shen2021interpretable} addresses \textbf{C2}, but requires that image-level labels  are available (\textbf{C1}). Multiscale feature extraction methods improve ROI detection, but existing work used ROI annotations~\citep{rangarajan2022ultra} or breast side labels~\citep{zhang2020new} to train these models. Both are not available in the standard clinical workflow (\textbf{C1}).~\citet{quellec2016multiple} calculated hand-crafted features of image regions and applied standard MIL approaches~\citep{dietterich1997solving,maron1997framework,andrews2002support,zhang2005multiple} at the case level. However, their hand-crafted features require extensive hyper-parameter tuning, even for  ROI extraction and it is not end-to-end trainable.

In this paper, we formulate breast cancer prediction in real-world hospital settings as a multi-instance learning (MIL) problem~\citep{dietterich1997solving}. We propose a method for case-level breast cancer prediction that uses MIL at two levels: the image level, to account for the fact that not all images contain abnormalities, and the patch level, to account for the fact that not all image-patches contain ROI (cf. Fig.~\ref{fig:model-architecture} for an overview). 
Our end-to-end learning approach is suitable for a variety of real-world breast cancer prediction settings on mammography cases. 
Specifically, our method learns using only case-level labels, thus addressing the challenges of abnormality signs being present in only a subset of images per case \textbf{(C1)} in small regions \textbf{(C2)}, and handling a variable number of images per case \textbf{(C3)}.  
Further, we note that reproducibility of state-of-the-art is an issue in the field of breast cancer prediction, in particular due to unavailability of \emph{full} source code, including all training, (pre-)processing and evaluation settings.

Our contributions are:

\begin{enumerate}
    \item We develop a framework for case-level breast cancer prediction, with MIL at two levels -- patch level and image level and evaluate it on 2 public and 1 private dataset. Our case-level model outperforms the state-of-the-art case-level models and models trained with image-level labels, suggesting that manual annotation of individual images is not needed. Further, the two-level MIL outperforms single-level MIL, i.e., only image-level MIL.
    \item We introduce a novel domain-specific, image-level, side-wise MIL pooling that outperforms existing MIL aggregation methods in classification performance, relevant image identification and region of interest extraction.
    \item Our framework includes a dynamic training procedure to incorporate variable numbers of images per case (7\% in our private dataset are not 4-image cases), resulting in generally higher performance for minority groups than default training.  
    \item We release 
    a Python library\footnote{Open source implementation available at \label{fn:repository}\url{https://github.com/ShreyasiPathak/multiinstance-learning-mammography}} to promote reproducibility and unified comparison in the field. The library includes the source code for training two image-level and two case-level state-of-the-art models along with our own model variants, and all pre-processing and evaluation scripts. 
\end{enumerate}

\section{Analysis of State-of-the-Art}
In this section we review multi-instance learning in medical imaging and state-of-the-art (SoTA) models on breast cancer prediction. Further, we conduct experiments to identify the most promising feature extraction method for the mammogram images and evaluate the applicability of SoTA models in real hospital settings.

\subsection{Multi-instance Learning and its Application in Medical Imaging}
Multi-instance learning (MIL)~\citep{dietterich1997solving} is a type of weakly-supervised learning task, where the label of a bag of instances (case) is known and the labels for individual instances (images) are unknown. MIL tries to learn the relationship between the bag and the instances through various ways as described below.

Image information can be aggregated on score \textit{(instance-space, IS)}~\citep{maron1997framework} or feature \textit{(embedded-space, ES)}~\citep{chen2006miles,wei2016scalable} level with different aggregation strategies (MIL pooling), e.g. mean and max pooling~\citep{Feng_Zhou_2017}, Log-Sum-Exp~\citep{ramon2000multi,pinheiro2015image} and noisy-or~\citep{zhang2005multiple}. 
In \textit{IS MIL}, the model outputs class scores for each instance, which are aggregated by MIL pooling to output a bag-level score. Different MIL pooling functions correspond to different assumptions in IS~\citep{Foulds_Frank_2010,amores2013multiple}. Max pooling assumes that every positive bag has at least one positive instance, which is selected for the bag-level decision (standard multi-instance assumption). Mean pooling averages logits and assumes that every instance contributes equally to the bag-level decision (collective assumption), while attention pooling is used to weigh output scores if contribution of instances differ and have to be learned (weighted collective assumption). In \textit{ES MIL}, the information is aggregated at the feature embedding level, and mapped to a bag-level feature vector using similar aggregation strategies, e.g. attention~\citep{ilse2018attention} and self-attention followed by attention~\citep{rymarczyk2021kernel}.

MIL is frequently used in medical image classification, e.g., in whole slide images (WSI)~\citep{li2021dual,lu2021data,javed2022additive} by converting high-resolution slide images into a bag of patches, 3D medical images like brain MRI~\citep{li2023multi} and chest computed tomography (CT)~\citep{he2021synergistic} into a bag of 2D image slices. For WSI,~\citet{li2021dual} proposed a MIL block with a self-attention-like operation between the image patch with the highest instance score and all image patches in the slide. In 3D images,~\citet{he2021synergistic} use MIL over image patches to bring relevant patches closer and irrelevant patches further away and~\citet{li2023multi} select the 2D MRI slice with the highest score and its spatially nearby slices and calculates the average of these scores to generate the bag-level score. 
WSI may contain multiple abnormal tissue regions/patches ($\approx$20\%~\citep{li2021dual}) in a single image which is different from mammography, where a single mammogram image may contain only one abnormality. Similarly, 3D MRI images have multiple 2D image slices where the ROIs may be visible, however, mammography cases don't have as many views for a single breast side and either one of the views or either one of the breast sides might show ROIs. Therefore, we propose a two-level MIL pooling - a patch-level MIL pooling and a novel image-level MIL pooling that can weigh the views per breast side and each breast side separately.

\subsection{Comparison of SoTA Models for Breast Cancer Prediction}
Table~\ref{tab:relatedwork} summarizes SoTA methods for breast cancer prediction using mammography. 
To highlight differences across existing methods and our work, we compare the methods w.r.t the level at which the training and prediction is done, e.g., ROI, image, breast side or case level, the number of views accepted as input, and whether ROI annotations are used during training. We also report the dataset used, the number of instances, the abnormalities included in the dataset (e.g., mass, calcification, or all), the classification task, and the AUC score. We make three key observations: i) reported scores across methods cannot be compared, ii) not all papers in this domain release source code making reproducibility an issue, and iii) none of the methods addresses all the challenges (\textbf{C1-C3}) of breast cancer prediction in real hospital settings. These key observations will be explained in the following paragraphs.

We found that the reported scores cannot be compared, because datasets differ and not all papers use a publicly available dataset such as for example CBIS-DDSM (CBIS)~\citep{cbisddsm}, DDSM~\citep{heath1998current}, VinDr~\citep{Nguyenvindr} or MIAS~\citep{suckling1994mammographic}.   
Even methods that report results on the same public datasets (e.g., CBIS) differ in i) the data (e.g., amount and selection of a subgroup), ii) the reporting standards of the result, and iii) the input level for training and prediction. Methods \textit{differ in data} due to the addition of instances over time~\citep{shu2020deep,wei2022beyond}, restriction to a specific abnormality~\citep{rampun2018breast,Tsochatzidis2019}, increase in the training dataset by data augmentation~\citep{Levy_Jain_2016,ragab2021framework}, and use of custom dataset split~\citep{shen2019deep}.~\citet{wei2022beyond} reported a difference of 13\% in AUC for~\citet{shen2019deep} on official split (AUC 0.75) vs custom split (AUC 0.88) of \cbis.
\begin{table}[tbhp] 
\centering
\caption{Comparison of our work to existing state-of-the-art in breast cancer prediction with mammography. \textbf{Instances} are the number of \textbf{R}OIs/ \textbf{I}mages/ \textbf{C}ases/ \textbf{Br}east in the dataset, \textbf{Task} is the classification problem - malignant (M), benign (B), normal (N), abnormal (A = M+B) and rest means non-malignant. \textbf{Abnormality} are the types of abnormalities included. \textbf{Input} are the number of instances the model takes as input, e.g. 4 indicates L-CC, L-MLO, R-CC, R-MLO. \textbf{ROI} annotation used in training (\checkmark) vs not (\ding{55}). \textbf{Level} indicates the training label and prediction level. $^\ast$AUC on \cbis taken from~\citet{shu2020deep}, $^\dagger$uses hand-crafted features, $^\ddagger$our reproduction. Methods differ in \textcolor{lightgray}{data}, \textcolor{darkgray}{reporting standard, label}.}
\label{tab:relatedwork}
\resizebox{\textwidth}{!}{%
\begin{tabular}{+l^c^r^c^c^c^c^c^c}
\toprule \tabhead
\textbf{Paper} & \textbf{Dataset} & \textbf{Instances} & \textbf{Task} & \textbf{Abnormality} & \textbf{Input} & \textbf{ROI} & \textbf{Level} & \textbf{AUC}\\ \otoprule
\citet{rampun2018breast} & CBIS & 1,593 (R) & M/B & mass & 1 & \checkmark & ROI & \textcolor{lightgray}{0.84}\\
\citet{Tsochatzidis2019} & CBIS & 1,697 (R) & M/B & mass & 1 & \checkmark & ROI & \textcolor{lightgray}{0.80} \\
\citet{ragab2021framework} & CBIS & 5,272 (R) & M/B & mass & 1 & \checkmark & ROI & \textcolor{lightgray}{1.00} \\
\citet{khan2019multi} & MIAS-CBIS & 3,890 (R) & M/B & mass, calc & 4 & \checkmark & ROI-Case & \textcolor{lightgray}{0.77}\\
\citet{zhu2017deep} & CBIS & 1,644 (C) & M/B & mass, calc. & 1 & \ding{55} & Image & \textcolor{darkgray}{0.79}$^\ast$ \\
\citet{shen2019deep} & CBIS & 2,478 (I) & M/B & mass, calc. & 1 & \checkmark & Image & \textcolor{lightgray}{0.88} \\
\citet{shu2020deep} & CBIS & 3,071 (I) & M/B & mass, calc. & 1 & \ding{55} & Image & \textcolor{lightgray}{0.84}\\
\citet{shen2021interpretable} & CBIS & 1,644 (C) & M/B & mass, calc. & 1 & \ding{55} & Image & \textcolor{darkgray}{0.86} \\
\citet{wei2022beyond} & CBIS & 3,103 (I) & M/B & mass, calc. & 1 & \ding{55} & Image & \textcolor{darkgray}{0.83} \\
\citet{carneiro2015unregistered} & DDSM & 680 (I) & M/B & mass, calc. & 2 & \checkmark & Breast side & \textcolor{lightgray}{0.97}\\
\citet{akselrod2019predicting} & Private & 52,936 (I) & M/rest & all & 4 & \ding{55} & Breast side & \textcolor{lightgray}{0.91} \\
\citet{Wu_2020} & Private & 229,426 (C) & M/B/N & all & 4 & \checkmark & Breast side & \textcolor{lightgray}{0.89} \\
\citet{van2021multi} & \cbis & 708 (Br) & M/B & mass & 2 & \ding{55} & Breast side & \textcolor{lightgray}{0.80} \\
\citet{petrini2022breast} & CBIS & 2,694 (I) & M/B & mass, calc. & 2 & \checkmark & Breast side & \textcolor{lightgray}{0.85}\\
\citet{manigrasso2024mammography} & CSAW & 8,723 (C) & M/R & all & 4 & \checkmark & Breast side & \textcolor{lightgray}{0.88} \\
\citet{quellec2016multiple} & DDSM & 2,479 (C) & A/N & mass, calc.  & 4 & \ding{55} & Case & \textcolor{lightgray}{0.80}$^\dagger$\\
\citet{kim2018applying} & Private & 29,107 (C) & M/N & all & 4 & \ding{55} & Case & \textcolor{lightgray}{0.91} \\
 & CBIS & 1,645 (I) & M/B & mass, calc. & Any & \ding{55} & Case & 0.64$^\ddagger$ \\ 
\citet{mckinney2020international} & OPTIMAM & 102,640 (C) & M/rest & all & 4 & \checkmark & Case & \textcolor{lightgray}{0.89} \\ [3pt]
\textbf{This work}  & \cbisoff & 1,645 (C) & M/B & mass, calc. & Any & \ding{55} & Case & 0.77 \\
 &  VinDr & 5,000 (C) & M/rest & all & Any & \ding{55} & Case & 0.83 \\
 & Private & 21,013 (C) & M/rest & all & Any & \ding{55} & Case & 0.85 \\ \bottomrule
\end{tabular}
}
\end{table} 
The \textit{reporting standards of results differ} due to use of test time augmentation to report final scores (improving the performance by 1-2\%~\citep{wei2022beyond}) and use of model ensembles~\citep{shen2021interpretable,wei2022beyond}. Further, methods \textit{differ in training label and prediction level}, e.g., training with ROI labels for prediction at the ROI level~\citep{Levy_Jain_2016,rampun2018breast,Tsochatzidis2019}, prediction at image level~\citep{ribli2018detecting,agarwal2019automatic,shen2019deep}, prediction at side level~\citep{Wu_2020,petrini2022breast}, and case level~\citep{mckinney2020international,manigrasso2024mammography}; training with image labels for image-level predictions~\citep{zhu2017deep,shu2020deep,hu2021multi,shen2021interpretable}; training with breast side label for side-level predictions~\citep{carneiro2015unregistered,akselrod2019predicting,Wu_2020} and case-level predictions~\citep{manigrasso2024mammography}. 
Some methods also use fine-grained labels that are hard to obtain in practice to train classifiers, e.g., abnormality type~\citep{shen2019deep,kyono2020improving}. 
Thus, performance scores are not comparable across methods, which makes it impossible to select the best-performing method from scores alone.
We color-coded the AUC scores to indicate how comparable the scores from existing models are to our result on CBIS: \textcolor{lightgray}{lightgray} indicates differences in the dataset or the train-test split, \textcolor{darkgray}{darkgray} indicates results reported on the official CBIS split, but differences in reporting standards or training label used, black indicates fully comparable results on the official CBIS split. 

Reproducibility is an issue in this domain. 
Most researchers do not make their source code publicly available and some only provide inference code, which makes the reproduction of results difficult and impedes rigorous benchmark comparison.
We shortlisted two unsupervised ROI extraction methods~\citep{shu2020deep,shen2021interpretable} as image-level feature extractor for our MIL model. We also shortlisted the feature extractor of~\citet{kim2018applying} as it is the strong case-level baseline. We implemented and compared them to select the best feature extractor for our case-level MIL model. Finally, we release a library containing our source code for 4 SoTA models and our work to promote reproducibility. 

Among the models trained with case labels in Table~\ref{tab:relatedwork},~\citet{quellec2016multiple} use hand-crafted features,~\citet{kim2018applying} average the image scores to get case-level score and~\citet{mckinney2020international} use ROI annotations. To the best of our knowledge, there is no case-level breast cancer prediction model that addresses all the challenges of real hospital settings. 
We address this gap with our end-to-end breast cancer prediction model that, while trained with only case-level labels, has the capability to learn the importance of images, extract ROIs, and is applicable to cases with variable numbers of images.

\subsection{Reproducibility of SoTA Models} 
\label{ssec:results:feature-extractors}
Since source code is not available for any of the selected candidates (except \gmic,~\citet{shen2021interpretable}, which has only inference code available), we reproduced related work from the information given in the corresponding paper and by contacting the authors. We publish our implementation of these methods along with the training details.\textsuperscript{\ref{fn:repository}}

We selected the image-level feature extractor for our case-level model by reproducing and comparing multiple related methods. We compared models trained from scratch against pretrained models. 
Specifically, we compared \kimrn~\citep{kim2018applying}, an adapted single channel {\lmss ResNet} model trained from scratch, a fine-tuned pretrained \densenet with global average pooling {\lmss (avgpool)} or max pooling {\lmss (maxpool)} in the last layer before the classification head~\citep{shu2020deep} and a pretrained \rnIIIIV with {\lmss avgpool}~\citep{shen2021interpretable}. 
We also compared \rnIIIIV and \densenet with models capable of unsupervised ROI extraction, \gmicresnet~\citep{shen2021interpretable} or models capable of feature selection, i.e. a pretrained \densenet with region-based group max pooling {\lmss (RGP)} and global group-max pooling {\lmss (GGP)}~\citep{shu2020deep}.
We report performance (cf. Table~\ref{tab:results:feature-extractor}) as stated in the original papers and from our reproduction\footnote{For reproduction, we used the same bit resolution (8 bit) for all models, except for models from~\citet{shen2021interpretable} where authors explicitly mention 16 bit resolution.} on the official split of \cbis for image-level prediction, i.e., a model takes as input a mammogram image and is trained with the image label. 
\begin{table}[th!bp]
\centering
\caption{Comparing feature extractors for image-level prediction on official \cbis split, \cbisoff. Batch size set to 10. \gmic outperforms other models in terms of F1 score and our reproduced performance is lower than the reported performance.}
\label{tab:results:feature-extractor}
\begin{small}
\begin{tabular}{+l^c^c^c}
\toprule \tabhead
Model   & F1$^{\textit{Our}}$ & AUC$^{\textit{Our}}$ & AUC$^{\textit{Paper}}$  \\\otoprule
{\lmss DIB-MG}~\citep{kim2018applying}  & $0.54 \pm 0.02$ & $0.64 \pm 0.00$  & n.a.  \\ [2pt]
\densenet~\citep{shu2020deep} & & & \\
+ {\lmss avgpool}  & $0.62 \pm 0.03$  & $0.76 \pm 0.01$ & $0.76 \pm 0.00$  \\
+ {\lmss maxpool}  & $0.63 \pm 0.01$ & $0.74 \pm 0.00$ & $0.74 \pm 0.00 $   \\
+ {\lmss RGP} (k=0.7) & $0.62 \pm 0.01$  & $0.76 \pm 0.01$ & ${0.84} \pm 0.00 $ \\
+ {\lmss GGP} (k=0.7)  & $0.62 \pm 0.03$ & $0.76 \pm 0.02 $ & $0.82 \pm 0.00$  \\ [2pt]
\rnIIIIV~\citep{shen2021interpretable} & $\textbf{0.66} \pm 0.02$ & $0.78 \pm 0.01$ & $0.79 \pm 0.01$   \\
\gmicresnet~\citep{shen2021interpretable} & $\textbf{0.66} \pm 0.02$  & $\textbf{0.79} \pm 0.02$  & $0.83 \pm 0.00$ \\
\bottomrule
\end{tabular}
\end{small}
\end{table}

Similar to previous work~\citep{raff2019step}, we observe that results reported by the original authors differ from our reproduction, i.e., we observe lower performance values.\footnote{We attribute the performance gap for GMIC (AUC = 0.79, our reproduction vs 0.83~\citep{shen2021interpretable}) to a smaller hyperparameter budget. 0.83 seem reasonable to achieve with a larger budget, as we observed some individual runs can reach an AUC of 0.82.} 
This calls for higher rigor in reporting all relevant details, ideally accompanied by source code, such that originally reported performance can be reproduced (or at least similar performance can be reached).
We confirm previous findings that models with pretrained computer vision backbones (\densenet, \rnIIIIV) outperform backbones trained from scratch (\kimrn)~\citep{carneiro2015unregistered,Levy_Jain_2016,Tsochatzidis2019}.
The unsupervised ROI extraction model, \gmicresnet, is on par with a standard \rnIIIIV, and outperforms the other feature selection methods, {\lmss RGP} and {\lmss GGP} by $0.04$ points in F1 score.
For our further experiments, we selected {\gmicresnet} as the feature extractor for our case-level models due to its high performance and capability of unsupervised ROI extraction.

\begin{table}[th!bp]
\centering
\caption{Why do we need case-level training? Evaluation of existing models at the case level on the private hospital dataset \mgmvv shows that existing models cannot achieve as high performance as models tailored to the requirements of real hospital settings.}
\label{tab:results:crossdata-exp}
\begin{small}
\begin{tabular}{+l^l^l^c^c^c}
\toprule \tabhead
Model & Setting & Train Dataset & Train Level & F1 & AUC \\\otoprule
\gmic~\citep{shen2021interpretable} & cross-dataset & \cbiscus & Image &  $0.31 \pm 0.00$ & n.a. \\
\gmic~\citep{shen2021interpretable} & cross-dataset & \vindr & Image & $0.33 \pm 0.01$ & n.a. \\
\gmic~\citep{shen2021interpretable} & same-dataset &\mgmvv & Image & $0.44 \pm 0.01$ & n.a. \\ [0.1cm]
{\lmss DIB-MG}~\citep{kim2018applying} & cross-dataset & \cbiscus & Case  & $0.29 \pm 0.03$ & $0.52 \pm 0.03$\\
{\lmss DIB-MG}~\citep{kim2018applying} & cross-dataset & \vindr & Case & $0.31 \pm 0.00$ & $0.53 \pm 0.00$\\
{\lmss DIB-MG}~\citep{kim2018applying} & same-dataset & \mgmfv & Case & $0.33 \pm 0.00$  & $0.67 \pm 0.01$ \\ [0.1cm]
{\lmss DMV-CNN}~\citep{Wu_2020} & cross-dataset & \vindr & Case  & $0.30 \pm 0.00$ & $0.57 \pm 0.01$\\
{\lmss DMV-CNN}~\citep{Wu_2020} & same-dataset & \mgmfv & Case & $0.36 \pm 0.00$ & $0.68 \pm 0.01$\\ [0.1cm] 
\midrule
\ESAttSide (our) & cross-dataset & \cbiscus & Case  & $0.31 \pm 0.00$ & $0.54 \pm 0.03$ \\
\ESAttSide (our) & cross-dataset & \vindr & Case  & $0.37 \pm 0.03$ & $0.69 \pm 0.01$ \\
\ESAttSide (our) & same-dataset & \mgmvv & Case   & $\textbf{0.61} \pm 0.01$ & $\textbf{0.85} \pm 0.00$ \\
\bottomrule
\end{tabular}
\end{small}
\end{table}

\subsection{Applicability of SoTA in Real Hospital Settings}
We evaluated the applicability and performance of the SoTA models, GMIC~\cite{shen2021interpretable}, DIB-MG~\citep{kim2018applying} and DMV-CNN~\citep{Wu_2020}, on real hospital settings using our private mammography dataset, \mgmvv (cf. Table~\ref{tab:dataset:mammograms}) extracted from a real hospital setting. We compared two settings: i) training the SoTA models on 2 public datasets, CBIS and VinDr, and evaluating on \mgm (cross-dataset), and ii) training and evaluating on the same dataset, \mgm (same-dataset). As only case labels are available for \mgm, the image-level model GMIC (third row in Table~\ref{tab:results:crossdata-exp}) was trained on \mgmvv by transferring the case label to each image in the case. 
During inference, the malignant label was assigned to a case if any image in that case was predicted as malignant.
Note that the AUC score cannot be calculated in this setup (n.a. in the first three rows in Table~\ref{tab:results:crossdata-exp}) as only a label is assigned to the case and no probability. We show the performance of our proposed model, \ESAttSide in Table~\ref{tab:results:crossdata-exp} to strengthen the point of why case-level models tailored to the requirements of real hospital settings are important.

Results reported in Table~\ref{tab:results:crossdata-exp} show that i) the same-dataset setting achieves higher performance than the cross-dataset setting, and that ii) the case-level model tailored to real hospital settings achieves higher F1 than image-level models (third vs last row in Table~\ref{tab:results:crossdata-exp}). 
These results demonstrate that we cannot use existing SoTA breast cancer models (neither when trained on a different dataset nor on the same dataset) for achieving high performance on real private hospital datasets. This finding further highlights the motivation behind developing case-level models that can be trained in real hospital settings. As we can see, a
case-level model tailored for the data that is available in real hospital settings (e.g., our proposed model, \ESAttSide) is needed to achieve good performance on such data.  

\section{Approach}
\label{sec:approach}

In this section, we describe our case-level breast cancer prediction model for real hospital settings.

\begin{figure*}[thbp]
\centering
\includegraphics[scale=0.78]{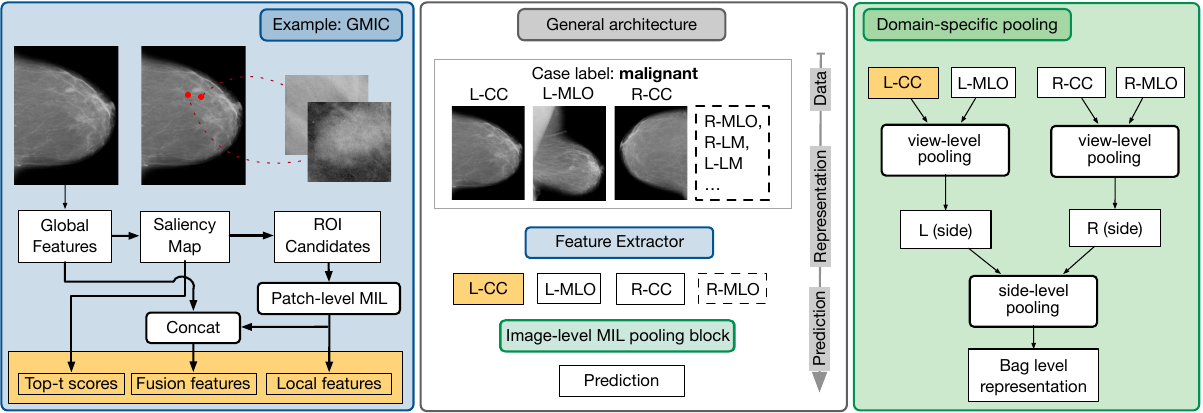}
\caption{Model architecture. \emph{Center:} overall architecture. An end-to-end trained feature extractor is applied to a variable number of input images (e.g. L-CC, L-MLO, R-CC), returning a feature representation per image. Image-level multi-instance learning (MIL) methods then aggregate these feature for a final decision.  \emph{Left:} An example feature extractor module, GMIC, capable of unsupervised ROI extraction. \emph{Right:} Our proposed domain-specific pooling block performing view-level pooling and then performing side-level pooling (based on the assumption that single images of one side highly correlate in the presence of abnormality). Dashed lines indicate absent views. \textit{Best viewed in color}.  }
\label{fig:model-architecture}
\end{figure*}

\subsection{Problem Definition}
\label{ssec:approach:problem-def}
Our task is to predict breast cancer (malignant or benign) in real hospital settings, i.e., the input is a mammography case containing a non-fixed number of images from breast sides (left, right). The groundtruth of malignant or benign is available at the case level and the ROIs in the breast are not annotated. We aim to develop a model that can predict the diagnosis for a mammography case and find the important images and ROIs in the case. 

Let $\mathcal{X} \times \mathcal{Y} = \{(X_1, y_1), \ldots, (X_n, y_n), \ldots, (X_N,y_N)\}$ be the training set with N mammography cases where $X_n$ is the n\textsuperscript{th} mammography case and $y_n \in \{0, 1\}$ ($0$: benign, $1$: malignant) is the groundtruth of the case. 
A mammography case $X_n$ contains a set of M images taken for a patient during a mammography exam from breast side $s \in S=$ \{Left (L), Right (R)\} with view $v \in V=$ \{craniocaudal (CC), mediolateral oblique (MLO), lateromedial (LM), mediolateral (ML), exaggerated craniocaudal (XCCL)\}, i.e., $X_n = \{I_{n,1}, \ldots, I_{n,m}, \ldots, I_{n,M}\}$ where $I_{n,m}$ is the m\textsuperscript{th} image of side $s$ and view $v$, e.g. L-CC, from the n\textsuperscript{th} mammography case, $X_n$, and the label of each image is unknown. 
Note that CC and MLO are the standard views taken during a mammography exam, with other additional views, LM, ML and XCCL taken to rule out or confirm abnormalities.
Each image, $I_{n,m}$ is composed of J patches, i.e. $I_{n,m} = \{p_{n,m,1}, \ldots, p_{n,m,j}, \ldots, p_{n,m,J}\}$ where $p_{n,m,j}$ is the j\textsuperscript{th} patch in the m\textsuperscript{th} image from the n\textsuperscript{th} case and the labels of these patches are unknown. Each image in a case may or may not contain abnormal patches (ROIs). If a case contains an abnormality, usually only one of these patches from some images in a case may show the abnormality. For the sake of simplicity of notations, we denote n\textsuperscript{th} case as $X_n$, m\textsuperscript{th} image in the n\textsuperscript{th} case as $I_m$ and j\textsuperscript{th} patch in the  m\textsuperscript{th} image as $p_j$ in the rest of this section. Any notation henceforth with $n$, $m$, and $j$ subscript denotes case, image and patch-level values.

We address the problem through multi-instance learning (MIL). Unlike the \textit{single-instance learning (SIL)} setting where the model takes a single instance, i.e., one image $I_m$, or one patch $p_j$, with a known label as input, in the \textit{MIL setting}, classification is performed on a bag of instances, i.e., a case $X_n$ with a known bag label $y_n$. The instances in the bag have no ordering and no relation among each other (i.e., the bag is a set of instances). Each instance in the bag has a label, which remains unknown while only the label of the bag is known. Under the standard binary MIL assumption, a bag is considered positive if it contains at least one positive instance~\citep{dietterich1997solving}. We frame our problem as two-level MIL. If a mammogram case (bag) contains at least one malignant image (instance), the case is considered malignant, otherwise benign (MIL at image level). Further, if a mammogram image contains at least one malignant ROI, then the image is considered malignant, otherwise benign (MIL at patch-level). 

Our case-level framework is illustrated in Fig.~\ref{fig:model-architecture}. 
Each mammogram image in a case is passed to a feature extractor module shared among all images. The feature extractor can learn relevant ROIs and aggregate them to generate image-level feature representations. 
These feature representations are passed to the image-level MIL pooling block to generate the case-level prediction. Our case-level framework is feature extractor agnostic, i.e., any kind of image-level feature extractor can be plugged into our framework (cf. Section~\ref{ssec:exp:feature-extractor-agnostic}).

\subsection{Feature Extractor} 
\label{ssec:approach:mil-model:gmic}
We compared 4 image-level feature extractors including both non-ROI extraction methods and unsupervised ROI extraction methods (cf. Section~\ref{ssec:results:feature-extractors}) and selected \gmicresnet (referred as \gmic)~\citep{shen2021interpretable} as it was the best performing feature extractor and has the capability of extracting ROI candidates in an unsupervised manner. Specifically, following \citet{shen2021interpretable}, we take a mammogram image, $I_m$ as input (cf. Fig.~\ref{fig:model-architecture} left), learn global level features using {\lmss ResNet18}~\citep{resnet_HeZRS15} feature extractor, $f_G$, which are then passed through a 1$\times$1 convolution and sigmoid ($\sigma$) to generate a class-specific saliency map $A_m$, encoding the malignant probability for all regions (cf. Eq.~\ref{eq:1}). 
The top $t$ scores from this saliency map are called \textit{top-t global features} $h^{\textit{top-t}}_m$ (cf. Eq.~\ref{eq:2}).
\begin{align}
A_{m} &= \sigma(conv_{1\times1}(f_G(I_{m}))) \label{eq:1} \\
h^{\textit{top-t}}_{m} &= \text{top-t}(A_{m}) \label{eq:2}
\end{align}
The saliency map is then used to retrieve the top-$k$ overlapping ROI candidates (patches) $R_m$ from the original image. 
\begin{equation}
R_{m} = \{p_{j}| j \in K_m\} = \text{retrieve-roi}(I_{m},A_{m},k), \label{eq:3}
\end{equation}
where $K_m$ is a set of $k$ patch indices of the ROI candidates extracted by the model. Each patch is passed through another ResNet18 feature extractor $f_L$ and the last layer feature representation of all patches is aggregated with Gated Attention (\textit{patch-level MIL})~\citep{ilse2018attention}, generating \textit{local features} $h^{local}_m$. 
\begin{equation}
h^{local}_{m} = \sum_{j \in K_m}a_jf_L(p_{j}), \label{eq:4} 
\end{equation}
where $a_j$ are the attention scores weighing the ROI candidates, generated using Gated Attention (cf. Eq. 12), where $h_{m}$ is $f_L(p_{j})$. The global features and local features are concatenated into \textit{fusion features} as:  
\begin{equation}
h^{fusion}_{m} = concat(maxpool(f_G(I_{m})), h^{local}_{m}) \label{eq:5}
\end{equation}
Thus, we have 3 feature representations for each image in a case: \textit{top-t global features}, \textit{local features} and \textit{fusion features}. 

\subsection{Image-level MIL Pooling Block}
\label{ssec:approach:mil-model:mil-pooling}

We input the feature representation of all images in a case to the image-level MIL pooling block to learn which images are important for the case-level prediction and generate a case-level score. We perform MIL pooling individually on the three types of features \textit{top-t global features}, \textit{local features} and \textit{fusion features} (cf. Fig.~\ref{fig:model-architecture}) to generate case-level scores separately for the three feature types. 

\subsubsection{Paradigm} We investigate pooling in both, instance space (IS) and embedded space (ES) for these image-level features. Let $g$ be the classification function, $\rho$ be the MIL pooling operation and $\sigma$ be the sigmoid activation.
For IS, we map the image-level features $h_m$ to image-level probability, $\hat{y}_m$.
We denote the set of all image-level probabilities for a case by $B_n$ and apply MIL pooling $\rho$ to this set to obtain the case-level probability, $\hat{y}_n$ (cf. Eq.~\ref{eq:6}). 
\begin{small}
\begin{equation}
    \text{IS: } \quad \hat{y}_{m} = \sigma(g(h_{m})), \: \hat{y}_{m}\in B_n, \: \hat{y}_n = \rho(B_{n}) \label{eq:6}
\end{equation}
\end{small}
For ES, we denote the set of image-level feature representations for a case also by $B_n$ and
apply MIL pooling $\rho$ on this set to generate a case-level feature representation $h_n$.
$h_n$ is further used to obtain a case-level logit on which we apply sigmoid activation to convert the case-level logit to the probability of the malignant class $\hat{y}_n$ (cf. Eq.~\ref{eq:7}). 
\begin{small}
\begin{align}
    \text{ES: } \quad h_{m} \in B_n,\: h_n = \rho(B_{n}), \: \hat{y}_n = \sigma(g(h_n)) \label{eq:7}
\end{align}
\end{small}
For both IS and ES, we set classification function $g$ as \textit{average} for the top-t global features and as a fully connected layer for fusion features and local features. Note, here $h_{m}$ is the generic notation for the three feature types of an image: $h^{top-t}_{m}, h^{local}_{m}$ and $h^{fusion}_{m}$. Precisely, our operations generate $\hat{y}^{top-t}_n$, $\hat{y}^{local}_n$ and $\hat{y}^{fusion}_n$ case-level predictions and $h^{top-t}_n$, $h^{local}_n$ and $h^{fusion}_n$ case-level representation (for ES paradigm).

\subsubsection{MIL Pooling Operations} We use the following four MIL pooling operations $\rho(B_{n})$ in both IS and ES paradigm: mean, max~\citep{wang2018revisiting}, attention (Att)~\citep{ilse2018attention}, and gated-attention (GAtt)~\citep{ilse2018attention} defined in Eq.~\ref{eq:8}-\ref{eq:12}. Let $b_m \in B_n$ be either the probability $\hat{y}_m$ or feature representation $h_m$ of a single image:
\begin{small}
\begin{align}
\rho_{\text{Mean}} &= \frac{1}{|B_n|}\sum_{m=1}^{|B_n|} b_{m} \label{eq:8}\\
\rho_{\text{Max}} &= \max_{m=\{1,..,|B_n|\}} b_{m} \label{eq:9} \\
\rho_{\text{(G)Att}} &= \sum_{m=1}^{|B_n|}a_{m}b_{m} \label{eq:10}
\end{align}
\end{small}

The attention scores $a_{m}$ weighing the images in a case differ for Att~\citep{ilse2018attention} and GAtt~\citep{ilse2018attention} as follows: 
\begin{small}
\begin{align}
a_{m}^{\text{Att}} &= 
\frac{\exp\{\mathbf{w}^\top\tanh(\mathbf{V}\mathbf{h}_{m}^\top)\}}{\sum_{k=1}^{|B_n|}\exp\{\mathbf{w}^\top \tanh(\mathbf{V}\mathbf{h}_{k}^\top)\}} \label{eq:11} \\
a_{m}^{\text{GAtt}} &= \frac{\exp\{\mathbf{w}^\top(\tanh(\mathbf{V}\mathbf{h}_{m}^\top)\odot \sigma(\mathbf{U}\mathbf{h}_{m}^\top))\}}{\sum_{k=1}^{|B_n|}\exp\{\mathbf{w}^\top(\tanh(\mathbf{V}\mathbf{h}_{k}^\top)\odot \sigma(\mathbf{U}\mathbf{h_k}^\top))\}} \label{eq:12}
\end{align}
\end{small}
where $\mathbf{w}, \mathbf{V}, \mathbf{U}$ are trainable parameters, $\odot$ is element-wise multiplication. 

\subsubsection{Domain-specific Pooling Block}
\label{ssec:approach:domain-pooling}
In standard MIL, all instances of a bag are conceptually equal, resulting in image-wise aggregation of the feature embedding \textit{(image-wise pooling)} as shown in Eq.~\ref{eq:8}-\ref{eq:10}, where set $B_n$ contains feature representations (ES paradigm) or probabilities (IS paradigm) of all images in a case.
A mammography case, however, typically contains images of both breast sides (left and right).
Malignancies can either not occur, occur on one side or on both sides. 
We propose a domain-specific pooling block performing side-wise aggregation in embedded space \textit{(side-wise pooling)} to find and weigh the condition of each side separately for creating the aggregated case-level feature embedding (cf. Fig.~\ref{fig:model-architecture}, right).  
We first combine the feature embeddings of the views per side (view $v \in V$) with pooling attention (view-level pooling) as:   
\begin{small}
\begin{align}
h^{top-t}_{L} &= \sum_{v \in V} a^{top-t}_{Lv} h^{top-t}_{Lv} & h^{top-t}_{R} &= \sum_{v \in V} a^{top-t}_{Rv} h^{top-t}_{Rv} \label{eq:13} \\
h^{local}_{L} &= \sum_{v \in V} a^{local}_{Lv} h^{local}_{Lv} & h^{local}_{R} &= \sum_{v \in V} a^{local}_{Rv} h^{local}_{Rv} \label{eq:14} \\
h^{fusion}_{L} &= \sum_{v \in V} a^{fusion}_{Lv} h^{fusion}_{Lv} & h^{fusion}_{R} &= \sum_{v \in V} a^{fusion}_{Rv} h^{fusion}_{Rv} \label{eq:15}
\end{align}
\end{small}
where subscript $L$ and $R$ indicate the left and right breast. We then combine the resulting feature embedding per breast by another pooling attention operation (side-level pooling) as: 
\begin{small}
\begin{align}
h^{top-t}_{n} =\sum_{s\in S} a^{top-t}_{s} h^{top-t}_{s} \qquad\qquad\qquad \: h^{local}_{n} = \sum_{s\in S} a^{local}_{s} h^{local}_{s} \qquad\qquad\qquad h^{fusion}_{n} = \sum_{s\in S} a^{fusion}_{s} h^{fusion}_{s} \label{eq:16}
\end{align}
\end{small} 
We use separate attention block for the view-level and the side-level pooling for each feature type. These three types of case-level feature representations are used to generate three case-level scores in the ES paradigm. We refer to this model as \ESAttSide. We also develop a version of this model in IS paradigm by applying the attention scores from the view-level and side-level pooling to the image-level probabilities to generate a case-level score. We refer to this model as \ISAttSide. We compare our proposed domain-specific side-wise pooling model, \ISAttSide in IS paradigm and \ESAttSide in ES paradigm to image-wise pooling across IS and ES paradigms with various MIL pooling operations and refer to these models as \ISMeanImg, \ISMaxImg, \ISAttImg, \ISGattImg, \ESMeanImg, \ESMaxImg, \ESAttImg and \ESGattImg (cf. Section~\ref{ssec:exp:mil-comparison}).

\subsection{Variable Number of Images per Case in MIL}
In real hospital settings, a variable number of images is available per patient. MIL is inherently capable of handling variable bag sizes. 
However, it fails to do so due to the way pytorch implements keeping track of past updates for adaptive optimizers such as Adam~\citep{kingma2017adam}. Even if a component is not present in the forward pass and hence should not change, the parameters are still updated via the gradient history \textit{(default training)}.
This affects in particular the (gated) attention modules, which for example do not need to be updated when only a single image is present in the bag. We propose a \textit{dynamic training} approach to handle any number of input images without such undesired parameter updates. We group training batches by same combination of image types, e.g., L-CC+L-MLO in one batch, L-CC+R-CC+R-MLO in another. After each mini-batch weight update, we reset the weights and the optimizer state of the unused components to the last state, where weights were updated directly from both the input and gradient history and not from the gradient history alone. Dynamic training is specifically applicable for our (G)Att MIL models. For (G)Att$^{img}$ MIL, we switch off updates of attention modules, if only a single image is present. In Att$^{side}$, if a single image is present per side, we switch off the view-level attention module and if images are present for one side, we switch off the side-level attention module. The differences between default and dynamic training are shown in detail in Table~\ref{tab:approach:default-vs-dynamic}.   
\begin{table}[th!bp]
\centering
\caption{Training of variable-image cases in \ESAttSide and \ESAttImg: default and dynamic training scheme. The presence of gradients (grad), update from gradient history (hist) and attention pooling weights update (att-wt) are denoted by (\checkmark) vs their absence denoted by (\ding{55}). 1L/1R denotes 1 image from Left (L) breast or 1 image from right (R) breast, nL/mR denotes n>1 images from left breast or m>1 images from right breast, 1L+1R denotes 1 image from left and 1 image from right breast and nL+mR denotes n>1 images from left and m>1 images from right breast.}
\label{tab:approach:default-vs-dynamic}
\resizebox{\textwidth}{!}{%
\begin{tabular}{+l^c^c^c^c^c^c^c^c^c^c^c^c^c^c^c^c^c^c}
\toprule \tabhead
 & \multicolumn{12}{c}{\textbf{\ESAttSide}} & \multicolumn{6}{c}{\textbf{\ESAttImg}} \\
\cmidrule(l{2pt}r{2pt}){2-13}\cmidrule(l{2pt}r{2pt}){14-19}
\textbf{Group} & \multicolumn{6}{c}{\textbf{Default}} & \multicolumn{6}{c}{\textbf{Dynamic}} & \multicolumn{3}{c}{\textbf{Default}} & \multicolumn{3}{c}{\textbf{Dynamic}} \\ 
\cmidrule[\heavyrulewidth](l{2pt}r{2pt}){2-7}\cmidrule[\heavyrulewidth](l{2pt}r{2pt}){8-13}\cmidrule[\heavyrulewidth](l{2pt}r{2pt}){14-16}\cmidrule[\heavyrulewidth](l{2pt}r{2pt}){17-19}
 &  \multicolumn{3}{c}{view-level} &  \multicolumn{3}{c}{side-level} &  \multicolumn{3}{c}{view-level} &  \multicolumn{3}{c}{side-level} & \multicolumn{3}{c}{image-wise} & \multicolumn{3}{c}{image-wise}\\ 
 \cmidrule(l{2pt}r{2pt}){2-4}\cmidrule(l{2pt}r{2pt}){5-7} \cmidrule(l{2pt}r{2pt}){8-10}\cmidrule(l{2pt}r{2pt}){11-13}\cmidrule(l{2pt}r{2pt}){14-16}\cmidrule(l{2pt}r{2pt}){17-19}
& grad & hist & att-wt & grad & hist & att-wt & grad & hist & att-wt & grad & hist & att-wt & grad & hist & att-wt & grad & hist & att-wt \\
  \cmidrule(l{2pt}r{2pt}){2-4}\cmidrule(l{2pt}r{2pt}){5-7} \cmidrule(l{2pt}r{2pt}){8-10}\cmidrule(l{2pt}r{2pt}){11-13}\cmidrule(l{2pt}r{2pt}){14-16}\cmidrule(l{2pt}r{2pt}){17-19}
1L/1R & \ding{55} & \checkmark & \checkmark & \ding{55} & \checkmark & \checkmark & \ding{55} & \ding{55} & \ding{55} & \ding{55} & \ding{55} & \ding{55} & 
\ding{55} & \checkmark & \checkmark &
 \ding{55} & \ding{55} & \ding{55} \\
nL/mR & \checkmark & \checkmark & \checkmark & \ding{55} & \checkmark & \checkmark & \checkmark & \checkmark & \checkmark & \ding{55} & \ding{55} & \ding{55}  & 
\checkmark & \checkmark & \checkmark &
 \checkmark & \checkmark & \checkmark \\
1L+1R & \ding{55} & \checkmark & \checkmark & \checkmark & \checkmark & \checkmark & \ding{55} & \ding{55} & \ding{55} & \checkmark & \checkmark & \checkmark &
\checkmark & \checkmark & \checkmark &
 \checkmark & \checkmark & \checkmark \\
nL+mR & \checkmark & \checkmark & \checkmark & \checkmark & \checkmark & \checkmark & \checkmark & \checkmark & \checkmark & \checkmark &\checkmark  & \checkmark &
\checkmark & \checkmark & \checkmark &
 \checkmark & \checkmark & \checkmark \\ \bottomrule
\end{tabular}}
\end{table}

\subsection{Training Loss}
We follow the loss definition by  GMIC~\citep{shen2021interpretable}, but instead of calculating the loss at the image level, we calculate it at the case level as: 
\begin{align}
L(y_n, \hat{y_n}) = BCE(y_n, \hat{y}^{top-t}_n) + BCE(y_n, \hat{y}^{local}_n) + BCE(y_n, \hat{y}^{fusion}_n) + \beta \sum_{m=1}^{|X_n|} |A_{n,m}| \label{eq:loss}
\end{align}
where BCE is binary cross entropy, $\beta$ controls the influence of the regularization term, $|A_{n,m}|$ is the L1 norm of the saliency map of an image, $y_n$ is the groundtruth case label and $\hat{y}_n$ is the predicted case label of the n\textsuperscript{th} case. 
At test time, we predict the case label from the fusion features, $\hat{y_n}^{fusion}$ as in the original paper~\citep{shen2021interpretable}.

\section{Experimental Setup}
In this section, we describe the datasets, preprocessing and settings for model training and evaluation. Our retrospective study was approved by the institutional review board of Hospital Group Twente (ZGT), The Netherlands.

\subsection{Datasets}
We evaluate our method on the two public benchmarks CBIS-DDSM (1.6k cases) and VinDr (5k cases) and the private dataset \mgm (21k cases). Table~\ref{tab:dataset:mammograms} shows an overview of all datasets. 

\textbf{CBIS-DDSM (CBIS)}~\citep{cbisddsm} is a public dataset adapted in 2016 from the DDSM dataset, which was collected in 1997 and it comprises 3,103 mammograms from 1,566 patients including only CC and MLO views. It contains manually extracted ROIs with groundtruth labels at the ROI level. 
We transfer the ROI-level labels to the image level (R$\rightarrow$I) by assigning malignant when any ROI (lesion) in the image has the label malignant, otherwise we label the image as benign. Further, we infer case-level labels from the image labels (I$\rightarrow$C) with the groundtruth of a case as malignant if any image in a case is malignant, otherwise benign. For \cbis, a case is defined as all images taken for a patient for a specific abnormality (mass or calcification). The case-level dataset contains 753 cases of calcification and 892 cases of mass, resulting in 1,645 cases in \cbis. 

\textbf{VinDr}-Mammo~\citep{Nguyenvindr} is a public dataset from two hospitals in Vietnam, consisting of 5,000 4-image cases collected between 2018 and 2020. 
Each image has a BI-RADS~\citep{sickles2013acr} category assigned, i.e., the  radiologists' assessment of probability of malignancy, ranging from 1 to 5 (least to most malignant) for \vindr. A case in VinDr is defined as the 4 images taken during a mammography exam. We assign the highest BI-RADS category among the images in a case as the case-level label (I$\rightarrow$C). 
Then, we map BI-RADS categories 4 and 5 to malignant, others to benign, following~\citet{carneiro2015unregistered,zhu2017deep}, resulting in 4,519 benign and 481 malignant cases. 

\begin{table}[th!bp]
\centering
\setlength{\extrarowheight}{1pt}
\caption{Dataset Statistics of \cbis, \vindr and \mgm. It shows the number of patients, cases and images, BI-RADS scores (Bi-S), whether the groundtruth labels are available on ROI (R), image (I) or case (C) level and number of malignant cases where the case label is not equal to all image labels in the case (CL$\neq$IL).
Plots of class distribution (B: benign, M: malignant)
and case distribution grouped by the number of images per case. \mgmfv contains only the cases with 4 standard (4-std.) images (light green bar in Case Dist). $^\dagger$One of the cases of \vindr has two LCC images. We used one of these images for our MIL models and both images for SIL models.}
\label{tab:dataset:mammograms}
\begin{tabular}{+lccc}
\toprule 
\textbf{Stats.}  & \bf\cbis & \bf \vindr & \bf \mgmfv / VV \\ \otoprule
Patients        & 1,566  & n.a. &  15,170 / 15,991 \\
Cases        & 1,645 & 5,000$^\dagger$ & 19,614 / 21,013 \\
Images        & 3,103 & 20,000 &  78,456 / 84,299 \\
Bi-S           & 0 to 5 & 1 to 5 & 0 to 6 \\
Label & R$\rightarrow$I$\rightarrow$C & R$\rightarrow$I$\rightarrow$C & C \\
CL$\neq$IL     &   21/774 (3\%)     &   468/481 (97\%)     &   n.a.     \\
{Class Dist.}& 
    \raisebox{-.5\totalheight}{\includegraphics[scale=0.155]{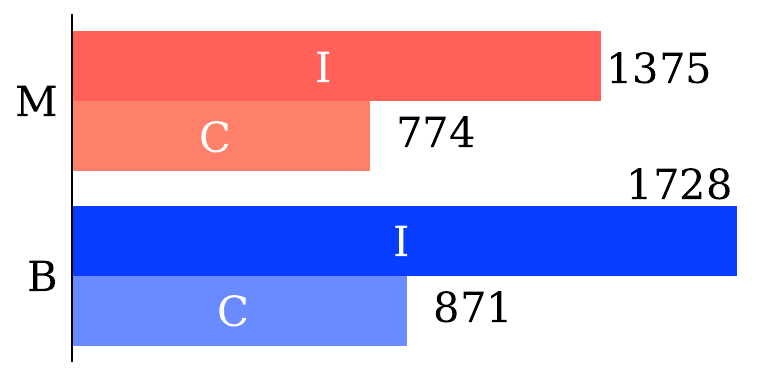}}  & 
    \raisebox{-.5\totalheight}{\includegraphics[scale=0.155]{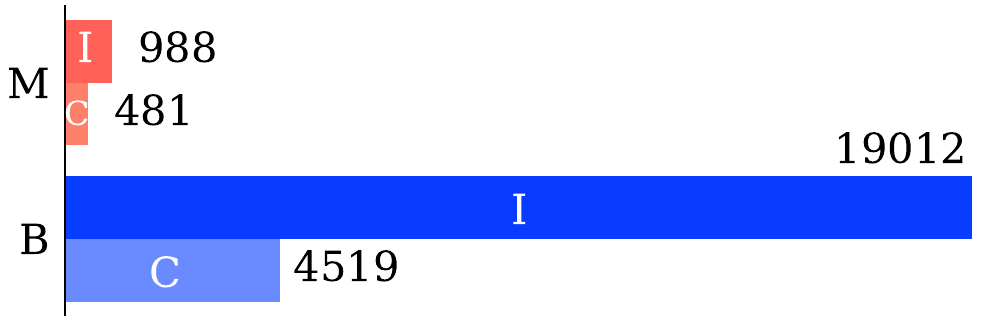}} & 
    \raisebox{-.5\totalheight}{\includegraphics[scale=0.155]{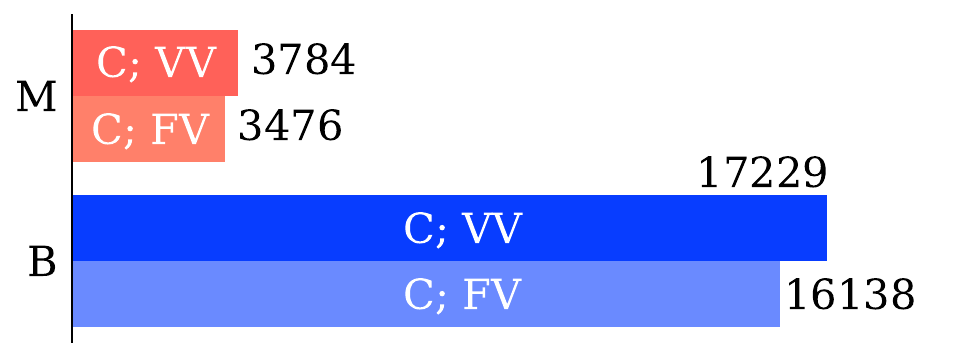}} \\ 
{Case Dist.} & 
    \raisebox{-.55\totalheight}{\includegraphics[scale=0.18]{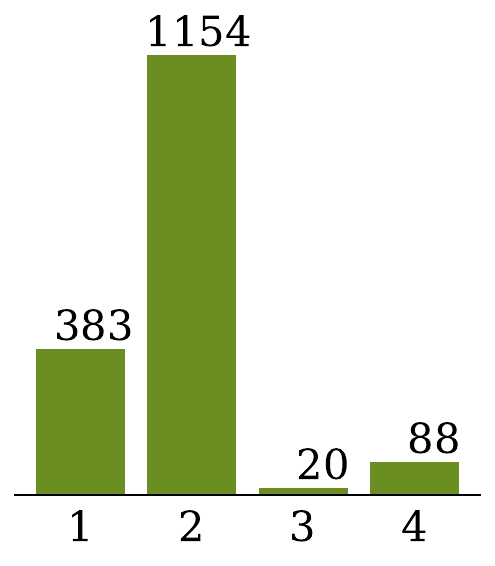}} &
  \raisebox{-.55\totalheight}{\includegraphics[scale=0.18]{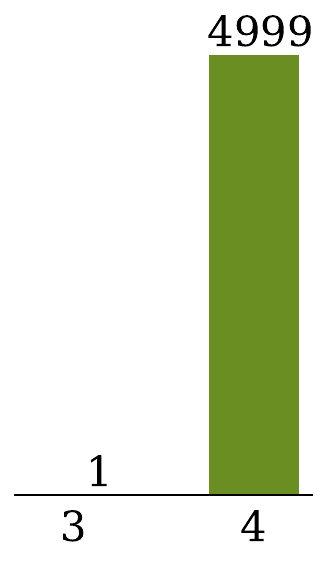}} & 
  \raisebox{-.5\totalheight}{\hspace{-0.6cm}\includegraphics[scale=0.18]{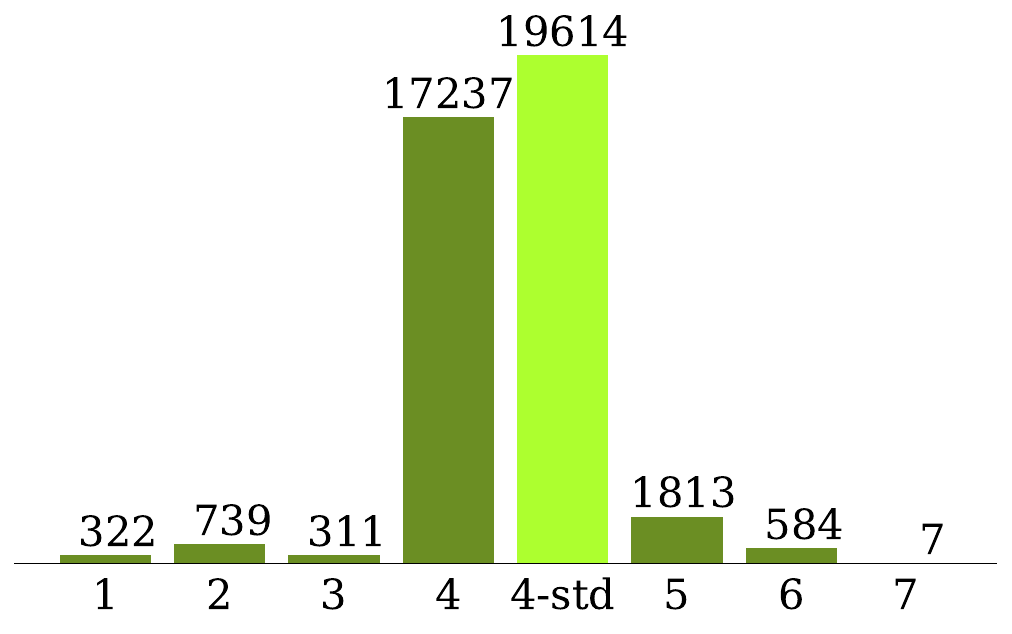}} \\\bottomrule
\end{tabular}
\end{table}

Our private dataset, \textbf{\mgm} collected from Hospital Group Twente (ZGT), The Netherlands between 2013 and 2020 contains 21,013 mammogram cases (17,229 benign and 3,784 malignant) from 15,991 patients.
Cases were either referred from a national screening program, sent to the hospital by general practitioners, or in-house patients requiring a diagnostic mammogram in the hospital. 
To investigate the impact of variable number of views, we create two versions of our dataset. \textbf{\mgmfv} contains the cases with the 4 standard views, i.e., CC and MLO, from L(eft) and R(ight) breast side~\citep{kim2018applying,Wu_2020}, and \textbf{\mgmvv} contains cases with all views taken during mammography, i.e., any of CC, MLO, LM, ML and XCCL (cf. Section~\ref{ssec:approach:problem-def} for full form of the views).
Table~\ref{tab:dataset:mammograms} Case Dist. shows the histogram of images per case in our \mgm dataset (\emph{4-std} bar in light green is the \mgmfv dataset and the complete dataset is \mgmvv). Note that the group with 4 images in Case Dist. refers to cases with any type of 4 images, whereas, \emph{4-std} are the cases having 4 images of standard views (93\% of cases in \mgmvv). 4-std includes the standard views of cases with an additional view (5, 6 or 7 views), hence there are more 4-std cases than 4-non-standard cases.
The cases in our dataset are labeled as either malignant or benign, where benign includes both normal cases with no tumor and cases with benign tumor.
In a standard hospital care setting, the diagnosis for a breast cancer patient (malignant or benign) is assigned after following the full diagnostic pathway. We extracted the pathways and assigned the label of the final diagnosis to each mammogram case. To obtain the final diagnosis, we used the financial code (Diagnosis Treatment Combination, DBC code) of a patient, because this is the most accurate information available in this hospital.
Thus, our assigned groundtruth reflects the true diagnosis of the patient.  

We further collected image-level and ROI annotations from two breast radiologists for a small subset of the \mgm dataset, \textbf{\mgmannot} containing 100 cases with 50 benign and 50 malignant cases and 413 images. We use this subset to validate the fine-grained results of our model in Section~\ref{sec:experiments-results}.   

\subsection{Data Preprocessing}
We apply the following preprocessing to all datasets. 
We convert images from DICOM~\citep{dicom} format to PNG following~\citet{kim2018applying} and~\citet{shen2019deep}. We save 16 bit images for~\cbis,~\vindr and 12 bit images for~\mgm following the bit depth of the image in DICOM format.  
To remove irrelevant information, such as burned-in annotations and excess background, we first find the contour mask covering the largest area, i.e., the region of the breast.
This mask is then used to extract the breast region from the original image, automatically leading to the removal of surrounding burned-in annotations. We finally use a bounding box around the extracted portion to crop any excess background (cf. Fig. \ref{sup:fig:data:preprocessing} for example images. All preprocessing code is available in the repository\textsuperscript{\ref{fn:repository}}).

\subsection{Data Augmentation} 
We follow data augmentation reported in the original papers of the respective SoTA model and use our own values for some details that were missing in the original papers.

For \kimrn~\citep{kim2018applying}, we randomly perturb the brightness and contrast by 10\%, resize to 1600x1600 and normalize the images to the range of [-1,1]. Before resizing, we zero-pad the shorter side of the image to 1600 (left side for RCC/RMLO and right side for LCC/LMLO). The images are not flipped horizontally to retain their original orientation.

For RGP and GGP models~\citep{shu2020deep}, we resize the images to 800x800, followed by normalization within the range [0,1]. Then, the images are randomly flipped horizontally with a probability of 0.5, contrast and saturation are randomly perturbed by 20\%, followed by random rotation of 30 degrees and addition of gaussian noise with mean 0 and standard deviation 0.005.

For MIL models with the \gmic feature extractor~\citep{shen2021interpretable}, we resize the images to 2944x1920 for \mgm and \cbisddsm following the original paper. We resize \vindr to 2700x990, as the original image size is smaller than the resize dimension used in GMIC (specifically the width). So, we resize the images in \vindr to the mean of height and width of all images in the dataset. We randomly flip the images with probability 0.5, apply random affine transformations consisting of rotation by 15 degrees, translation up to 10\% of the image size, scaling by a random factor between 0.8 and 1.6, random shearing by 25 degrees, and added gaussian noise of mean 0 and std dev 0.005.

During inference, the images were only resized and normalized. The single channel of grayscale images was duplicated into 3 channels to fit pretrained models and normalized to ImageNet mean and standard deviation. 
For training from-scratch models (DIB-MG), only the single grayscale channel was used as input and normalized to the range of [-1,1]~\citep{kim2018applying}.

\subsection{Training and Evaluation} 
For \cbis, we used the official train-test split (\cbisoff) for feature extractor selection in the SIL setting (Table~\ref{tab:results:feature-extractor}). 
For  MIL vs SIL (Table~\ref{tab:results:sil-vs-mil}) and MIL models (Table~\ref{tab:results:mil-pooling}) we used a custom label-stratified split with 15\% test set (\cbiscus) such that all cases of a patient were exclusively contained in the same subset. For \vindr, we used the official train-test split for all models and for \mgm, we used a label-stratified split with 15\% test set. 
The training sets were further divided into 90-10\% train-validation splits and all cases of a patient were exclusively contained in the same subset. SIL models were trained on the same train-val-test splits as MIL. In Table~\ref{tab:results:mil-pooling}, MIL models on \cbiscus, in Table~\ref{tab:results:sil-vs-mil}, \ESAttSide  on \cbiscus and \mgmvv and in Table~\ref{tab:results:sota-comparison}, \ESAttSide on \cbis are trained with default training.

For hyperparameter selection, we performed a random search over 20 hyperparameter combinations of learning rate, weight decay and regularization term $\beta$ for {\lmss GMIC} with $t=2\%$ for top-t features in the SIL setting on \cbis and selected the combination with the highest AUC on the validation set. We used these hyperparameters for all models with GMIC as feature extractor. 

All models were trained using Adam, except {\lmss \kimrn}, which used SGD.
We set the batch size (bs) to the largest possible value for the GPU memory for each experiment (indicated in respective table captions). In SIL, batch size refers to the number of images and in MIL to the number of cases. In all models, we used a weighted cost function for training by upweighting the error of the malignant class by the ratio of benign to malignant cases. All images from the right breast side were flipped horizontally. 

We used PyTorch 1.11.0, CUDA 11.3 and ran all experiments on a single GPU (Nvidia A6000 and A100). For reproducible results, we fixed the random seeds of weight initialization, data loading and set CUDA to behave deterministically. We ran models with 2 (\mgm) or 3 (other datasets) different random seeds for data splits and report mean and standard deviation. We report F1 score and AUC score. More details for reproducing our work (e.g. hyperparameters and random seeds) can be found in our repository\textsuperscript{\ref{fn:repository}}. We also provide the configuration files to setup model training in our repository to support benchmarking the models. 

\section{Experiments and Results}
\label{sec:experiments-results}
In this section, we describe the experiments we conducted, along with their guiding questions, obtained results and conclusions we can draw from them. Our side-wise MIL pooling model \ESAttSide trained on case-level labels works equally well as model trained on individual image labels, and case labels are not transferable to individual images (Section~\ref{ssec:exp:mil-vs-sil}). Our side-wise MIL pooling improves classification performance over common MIL pooling methods, in particular in the embedded-space paradigm (Section~\ref{ssec:exp:mil-comparison}). We quantitatively compare multiple MIL approaches in their ability to identify relevant images (Section~\ref{ssec:exp:relevant-images}) and extract correct ROIs in a case (Section~\ref{ssec:expsetup:unsupervised-roi}) and show that our model can identify both with satisfying quality. Our qualitative evaluation with clinicians also suggests the same  (Section~\ref{ssec:exp:unsupervised-roi:qualitative}). Evaluating whether our model is right for the right reasons, we find that in the majority of correctly classified cases it also extracts the correct ROI and that some cases are misclassified despite extracting the correct ROI (Section~\ref{ssec:exp:right-for-right-reason}). Our side-wise MIL pooling provides robust improvements across different feature extractors (Section~\ref{ssec:exp:feature-extractor-agnostic}). Our model \ESAttSide outperforms five SoTA baseline methods in F1 score on classification performance (Section~\ref{ssec:exp:comparison-to-sota}). Model trained on fixed-image cases perform reasonably well on variable-image cases during test time (Section~\ref{ssec:exp:variable-images}).

\subsection{MIL vs SIL Classification Performance}
\label{ssec:exp:mil-vs-sil}
\textit{Our goal: How do models trained on case labels compare to models trained on image labels in terms of classification performance? Can we use case-level labels that are readily available in real hospitals settings for model training without the need of any manual annotation of images?}
In Table~\ref{tab:results:sil-vs-mil}, we compared our proposed case-level model \ESAttSide (MIL) to image-level GMIC models (SIL) trained with two SIL settings: \SILil, where each image has its true label and \SILcl, where the case label is transferred to all images in that case. We report performance for the SIL models at the case level and image level (P. Level in the table). At the image level for SIL models, we always evaluate with the true image labels and at the case level for SIL models, we assign a case as malignant if any of the images within that case are predicted as malignant by the SIL model. Note that this setup does not assign probabilities to cases, thus for case-level prediction with SIL models, AUC score is marked as n.a.
\cbis and \vindr contain both image and case labels, whereas \mgm contains only the latter, thus \SILil results are not reported on \mgmvv. 

Table~\ref{tab:results:sil-vs-mil} shows that the case-level model \ESAttSide generally has similar or better performance than models trained on image labels (\SILil) on all datasets. 
\begin{table}[thbp]
\centering
\caption{SIL vs. MIL comparison on \cbiscus, \vindr, \mgmvv; SIL (bs=10). Prediction level (P. Level). The case-level model \ESAttSide outperforms SIL models in terms of F1 score, suggesting manual annotation of images is not needed for training breast cancer models.}
\label{tab:results:sil-vs-mil}
\begin{small}
\begin{tabular}{+llcccccc}
\toprule \tabhead
Model & \textbf{P. Level} & \multicolumn{2}{c}{\textbf{\cbiscus}} & \multicolumn{2}{c}{\textbf{VinDr}} & \multicolumn{2}{c}{\textbf{\mgmvv}} \\ 
& & \textbf{F1} & \textbf{AUC} & \textbf{F1} & \textbf{AUC} & \textbf{F1} & \textbf{AUC} \\ \otoprule
\SILil  & Image  & $0.66 \pm 0.03$ & $\textbf{0.79} \pm 0.05$ & $0.27 \pm 0.01$ & $\textbf{0.83} \pm 0.01$ & n.a. & n.a. \\
\SILil  & Case  & $0.68 \pm 0.03$ & n.a. & $0.30 \pm 0.00$ & n.a. & n.a. & n.a. \\
\SILcl  & Image  & $0.65 \pm 0.02$ & $0.77 \pm 0.01$ & $0.23 \pm 0.05$ & $0.82 \pm 0.01$ & $0.44 \pm 0.01$ & $0.75 \pm 0.01$ \\
\SILcl  & Case  & $0.67 \pm 0.02$ & n.a. & $0.26 \pm 0.05$ & n.a. & $0.44 \pm 0.01$ & n.a.\\[3pt]
\ESAttSide & Case  & $\textbf{0.70} \pm 0.01$ & $0.78 \pm 0.02$ & $\textbf{0.48} \pm 0.03$ & $\textbf{0.83} \pm 0.02$ & $\textbf{0.61} \pm 0.01$ & $\textbf{0.85} \pm 0.00$ \\\bottomrule
\end{tabular}
\end{small}
\end{table}
This suggests that weak case-level labels are as good or even better for training breast cancer prediction models than exact image labels. Thus, \textbf{manual labelling of images is not needed for model training and case labels, which are readily available at the hospital can be used}. Further, \SILil models outperform \SILcl models, which suggests that transferring case labels to individual images can result in error-prone training of image-level models. Thus, \textbf{case labels of mammography exams are not transferable to images for model training without loss in performance}.  
The performance difference in \SILil and \SILcl models is less pronounced for~\cbis compared to \vindr as \cbis has only 3\% of malignant cases where the labels for cases and images do not match, while \vindr has 97\% of such cases (cf. Table~\ref{tab:dataset:mammograms}).

\subsection{Comparison of Image-level MIL pooling} 
\label{ssec:exp:mil-comparison}
\emph{Our goal: How does our proposed side-wise MIL pooling perform in comparison to other MIL pooling methods in terms of classification performance?} We compared our proposed side-wise pooling Att$^{side}$ with image-wise ($^{img}$) MIL pooling operations (Mean, Max, Att and GAtt) across both IS and ES paradigm. 

Table~\ref{tab:results:mil-pooling} shows our proposed side-wise pooling model \ESAttSide generally outperforms the other MIL models across all datasets.
\begin{table}[thbp]
\centering
\caption{Comparison of various MIL pooling approaches in instance (IS)  and embedded (ES) space. Batch sizes (bs) of \cbiscus: bs=3, \vindr: bs=7, and \mgmfv: bs=5. Our side-wise pooling \ESAttSide outperforms other pooling approaches. *trained with gradient accumulation.}
\label{tab:results:mil-pooling}
\begin{small}
\begin{tabular}{+l^c^c^c^c^c^c}
\toprule \tabhead
\textbf{Model} & \multicolumn{2}{c}{\textbf{\cbiscus}} & \multicolumn{2}{c}{\textbf{VinDr}} & \multicolumn{2}{c}{\textbf{\mgmfv}}\\ 
 & \textbf{F1} & \textbf{AUC} & \textbf{F1} & \textbf{AUC} & \textbf{F1} & \textbf{AUC} \\ \otoprule
\ISMeanImg    & $0.64 \pm 0.04$ & $0.77 \pm 0.02$ &  $0.42 \pm 0.05$ & $0.80 \pm 0.02$ & $0.46 \pm 0.02$ & $0.76 \pm 0.03$* \\
\ISMaxImg   & $0.67 \pm 0.03$ & $0.76 \pm 0.07$ & $0.46 \pm 0.05$ & $0.81 \pm 0.02$ & $0.46 \pm 0.07$ & $0.76 \pm 0.05$*  \\
\ISAttImg  & $0.66 \pm 0.03$ & $0.78 \pm 0.02$ & $0.42 \pm 0.04$ & $0.81 \pm 0.02$ & $0.54 \pm 0.03$ & $0.83 \pm 0.02$*\\ 
\ISGattImg   & $0.66 \pm 0.04$ & $0.77 \pm 0.01$ & $0.41 \pm 0.02$ & $0.81 \pm 0.02$ & $0.53 \pm 0.02$ &  $0.82 \pm 0.01$* \\ 
\ISAttSide  & $0.67 \pm 0.03$ & $\textbf{0.79} \pm 0.02$ &  $0.46 \pm 0.08$ &  $0.80 \pm 0.01$ & $0.51 \pm 0.00$ & $0.81 \pm 0.01$* \\ [3pt]
\ESMeanImg  & $0.65 \pm 0.04$ & $0.76 \pm 0.02$ & $0.34 \pm 0.04$ & $0.80 \pm 0.00$ & $0.54 \pm 0.02$ & $0.82 \pm 0.03$ \\
\ESMaxImg  & $0.63 \pm 0.04$ & $0.76 \pm 0.01$ & $0.46 \pm 0.06$ & $0.80 \pm 0.01$ & $0.52 \pm 0.03$ & $0.82 \pm 0.00$ \\
\ESAttImg  & $0.63 \pm 0.05$ & $0.75 \pm 0.05$ & $0.44 \pm 0.04$ & $0.81 \pm 0.01$ & $0.56 \pm 0.02$ & $0.84 \pm 0.01$ \\ 
\ESGattImg  & $0.63 \pm 0.08$ & $0.74 \pm 0.07$ & $\textbf{0.48} \pm 0.04$ & $0.82 \pm 0.01$ & $0.55 \pm 0.03$ & $0.83 \pm 0.02$ \\ 
\ESAttSide   & $\textbf{0.70} \pm 0.01$ & $0.78 \pm 0.02$ & $\textbf{0.48} \pm 0.03$ & $\textbf{0.83} \pm 0.02$ & $\textbf{0.59} \pm 0.00$ & $\textbf{0.85} \pm 0.01$ \\
\bottomrule
\end{tabular}
\end{small}
\end{table}
Also, within IS and ES paradigm, our side-wise pooling block Att$^{side}$ has highest or competitive performance to other MIL pooling method. This shows that \textbf{side-wise MIL pooling improves over image-wise pooling in the embedded-space paradigm}. 
Comparing IS to ES per individual pooling method, we cannot conclude that one paradigm is superior over the other as we observe mixed results (IS generally performs better in \cbis, ES better in \vindr and \mgmfv).
Our side-wise pooling model \ESAttSide always outperforms simple averaging of instance scores (\ISMeanImg) and feature representation (\ESMeanImg). 
Note that \mgmfv in IS paradigm was trained with gradient accumulation with batch size 5 and accumulation step 2.\footnote{due to sudden unavailability of the GPU which could fit batch size 5} Table \ref{sup:tab:results:grad-accum} shows training with gradient accumulation results in slightly lower performance than training without. However, even with gradient accumulation our \ESAttSide achieves highest F1 score on \mgmfv. 

\subsection{Quantitative Assessment of Ability to Identify Relevant Images}
\label{ssec:exp:relevant-images}
\begin{figure}[thbp]
\centering
    \includegraphics[scale=0.20]{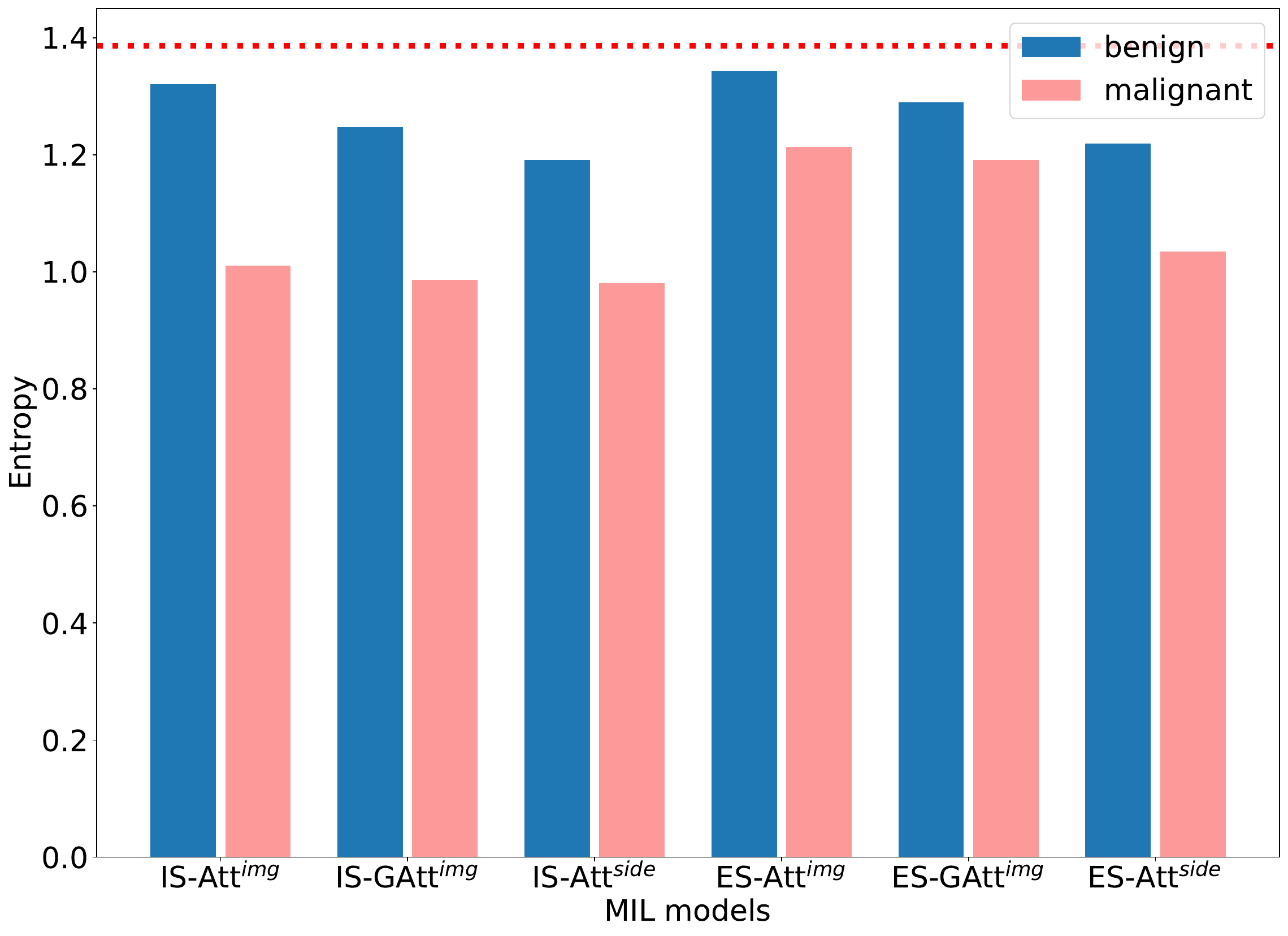}
    \caption{Entropy of attention score distribution of attention-based models for malignant and benign class for \mgmfv. Red dashed line shows the entropy for uniform weights of all views, i.e., attention score of 0.25 for four views. Bars for the malignant class are away from uniform distribution indicating that the models can differentiate well among the images in malignant cases.}
    \label{fig:attwt-entropy-zgt}
\end{figure}

\emph{Our goal: To what extent can MIL models find the important (malignant) images in the malignant cases and how does our proposed model \ESAttSide perform in this aspect compared to the other MIL and SIL models?}
We first analyzed whether the MIL models are learning to focus differently on the images. We inspected the entropy of the attention score distribution of the images for malignant and benign cases on \mgmfv. Fig.~\ref{fig:attwt-entropy-zgt} shows that the attention score distribution of images for both malignant and benign cases differs from the uniform distribution (shown by the red dashed line), indicating that the MIL models are assigning unequal importance to the images. 
Attention scores vary more for malignant than for benign cases as we would expect, which suggests that MIL models are learning to differentiate well among the images in a malignant case and hopefully, to focus on the images with relevant ROIs. 

Next, to understand the extent to which MIL models can identify important, i.e., malignant, images in malignant cases, we used the importance given to the images by the MIL models to assign proxy labels to these images and calculated their match with the groundtruth of these images. 
For IS and ES attention pooling based models, we used the attention scores assigned to the images in the MIL models as proxy to assign predicted image labels. 
We converted the attention score to predicted image labels for all truly malignant cases as follows: image $k$ with attention score $a_k>0.25$ was assigned malignant, otherwise benign. Then, we calculated the F1 score with respect to the true image-level labels. For IS models, we additionally used the image probability as a proxy to assign predicted image labels (malignant if probability $>0.5$, otherwise benign). We performed this evaluation only for malignant cases as all images in a benign case have the same groundtruth. We performed this analysis on \vindr, as this is the largest dataset with known image labels in our paper and has the highest percentage of cases with mixed image labels (Table~\ref{tab:dataset:mammograms} CL$\neq$IL) and also on \mgmannot.

\begin{figure}[htbp]
\centering
\begin{subfigure}[b]{0.5\textwidth}
        \centering
        \includegraphics[scale=0.175]{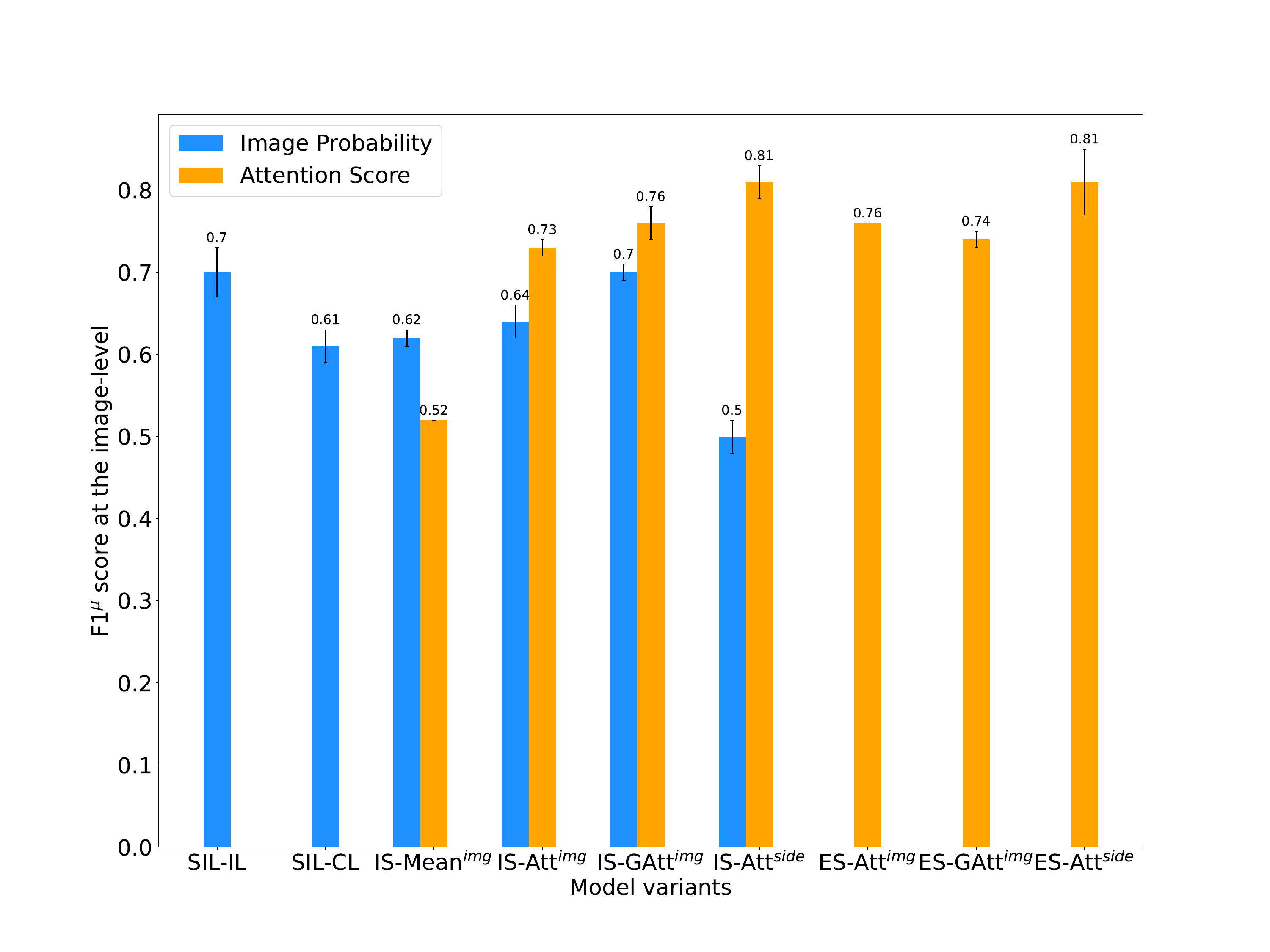}
        \caption{\vindr}
    \end{subfigure}%
    \begin{subfigure}[b]{0.5\textwidth}
        \centering
        \includegraphics[scale=0.17]{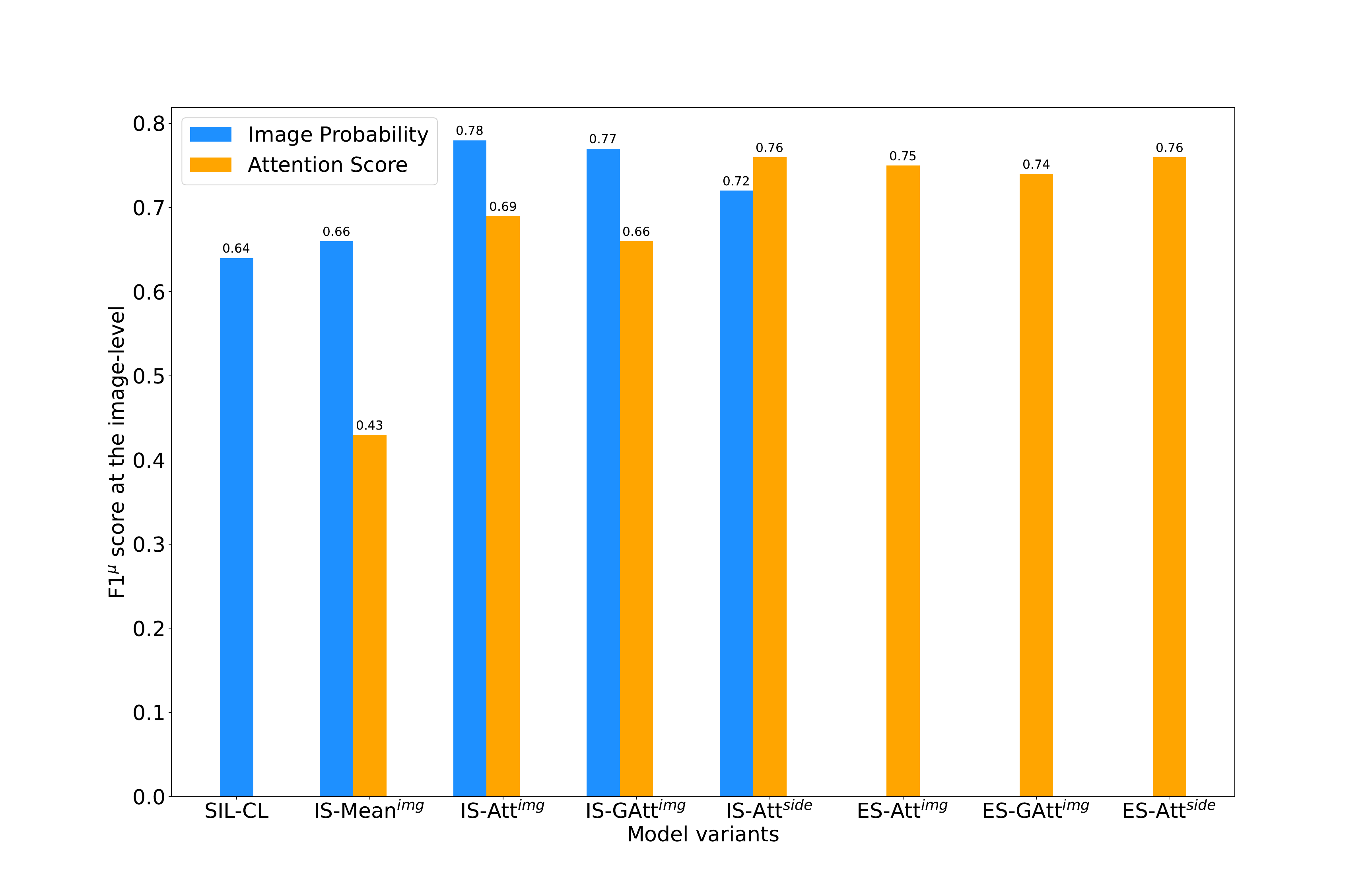}
        \caption{\mgmannot}
    \end{subfigure}
\caption{Agreement of images identified as relevant, i.e., malignant (attention score $>0.25$ or image probability $>0.5$ for IS models), for truly malignant cases. Our proposed model \ESAttSide generally outperforms other models in finding the malignant images.}
\label{fig:res:attmodel-imglabel-f1}
\end{figure}

Fig.~\ref{fig:res:attmodel-imglabel-f1} shows that for \vindr, our side-wise pooling models \ESAttSide and \ISAttSide have better capability to find the malignant images in malignant cases using the image labels from attention scores (F1$=0.81$) than the other models. 
However, \ISAttSide has a lower F1 score using the image labels from image probabilities than the other IS models. For \mgmannot, our side-wise pooling models \ESAttSide and \ISAttSide have better capability to identify malignant images (F1$=0.76$) when compared across F1 scores from attention scores. F1 score from image probabilities of \ISAttImg is highest overall. Almost all MIL attention models outperform the F1 score from the uniform weights of the baseline model \ISMeanImg, where all images have equal importance. Further, all attention pooling models have better capability of identifying malignant images in malignant case than SIL models for both datasets. Also, \SILil has better F1 score than \SILcl showing that training with image labels which are transferred from case labels results in a more error-prone model. In conclusion, \textbf{our side-wise pooling model has a better capability of identifying the malignant images within a malignant case compared to other MIL and SIL models}.

\subsection{Quantitative Assessment of Unsupervised ROI Extraction}
\label{ssec:expsetup:unsupervised-roi}
\emph{Our goal: How well can an unsupervised ROI extractor (e.g., \gmic) extract the correct ROIs from the images in MIL and SIL models? How does our proposed model \ESAttSide perform in this aspect?} We investigate the quality of the unsupervised ROI extractor \gmic~\citep{shen2021interpretable} quantitatively in this section. 

We calculated the intersection over union (IoU) and dice similarity coefficient (DSC) of the extracted ROI candidates with the ROI groundtruth annotations available in the two public datasets \cbis and \vindr, and an annotated subset of our private dataset \mgmannot (Table~\ref{tab:results:roi-extraction}). 
\begin{table}[thbp]
\centering
\caption{IoU and DSC score for ROI extraction in \cbiscus, \vindr, \mgmannot across all models; \ESAttSide achieves similar performance to SIL.}
\label{tab:results:roi-extraction}
\begin{small}
\begin{tabular}{+lllllll}
\toprule \tabhead
  & \multicolumn{2}{c}{\textbf{\cbiscus}} & \multicolumn{2}{c}{\textbf{VinDr}} & \multicolumn{2}{c}{\textbf{\mgmannot}} \\ 
\textbf{Model}  & \textbf{IoU} & \textbf{DSC} & \textbf{IoU} & \textbf{DSC} &  \textbf{IoU} & \textbf{DSC} \\ \otoprule
\SILil & $0.07 \pm 0.02$ & $\textbf{0.11} \pm 0.04$ & $0.29 \pm 0.02$ & $0.41 \pm 0.03$ & n.a. & n.a.\\
\SILcl & $0.07 \pm 0.07$ & $0.10 \pm 0.09$ & $0.25 \pm 0.02$ & $0.35 \pm 0.02$ & $0.19$ & $0.28$\\
\ISMeanImg & $0.05 \pm 0.05$ & $0.07 \pm 0.07$ & $0.23 \pm 0.03$ & $0.33 \pm 0.04$ & $0.15$ & $0.23$\\  
\ISMaxImg & $0.03 \pm 0.04$ & $0.05 \pm 0.07$ & $0.25 \pm 0.02$ & $0.35 \pm 0.03$ & $0.00$ & $0.01$\\ 
\ISAttImg & $0.02 \pm 0.02$ & $0.04 \pm 0.04$ & $0.22 \pm 0.04$ & $0.30 \pm 0.05$ & $0.17$ & $0.26$ \\ 
\ISGattImg & $0.04 \pm 0.05$ & $0.06 \pm 0.07$ & $0.24 \pm 0.07$ & $0.37 \pm 0.08$ & $0.17$ & $0.27$\\ 
\ISAttSide & $0.03 \pm 0.05$ & $0.05 \pm 0.08$ & $0.28 \pm 0.03$ & $0.40 \pm 0.03$ & $\textbf{0.20}$ & $\textbf{0.30}$\\ 
\ESMeanImg & $0.04 \pm 0.04$ & $0.06 \pm 0.06$ & $0.23 \pm 0.03$ & $0.33 \pm 0.03$ & $0.19$ & $0.29$\\ 
\ESMaxImg & $0.00 \pm 0.00$ & $0.01 \pm 0.01$ & $0.28 \pm 0.04$ & $0.39 \pm 0.05$ & $0.19$ & $0.29$\\ 
\ESAttImg & $0.02 \pm 0.02$ & $0.02 \pm 0.03$ & $0.26 \pm 0.05$ & $0.36 \pm 0.07$ & $0.19$ & $0.29$\\ 
\ESGattImg & $0.01 \pm 0.02$ & $0.02 \pm 0.03$ & $0.29 \pm 0.02$ & $0.40 \pm 0.03$ & $\textbf{0.20}$ & $\textbf{0.30}$\\ 
\ESAttSide & $\textbf{0.08} \pm 0.04$ & $\textbf{0.11} \pm 0.06$ & $\textbf{0.30} \pm 0.03$ & $\textbf{0.42} \pm 0.05$ & $0.19$ & $0.29$\\  \bottomrule
\end{tabular}
\end{small}
\end{table}
We calculated the highest match of the top-6 ROI candidates with any of the abnormalities in one image and report the scores for all MIL and SIL models. Our proposed side-wise pooling model, \ESAttSide, generally achieves higher IoU and DSC score than the other MIL and SIL models, except for \mgmannot, where \ESAttSide is 0.01 points lower than the highest score. Case-level models (MIL) outperform or have similar capability to image-level models (SIL) in  ROI candidate extraction. The low scores in \cbis compared to the other two datasets might be due to the lower image quality in \cbis. 
We further found that the IoU and DSC scores slightly decreases when we evaluate whether the model found all abnormalities in one image (cf. Table \ref{sup:tab:results:mean-roi-extraction}). The scores decrease further when we evaluate whether the highest patch-level attention scores are assigned to the correct ROIs (cf. Table \ref{sup:tab:results:maxattscore-roi-extraction}). 
Thus, \textbf{case-level models have similar capability to extract ROIs as image-level models and our side-wise pooling model \ESAttSide generally outperforms other MIL methods in ROI extraction}. 

\subsection{Qualitative Assessment of Image and ROI Identification} 
\label{ssec:exp:unsupervised-roi:qualitative}
\emph{Our goal: How well can our \ESAttSide model extract ROI candidates and identify the relevant images qualitatively?} We qualitatively evaluated the case-level visualization and the ROI candidates for \mgm by i) anecdotal evidence and ii) through a semi-structured interview with radiologists. For both evaluations, we divided the cases into four groups: true positive (malignant detected as malignant), true negative (benign detected as benign), false positive (benign detected as malignant), and false negative (malignant detected as benign) cases.

\begin{figure}[tbhp]
\centering
\includegraphics[scale=0.28]{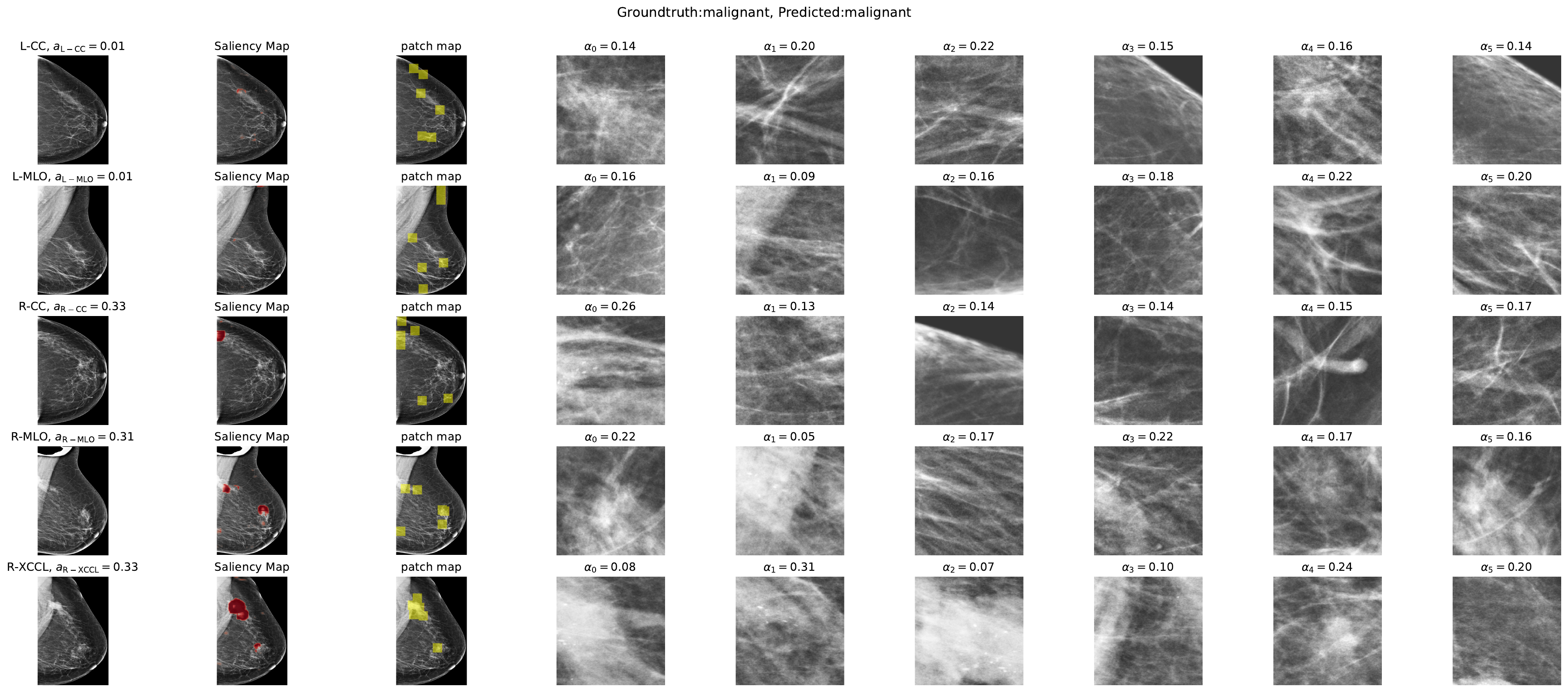}
\caption{Output visualization of the malignant mammogram case from Fig.~\ref{fig:intro:case-5image-mgm-example} by our \ESAttSide model. Views from right breast are flipped horizontally for input to the model.
Each row shows the original mammogram image, the saliency map from the GMIC Global Network (important regions shown in red), extracted patches marked (in yellow) on the image and then the 6 ROI candidates corresponding to the extracted patches. The associated scores of the images (e.g., $a_{\mathrm{L-CC}}$) and patches (e.g., $\alpha_0$) are the attention scores assigned by the image- and patch-level MIL pooling in our model respectively. Our side-wise MIL pooling assigned highest attention scores to the three malignant images - R-CC, R-MLO and R-XCCL. The malignant mass in the right breast is successfully extracted by the model (cf. Fig.~\ref{fig:intro:case-5image-mgm-example} for groundtruth annotation), i.e., second patch in R-MLO image (4\textsuperscript{th} row) and the third patch in R-XCCL image (5\textsuperscript{th} row). However, these patches are not assigned the highest patch-level attention score. These patches are also highlighted in red in the saliency map of the images.}
\label{fig:mammogram-case-patch-vis}
\end{figure}

\textit{Anecdotal evidence:} Fig.~\ref{fig:mammogram-case-patch-vis} shows an example of a true positive case from \ESAttSide for \mgm. Our model extracted the ROIs correctly and also provided higher attention scores to the right breast images containing the abnormality. 
We further show one example from each group from \ESAttSide in the supplementary material - true negative (cf. Fig. \ref{sup:fig:case-5image-tn}), false positive (cf. Fig. \ref{sup:fig:abnormality-6imagecase-fp}), and false negative (cf. Fig. \ref{sup:fig:abnormality-oneview-fn}). For all examples, all models extracted the correct ROIs, though not all diagnoses were correct. 
 
\textit{Evaluation with radiologists}: We qualitatively evaluated the top-6 ROI candidates for 61 cases from the \mgm test set through a semi-structured interview with two radiologists.
We asked the following questions during the interview. 
For true positive and false negative cases: ``Do you see any relevant ROI among the extracted ROI candidates?'' and   ``Does the relevant ROI have the highest attention score?''. For true negative cases we asked ``What kind of patches do you see?''.  For misclassifications, we asked the radiologists to describe which kind of mistakes were made.
Clinicians found the extracted ROIs to be relevant, and confirmed that the patches with abnormalities were extracted for the malignant cases. Anecdotally, radiologists were surprised that the model found the mass abnormality for one of the cases with high breast density, which is usually rather hard to detect for humans. 
While important ROIs were correctly extracted, they were not always associated with the highest attention score.
However, highest image-level attention scores were more often correctly associated with the important images for the decision. For true negative cases, extracted ROIs usually showed normal tissues or benign abnormalities.
For the malignant and benign misclassifications, the radiologists commented that the correct ROIs were extracted, but that the decision on malignant or benign would require additional diagnostic tests. Thus, \textbf{qualitative assessment of case-level visualization shows satisfying quality of extracted ROIs and identification of relevant images, though there is potential for improvement.}

\subsection{ROI Extraction vs Diagnosis at the Case Level}
\label{ssec:exp:right-for-right-reason}
\emph{Our goal: How many of the cases have the correct diagnosis for the right reasons, i.e., the extraction of the correct ROI?} We analyzed for how many cases the model made the correct prediction and extracted the correct ROI (a step towards ``right for the right reasons''). We calculated confusion matrices with true and predicted case-level labels and an additional check whether the correct ROI was extracted. 

\begin{figure}[htbp]
\centering
\begin{subfigure}[b]{0.5\textwidth}
        \centering
        \includegraphics[scale=0.65]{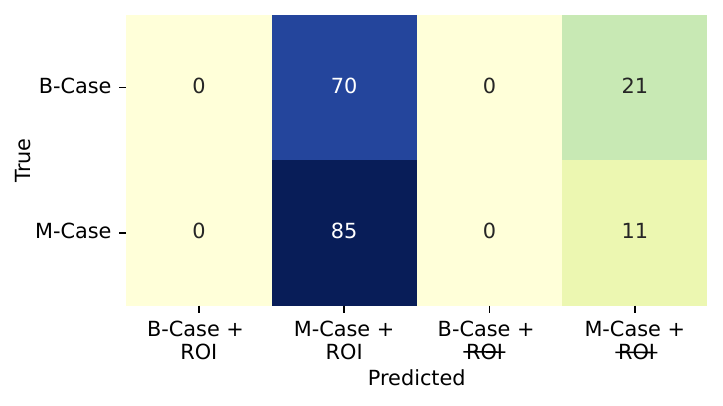}
        \caption{\ISMeanImg}
        \label{fig:conf-mat-roi-diagnosis-vindr:ismeanimg}
    \end{subfigure}%
    \begin{subfigure}[b]{0.5\textwidth}
        \centering
        \includegraphics[scale=0.65]{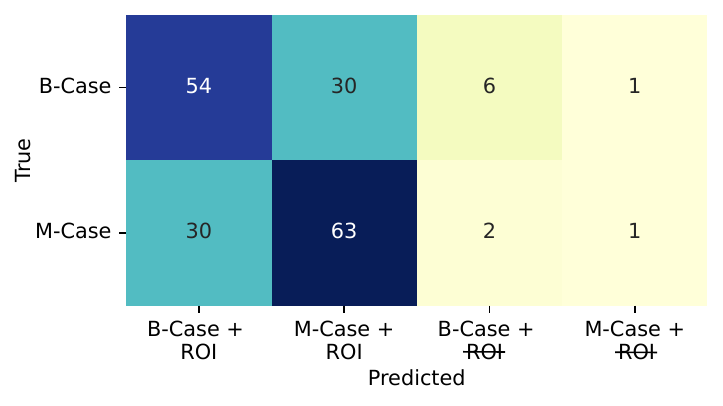}
        \caption{\ESAttSide}
        \label{fig:conf-mat-roi-diagnosis-vindr:esattside}
    \end{subfigure}
\caption{Confusion matrix for benign and malignant cases coupled with a check on ROI extraction for \ISMeanImg and \ESAttSide models on the test set of \vindr. \{B,M\}-Case on the y-axis denotes the number of true benign and malignant cases. \{B,M\}-Case + ROI on the x-axis denotes the number of cases predicted as benign or malignant with correct extraction of the ROI, while \{B,M\}-Case + \sout{ROI} denotes the number of cases where the correct ROI was not extracted. Majority of the cases were correctly classified with the correct ROI extracted by \ESAttSide.}
\label{fig:conf-mat-roi-diagnosis-vindr}
\end{figure}

In Fig.~\ref{fig:conf-mat-roi-diagnosis-vindr}, we compare the confusion matrix of our proposed side-wise pooling model \ESAttSide (Fig.~\ref{fig:conf-mat-roi-diagnosis-vindr:esattside}) with the baseline MIL pooling model \ISMeanImg (Fig.~\ref{fig:conf-mat-roi-diagnosis-vindr:ismeanimg}) on all cases in the test set of \vindr that contain an abnormality. 
For \ESAttSide, we found that 54 benign cases were correctly predicted as benign along with correct ROI extraction (B-Case + ROI), and 63 malignant cases were correctly predicted as malignant with correct ROI extraction (M-Case + ROI). 30 benign and 30 malignant cases were misclassified as the other class though the correct ROI was extracted. There are 10 cases where the correct ROI was not in the top-6 ROI candidates extracted by the model (+ \sout{ROI}). Those cases were mostly classified as benign. \ESAttSide has a much greater number of cases where the correct ROI was extracted than \ISMeanImg. In \ISMeanImg, all benign and malignant cases with correctly extracted ROI and cases where the correct ROI was not extracted were classified as malignant. \textbf{The majority of the cases were correctly classified with the correct ROI extracted by our \ESAttSide model. However, some of the benign and malignant cases were misclassified as the other class even when the correct ROI was found}. Further, a case can be correctly diagnosed without the model finding the right reason (i.e., the correct ROI), albeit this constitutes a small portion of the whole set.

\subsection{Generality of Side-wise MIL Pooling Block}
\label{ssec:exp:feature-extractor-agnostic}
\emph{Our goal: How well does our side-wise MIL pooling block perform across various feature extractors? And is two-level MIL better than one-level MIL for breast cancer prediction?} We study the generality of our side-wise MIL pooling block \ESAttSide across three feature extractors: ResNet34~\citep{shen2021interpretable}, RGP~\citep{shu2020deep}\footnote{The image size for RGP is 800x800 whereas for \rnIIIIV and \gmic is 2944x1920 taken from the original work. RGP did not fit in 1 GPU on increasing the image size as the base feature extractor is DenseNet169, so we kept the image size unchanged.} and \gmic~\citep{shen2021interpretable} by comparing its performance with the baseline MIL pooling \ISMeanImg on \cbiscus and \mgmfv. Here, RGP denotes DenseNet169 combined with RGP pooling from~\citet{shu2020deep}. Further, we also compared the models across two-level MIL (image- and patch-level) and one-level MIL (image-level only). 

Table~\ref{tab:results:feature-extractor-agnostic} shows that the side-wise MIL pooling block of \ESAttSide improves performance over simple image-level score aggregation \ISMeanImg for all feature extractors, ResNet34, RGP and \gmic on \cbiscus. 
\begin{table}[thbp]
\centering
\caption{Impact of our domain-specific side-wise pooling block across three feature extractors, ResNet34, DenseNet169 + RGP (RGP) and \gmic on \cbiscus and \mgmfv. MIL at patch level (P) and/or image level (I): \ding{55} - absence, \checkmark - presence. Side-wise pooling block improves performance generally across all features extractors showing the generality of our pooling method.}
\label{tab:results:feature-extractor-agnostic}
\begin{small}
\begin{tabular}{+lllll^c^c}
\toprule \tabhead
 & \multicolumn{2}{c}{\textbf{MIL}} & \textbf{Feat.} & \textbf{Image} & \textbf{F1} & \textbf{AUC} \\ 
 & \textbf{P} & \textbf{I} & \textbf{Ext.} & \textbf{Aggr.} &  &  \\ \otoprule
\multirow{6}{*}{\rotatebox{90}{\cbiscus}}  & \ding{55} & \checkmark & ResNet34  & \ISMeanImg & $0.60 \pm 0.08$ & $0.69 \pm 0.10$ \\
 & \ding{55} & \checkmark & ResNet34  & \ESAttSide &  $0.63 \pm 0.09$ & $0.75 \pm 0.05$ \\
 & \ding{55} & \checkmark & RGP  & \ISMeanImg & $0.67 \pm 0.01$ & $0.74 \pm 0.04$ \\
 & \ding{55} & \checkmark & RGP  & \ESAttSide & $0.66 \pm 0.03$ & $0.77 \pm 0.02$ \\ 
& \checkmark & \checkmark & GMIC  & \ISMeanImg & $0.64 \pm 0.04$ & $0.77 \pm 0.02$\\
& \checkmark & \checkmark & GMIC & \ESAttSide & $\textbf{0.70} \pm 0.01$ & $\textbf{0.78} \pm 0.02$   \\ \midrule
\multirow{6}{*}{\rotatebox{90}{\mgmfv}} & \ding{55} & \checkmark & ResNet34  & \ISMeanImg & $0.44 \pm 0.03$ & $0.75 \pm 0.03$ \\
 & \ding{55} & \checkmark & ResNet34  & \ESAttSide & $0.42 \pm 0.08$ & $0.74 \pm 0.07$ \\
 &  \ding{55} & \checkmark & RGP  & \ISMeanImg & $0.52 \pm 0.01$ & $0.81 \pm 0.02$ \\
 & \ding{55} & \checkmark & RGP  & \ESAttSide & $\textbf{0.59} \pm 0.01$ & $\textbf{0.86} \pm 0.00$ \\
& \checkmark & \checkmark & GMIC  & \ISMeanImg & $0.46 \pm 0.02$ & $0.76 \pm 0.03$\\
& \checkmark & \checkmark & GMIC & \ESAttSide & $\textbf{0.59} \pm 0.00$ & $0.85 \pm 0.01$   \\ [3pt]
\bottomrule
\end{tabular}
\end{small}
\end{table}
On \mgmfv, \ESAttSide is better than \ISMeanImg for RGP and \gmic, and slightly worse for \rnIIIIV (with a high standard deviation). 
RGP generally performs better than \rnIIIIV and GMIC for \ISMeanImg.
However, adding \ESAttSide to \gmic makes it competitive to RGP or even better. This \textbf{shows the generality and advantage of our side-wise MIL pooling for different feature extractors}.

On comparing one-level with two-level MIL, \gmic + \ESAttSide (two-level) outperforms ResNet34 + \ESAttSide (one-level) on both datasets. However, on \mgmfv, RGP + \ESAttSide (one-level) has similar performance to \gmic + \ESAttSide (two-level). The similar performance of RGP is attributable to the fact that it can select important features from an image (making it intuitively similar to unsupervised ROI extraction). In conclusion, \textbf{two-level MIL (feature selection for each image and weighing images properly in a case) is better than one-level MIL (only weighing images properly in a case).}    

\subsection{Comparison to State-of-the-Art Models}
\label{ssec:exp:comparison-to-sota}
We compared our side-wise MIL pooling model \ESAttSide to our reproduction of two SoTA case-level models, {\lmss DMV-CNN}~\citep{Wu_2020} and \kimrn~\citep{kim2018applying}, and three SoTA image-level models, {\lmss \rnIIIIV}, {\lmss RGP} and {\lmss GMIC}, on \cbis, \vindr and \mgmfv. We used the view-wise feature concatenation version for {\lmss DMV-CNN} without any application of BI-RADS pretraining and ROI heatmap (due to their unavailability for our \mgmfv dataset).
Table~\ref{tab:results:sota-comparison} shows that \textbf{our side-wise MIL pooling model \ESAttSide outperforms all other models on all datasets in F1 score}.
{\lmss DMV-CNN} requires four fixed views and cannot be trained on \cbis, where cases with four images are a minority (Table~\ref{tab:dataset:mammograms} case dist.), indicated by  ``n.a.'' in the result table.
\begin{table}[thbp]
\centering
\caption{Comparison of our proposed model \ESAttSide to SoTA breast cancer models (our implementation). Level indicates level at which training and prediction was done, Paradigm (Par.) and aggregation method of images in a case. \ESAttSide has higher F1 score than other models. Batch sizes (bs) \cbis: bs=3, \vindr: bs=7, and \mgmfv: bs=5.*models trained from scratch.}
\label{tab:results:sota-comparison}
\resizebox{\textwidth}{!}{
\begin{tabular}{++lll^c^c^c^c^c^c^c^c^c^c}
\toprule \tabhead
\textbf{Model} & \textbf{Level} & \textbf{Par.} & \textbf{Aggr.} & \multicolumn{2}{c}{\textbf{\cbiscus} (\textbf{\cbisoff})} & \multicolumn{2}{c}{\textbf{VinDr}} & \multicolumn{2}{c}{\textbf{\mgmfv}}\\ 
 & & & & \textbf{F1} & \textbf{AUC} & \textbf{F1} & \textbf{AUC} & \textbf{F1} & \textbf{AUC} \\ \otoprule
{\lmss ResNet34}~\citep{shen2021interpretable} & Image & n.a. & n.a. & $0.65 \pm 0.02$ & $0.76 \pm 0.02$ & $0.30 \pm 0.02$ &  $0.85 \pm 0.01$ & n.a & n.a \\
 &  &  &  & ($0.63 \pm 0.03$) & ($0.76 \pm 0.01$) &  &   & &  \\
{\lmss RGP}~\citep{shu2020deep} & Image & n.a. & n.a. & $0.67 \pm 0.01$ & $0.78 \pm 0.02$ & $0.32 \pm 0.00$ & $\textbf{0.86} \pm 0.02$  & n.a & n.a \\
 &  &  &  & ($0.62 \pm 0.01$) & ($0.76 \pm 0.01$) &  &   &  &  \\
{\lmss GMIC}~\citep{shen2021interpretable} & Image & n.a. & n.a. & $0.66 \pm 0.03$ & $\textbf{0.79} \pm 0.05$ & $0.27 \pm 0.01$ & $0.83 \pm 0.01$ & n.a & n.a \\
 &  &  &  & ($0.64 \pm 0.02$) & ($0.77 \pm 0.03$) &  &  &  &  \\
{\lmss DIB-MG*}~\citep{kim2018applying} & Case & IS & Mean & $0.52 \pm 0.03$ & $0.64 \pm 0.01$ & $0.32 \pm 0.03$ & $0.68 \pm 0.02$ & $0.32 \pm 0.00$ & $0.66 \pm 0.02$ \\
 &  &  &  & ($0.56 \pm 0.05$) & ($0.64 \pm 0.03$) &  &  &  &  \\
{\lmss DMV-CNN*}~\citep{Wu_2020} & Case & ES & Concat & n.a. & n.a. & $0.27 \pm 0.01$ & $0.75 \pm 0.04$ & $0.36 \pm 0.00$ & $0.68 \pm 0.01$ \\ 
\ESAttSide & Case & ES & Side Att. & $\textbf{0.70} \pm 0.01$ & $0.78 \pm 0.02$ & $\textbf{0.48} \pm 0.03$ & $0.83 \pm 0.02$ & $\textbf{0.59} \pm 0.00$ & $\textbf{0.85} \pm 0.01$ \\
 &  &  &  & ($0.67 \pm 0.03$) & ($0.77 \pm 0.01$) &  &  &  &  \\
\bottomrule
\end{tabular}}
\end{table}

\subsection{Training with Variable-image Cases} 
\label{ssec:exp:variable-images}
\emph{Our goal: How well does dynamic training perform compared to default training for training models on variable-image cases? How well do models trained on the four images of standard views perform on variable-image cases including additional views unseen during training compared to models trained on variable-image cases?} 

\begin{figure}[htbp]
\centering
    \begin{subfigure}[b]{0.36\textwidth}
        \centering
        \includegraphics[scale=0.18]{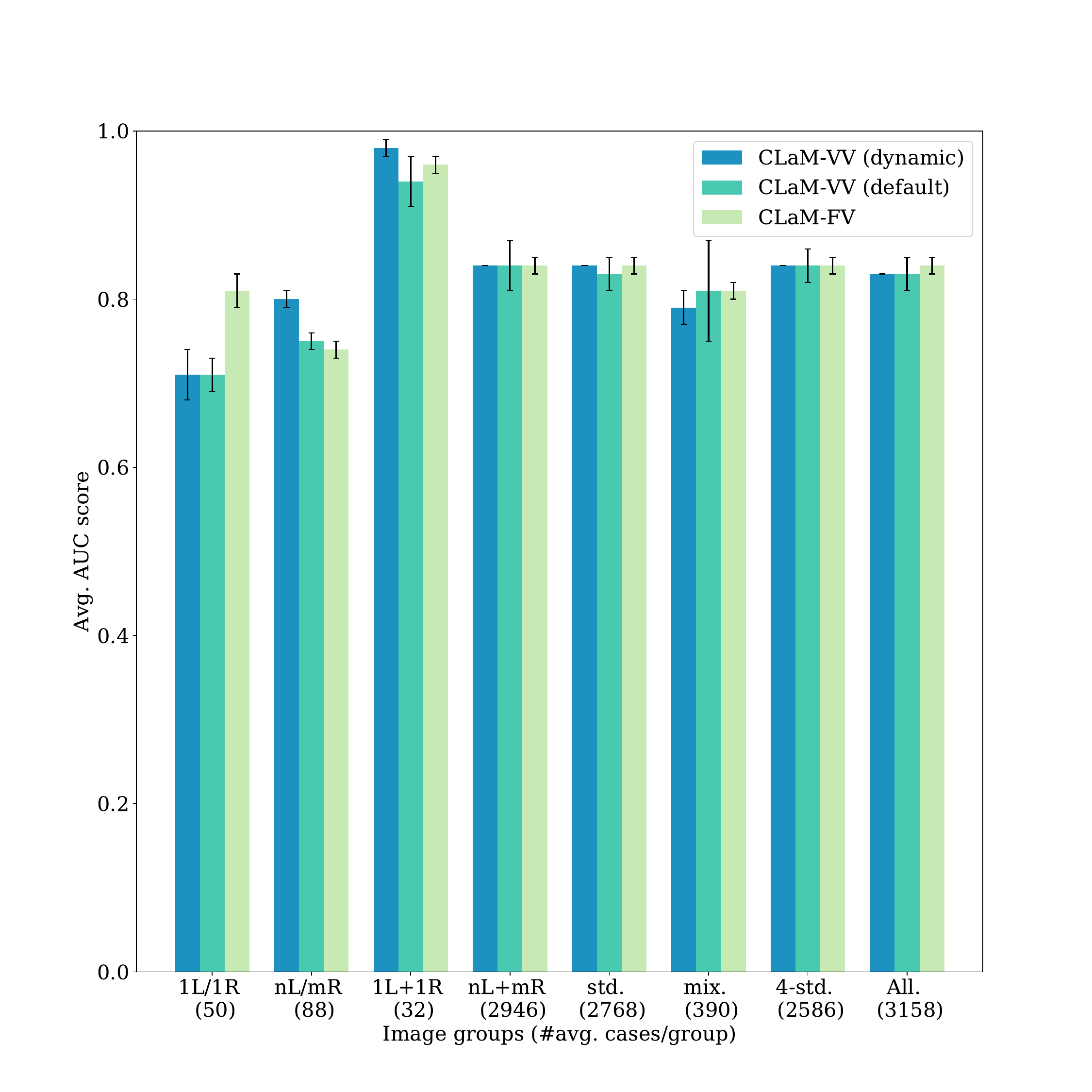}
        \caption{\ESAttSide on \mgmvv}
        \label{fig:res:variable-images:attside-mgm}
    \end{subfigure}%
    \begin{subfigure}[b]{0.32\textwidth}
        \centering
        \includegraphics[scale=0.17]{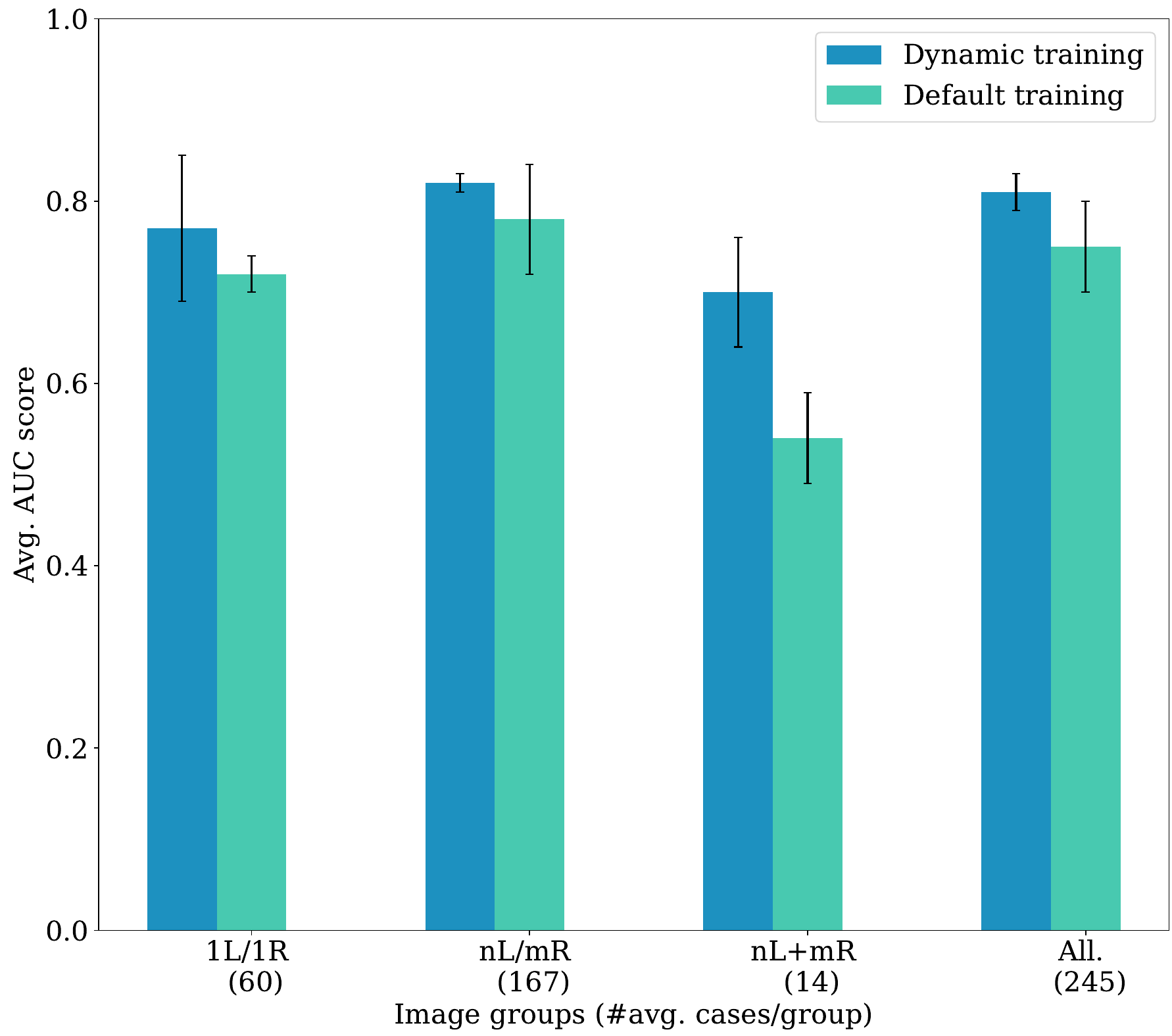}
        \caption{\ESAttImg on \cbiscus}
        \label{fig:res:variable-images:attimg-cbis}
    \end{subfigure}%
    \begin{subfigure}[b]{0.32\textwidth}
        \centering
        \includegraphics[scale=0.17]{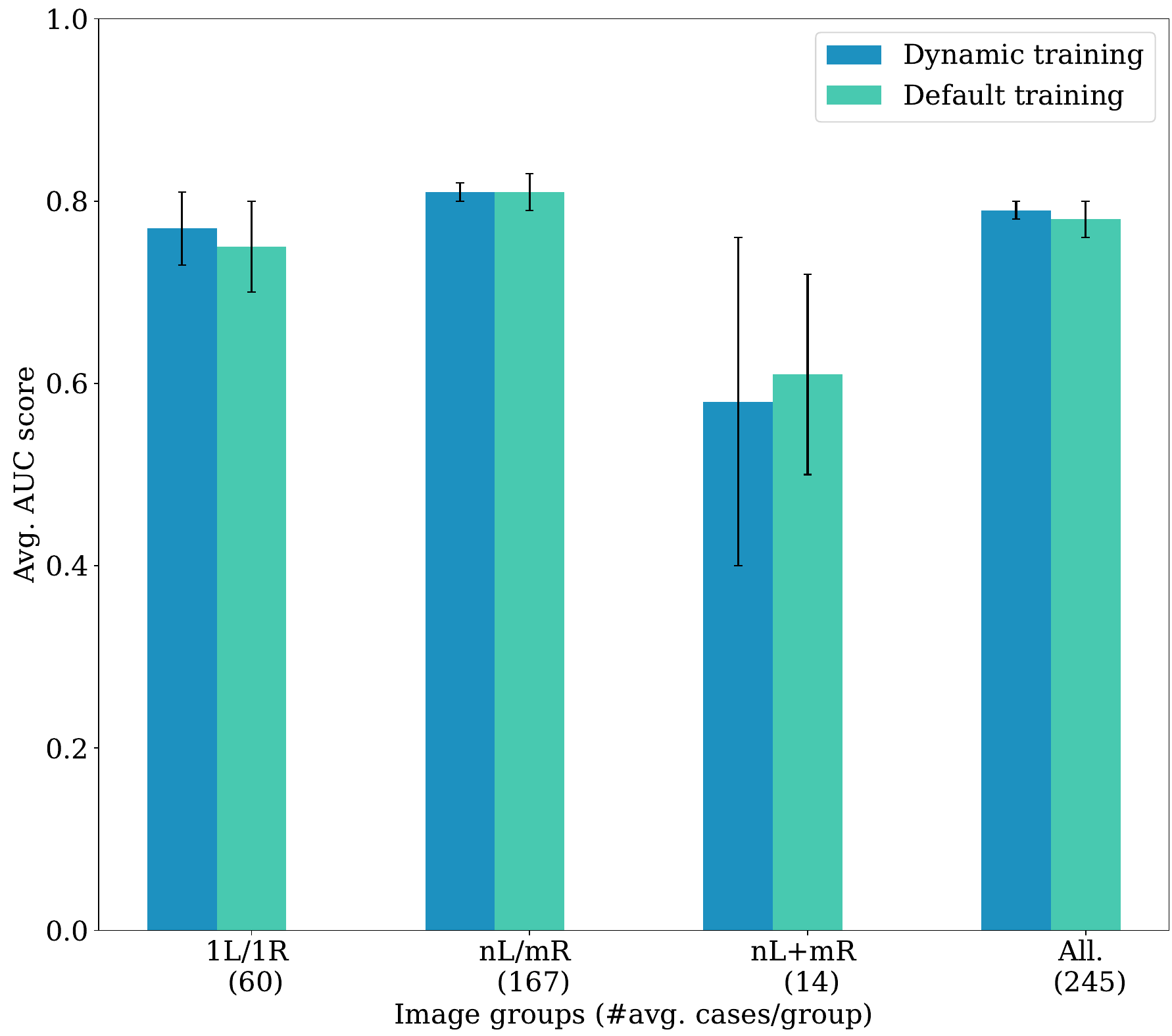}
        \caption{\ESAttSide on \cbiscus}
        \label{fig:res:variable-images:attside-cbis}
    \end{subfigure}
\caption{Dynamic and default training performance of (a) \ESAttSide on \mgmvv, (b) \ESAttImg and (c) \ESAttSide on \cbiscus, grouped by different types of cases. (a) includes fixed-view training performance. y-axis shows AUC score with standard deviation (error bar). x-axis shows groups of cases, e.g., 1L/1R denotes cases containing only a single image from either side of the breast (left (L) or right (R)).}
\label{fig:res:variable-images}
\end{figure}

\textit{Dynamic vs default training:} 
We compared variable-image training with dynamic vs default scheme on \mgmvv and \cbiscus. 
Fig.~\ref{fig:res:variable-images} shows the comparison of the two schemes with \ESAttSide on \mgmvv and \cbiscus, and \ESAttImg on \cbiscus.
We report performance across various groups of cases: cases with a single image from any breast (1L/1R), cases with images from 1 breast (nL/mR), one image from both breasts (1L+1R), multiple images from both breasts (nL+mR), any number of images of standard views (std.), 4 images of standard views (4-std), cases with at least one additional view (mix.), and all cases (All).
On \cbis, dynamic training outperforms default training in all groups for \ESAttImg (Fig.~\ref{fig:res:variable-images:attimg-cbis}) and in the 1L/1R and All group for \ESAttSide while being on par in the nL/mR group (Fig.~\ref{fig:res:variable-images:attside-cbis}).
On \mgmvv, dynamic and default training perform similar for the three majority groups nL+mR, 4-Std and All, while dynamic training outperforms default training in the two minority groups nL/mR and 1L+1R (Fig. \ref{fig:res:variable-images:attside-mgm}). Default training outperforms dynamic training in the group mix, but has high standard deviation. For most groups, we observe default training has generally higher standard deviation than dynamic training on \mgm.
Overall dynamic training is slightly better than default training with performance gains seen in particular in some minority groups. 

\textit{Fixed-image vs variable-image training:}
We compared training \ESAttSide on variable-image cases (\mgmvv, variable amount of images per case) with training on fixed-image cases (\mgmfv, always four images of standard views). Both variants were trained for max. 30 epochs with early stopping (patience epoch 10) and tested on \mgmvv.
The results in Fig.\ref{fig:res:variable-images:attside-mgm} show that fixed-image training slightly outperforms variable-image training (group All), and that both variants are roughly on par for the 4-std and mix group, despite that the latter contains additional views unseen by fixed-image training.
The performance gains of fixed-image training are largely driven by the better performance on single images (group 1L/1R), while variable-image training performs better on two minority groups (images only from one side, nL/mR, and one image from each side, 1L+1R).

Overall, we conclude that it is not necessary to train with variable-image cases to obtain reasonable performance on variable-image cases (including unseen views) during prediction when there are enough fixed-image cases. \textbf{Models trained on fixed-image cases work good enough for inference on variable-image cases, given that the fixed-image models is inherently able to handle variable-image cases (like our model). When fixed-image cases are the minority, our dynamic training is preferable.} We emphasise that models developed for real hospital settings should always be able to handle variable-image cases and work well on them.  

\section{Discussion}
\textbf{Our proposed case-level model.} Our results show that our proposed case-level model, \ESAttSide is better than or on par with image-level models, eliminating the need for manual annotation of images. This makes it possible to train breast cancer prediction models with only case-level labels, i.e., the diagnosis, that is readily available at the hospital, thus enabling model training in real hospital settings. 
We performed an extensive analysis of the factors that influence the high performance of our side-wise pooling block in embedded-space. Specifically, we investigated two aspects, the ability to find relevant images within a case, and the ability to extract ROIs in an unsupervised manner. Our results show that our side-wise pooling model is better  at both, finding the important images and extracting ROIs, and this might contribute to the high classification performance of \ESAttSide. 
We conjecture that our case-level framework is applicable to other medical examinations where multiple views are taken and only the final diagnosis is provided at the end, and that our side-wise pooling can be applied to scenarios with more than one entity and multiple views per entity.

\textbf{Comparison to state-of-the-art.} Our case-level model, \ESAttSide outperforms our implementation of the SoTA models both at the image and the case level. On the official split of \cbis, the performance of \ESAttSide (AUC\footnote{Most related work only reports AUC. However, the score is misleading for imbalanced datasets, as our results on VinDr show (cf. Table~\ref{tab:results:mil-pooling}).}=0.77) is slightly below the performance reported in related work for an image-level model~\citep{wei2022beyond} (AUC=0.80 without ensemble and test time augmentation) and for a model trained with ROI annotation~\citep{petrini2022breast} (AUC=0.84 on \cbis cases with only 2 views, hence not fully comparable).
While~\citet{quellec2016multiple} considered abnormalities that are not visible in both views as false alarms, we take a different approach. As seen in Fig.~\ref{fig:intro:case-5image-mgm-example}, an abnormality occurring near the pectoral muscle can not be seen in the CC view, but in the MLO or additional views. Our side-wise MIL pooling block can learn to select the important view for the task without ignoring abnormalities that are only visible in one view. 
Our overlap score (DSC=0.42 on \vindr) between ROIs extracted by the model and the groundtruth ROIs is higher than scores reported in related work~\citep{shen2021interpretable} (DSC=0.33) and~\citep{liu2021weakly} (DSC=0.39) on their private dataset. 

\textbf{Performance across datasets.} We also observe differences in performance across datasets. Overall, the performance varies from F1 $=0.70$ on \cbis, F1 $=0.59$ on \mgmfv to F1 $=0.48$ on \vindr.  
Aside from varying data quantity and quality, the difficulty of the classification task can also impact performance.~\citet{khan2019multi} found normal vs. abnormal (benign and malignant) to be easier to classify (AUC $= 0.93$) than benign vs. malignant (AUC $=0.84$) on MIAS-CBIS. Similarly, in our datasets, the benign class of \cbis has only benign cases (i.e., cases that contain benign abnormalities), whereas in \mgm it additionally contains normal cases.
For \vindr, we followed \citet{carneiro2015unregistered} in using BI-RADS scores as a proxy to assign malignant and benign labels, but this may not reflect the true class of the cases (with a higher chance of some benign cases getting the groundtruth of malignant). We suspect that the lower performance of our approach on \vindr is rooted in this likely inaccuracy.

\textbf{Reproducibility and replicability.} Reproducibility and replicability are important aspects of scientific research for scientific progress~\citep{Ulmer2022}. We found a gap of 0.03-0.08 in AUC on \cbisoff while reproducing related work~\citep{shu2020deep,shen2021interpretable}. We call for transparency on hyperparameter settings and the preprocessing details to support reproducibility. We have made our hyperparameters, preprocessing, and training code public\textsuperscript{\ref{fn:repository}} to promote reproducibility and to set our work as a benchmark. 

\textbf{Main factors for high performance of case-level breast cancer prediction model.} The main factors are pretraining, unsupervised ROI extraction, choice of MIL pooling, and hyperparameter tuning.
First, we found that pretrained models (trained on ImageNet) increase the performance by at least 10\%. We have not experimentally verified the claim of achieving higher performance by pretraining models on BI-RADS scores~\citep{Wu_2020,shen2021interpretable}, because BI-RADS scores  are not readily available in hospital databases in a structured manner. 
The performance range of our models pretrained on ImageNet (AUC=0.85 \mgm) is similar to models pretrained on BI-RADS (AUC=0.82 view-wise model~\citep{Wu_2020} on private dataset), showing that both types of pretraining are beneficial.
Second, unsupervised ROI extraction methods (e.g., \gmic) or feature selection methods (e.g., RGP) improve performance over non-ROI feature extractors. We observed a 0.17 improvement in F1 score on replacing \rnIIIIV with \gmic or RGP for \mgmfv.  
Third, the choice of MIL pooling can impact the F1 score by up to 0.13 (\mgmfv).
Fourth, we found hyperparameter tuning to be important to get good performance. For instance, the AUC score varied from 0.78 to 0.82 for \gmic on the official \cbis split among the top-5 settings for model hyperparameters in SIL.\footnote{We report average over these top-5 settings in Table~\ref{tab:results:feature-extractor}, while for other tables, we report average over 3 seeds with the best hyperparameter.}

\textbf{Limitations and future work.} We found that highest attention scores were not assigned to the correct ROI candidate in the patch-level MIL pooling. In the future, it might be interesting to investigate whether solving this can further improve performance and align the model towards right for the right reasons. Unsupervised ROI extraction does a good job at finding ROIs but it can be further improved to find ROIs in all cases.  Our dynamic training approach performs better than the default training, however, it does not improve performance over fixed-image training for single image group in \mgm. We think that the learning dynamics of a model can depend on multiple things, e.g. the order of the batches and it may be interesting to investigate training of variable-instance bags in the future. 
Further, in this work, we assume that instances in a bag are conditionally independent given the bag (label). 
However, this assumption is unlikely to hold, as an abnormality visible in one view may also be visible in other views. In future work, we plan to investigate how views relate to each other, e.g., by cross-view attention~\citep{van2021multi,manigrasso2024mammography}, graph convolutional networks~\citep{liu2021act,manigrasso2024mammography} or relational networks~\citep{yang2020momminet,yang2021momminet}. Some of these approaches currently rely on ROI information during training and we plan to investigate whether and how they can be lifted to work with case-level labels only.

In the clinical workflow, a mammography exam alone is not always enough for diagnosis. This may be reflected in misclassifications found in our model. For some, the model was not able to correctly predict the class, even though it extracted the correct ROI. 
For example, in Fig. \ref{sup:fig:case-5image-tn}, the model classified the benign case containing calcifications as benign, but in Fig. \ref{sup:fig:abnormality-6imagecase-fp}, the model misclassified another benign case containing calcification. 
We plan to investigate the reason behind this observation - is it a mistake of our model or is mammography alone not sufficient to classify this case? Similarly, in Fig. \ref{sup:fig:abnormality-oneview-fn} the mass abnormality is similar in appearance to benign mass and it might be one of the reasons why the model misclassified this case as benign even though the ROI was extracted correctly. However, similarity in appearance to a certain class does not alone define whether the abnormality is benign or malignant. Previous mammogram history, age, position and size of the tumor are some of the other factors determining the diagnosis of the tumor. 
It is crucial to be aware of these limitations while predicting breast cancer from mammography alone.  

\section{Conclusion} 
We propose a framework for case-level breast cancer prediction using mammography that addresses the three challenges of real hospital settings - groundtruth only available at the case level, no ROI annotation available and variable number of images per case. Specifically, we propose a two-level MIL pooling approach at the patch and image level that can be trained end-to-end with only weak (case-level) labels. 
Our case-level model achieves similar performance to image-level models without the need of time-consuming manual annotation of images, can identify the malignant images in a malignant case and can extract relevant ROIs. 
Thus, our MIL model can reliably point out where an abnormality is, i.e., the breast side, image view and region of interest, which is crucially important for uptake in a clinical workflow.  
While our model is right for the right reasons in the majority of correct classifications, we also observed misclassifications despite that the model extracted the correct ROI. We plan to investigate whether these are truly mistakes of the model or whether they are (partly) due to the fact that also in the clinical workflow, a mammograpy exam alone is not always sufficient for diagnosis, but requires consideration of additional factors (such as history and demographics of the patient). Regardless of whether misclassifications of our model are attributed to these factors, future work should take them into account and focus on multimodal learning. Similarly, since data scales with patient intake, while manual annotation efforts do not, future work should focus on readily available label cues and unsupervised extraction of regions of interest. 

\section*{Declaration of Competing Interest}
The authors declare that they have no known competing financial interests or personal relationships that could have appeared to influence the work reported in this paper.  

\section*{CRediT authorship contribution statement}  
\textbf{Shreyasi Pathak:} Conceptualization, Methodology, Software, Investigation, Data Curation, Writing - Original Draft, Writing - Review \& Editing, Visualization. \textbf{J\"org Schl\"otterer:} Conceptualization, Methodology, Writing - Review \& Editing, Supervision. \textbf{Jeroen Geerdink:} Data Curation, Funding acquisition, Writing - Review \& Editing. \textbf{Jeroen Veltman:} Data Curation, Funding acquisition, Writing - Review \& Editing. \textbf{Maurice van Keulen:} Conceptualization, Writing - Review \& Editing, Supervision, Funding acquisition. \textbf{Nicola Strisciuglio:} Conceptualization, Writing - Review \& Editing, Supervision. \textbf{Christin Seifert:} Conceptualization, Methodology, Writing - Review \& Editing, Visualization, Supervision, Funding acquisition.

\section*{Acknowledgments} The authors would like to thank radiologists Onno Vijlbrief, Rob Bourez and Maikel Viskaal from Hospital Group Twente, The Netherlands for helping with domain knowledge, annotations and for the semi-structured interview of the case-level visualizations from the model. We also gratefully acknowledge Institute of Artificial Intelligence in Medicine (IKIM), Essen, Germany for providing computational resources for running the experiments, and special thanks to Prof. Folker Meyer from IKIM for his support during this process. This work is supported in part by grants from Hospital Group Twente, The Netherlands and by an unrestricted research grant from the Pioneers in Health Care (PIHC) Innovation Fund, established by the University of Twente, Saxion University of Applied Sciences, Medisch Spectrum Twente, ZiekenhuisGroep Twente and Deventer Hospital.

\bibliographystyle{elsarticle-harv} 
\bibliography{main}

\newpage

\appendix

\section{Data preprocessing}
Preprocessing code available in our repository\footnote{Open source framework available at \url{https://github.com/ShreyasiPathak/multiinstance-learning-mammography}}
\begin{figure}[htb!p]
\centering
    \begin{subfigure}[b]{0.50\textwidth}
        \centering
        \includegraphics[scale=0.85]{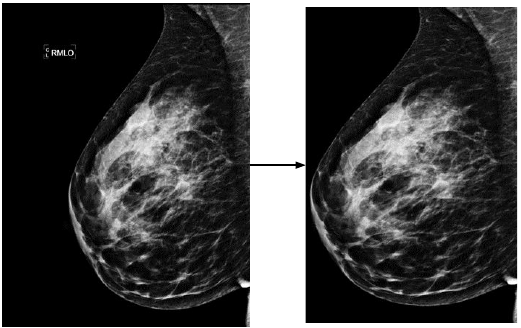}
        \caption{R-MLO}
    \end{subfigure}%
    \begin{subfigure}[b]{0.50\textwidth}
        \centering
        \includegraphics[scale=0.85]{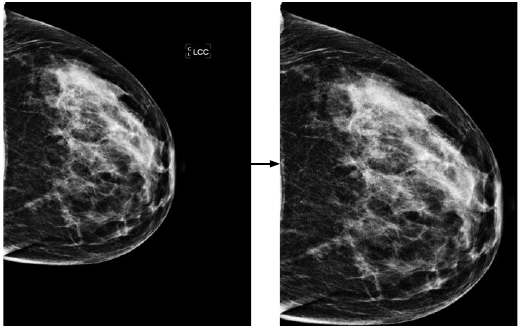}
        \caption{L-CC}
    \end{subfigure}
\caption{Data preprocessing: Our automated algorithm removes the extra black background and irrelevant information from the mammography images.}
\label{sup:fig:data:preprocessing}
\end{figure}

\section{Case-level Visualization}

\subsection{Case 2: Benign abnormality only seen in the right breast and case contains 5 images; True negative prediction (benign predicted as benign)}

\begin{figure}[thbp]
\centering
    \begin{subfigure}[b]{0.21\textwidth}
        \includegraphics[scale=0.21]{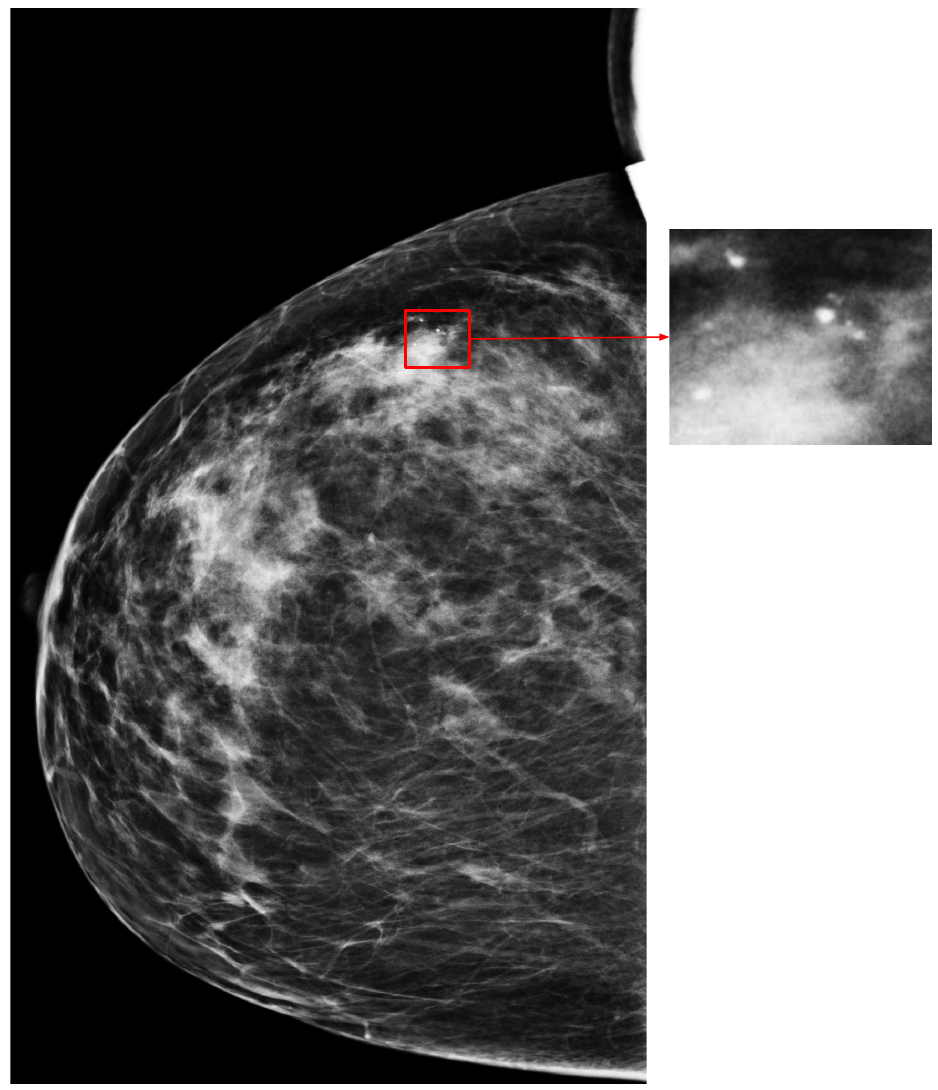}
        \caption{R-CC}
    \end{subfigure}%
    \begin{subfigure}[b]{0.23\textwidth}
        \includegraphics[scale=0.21]{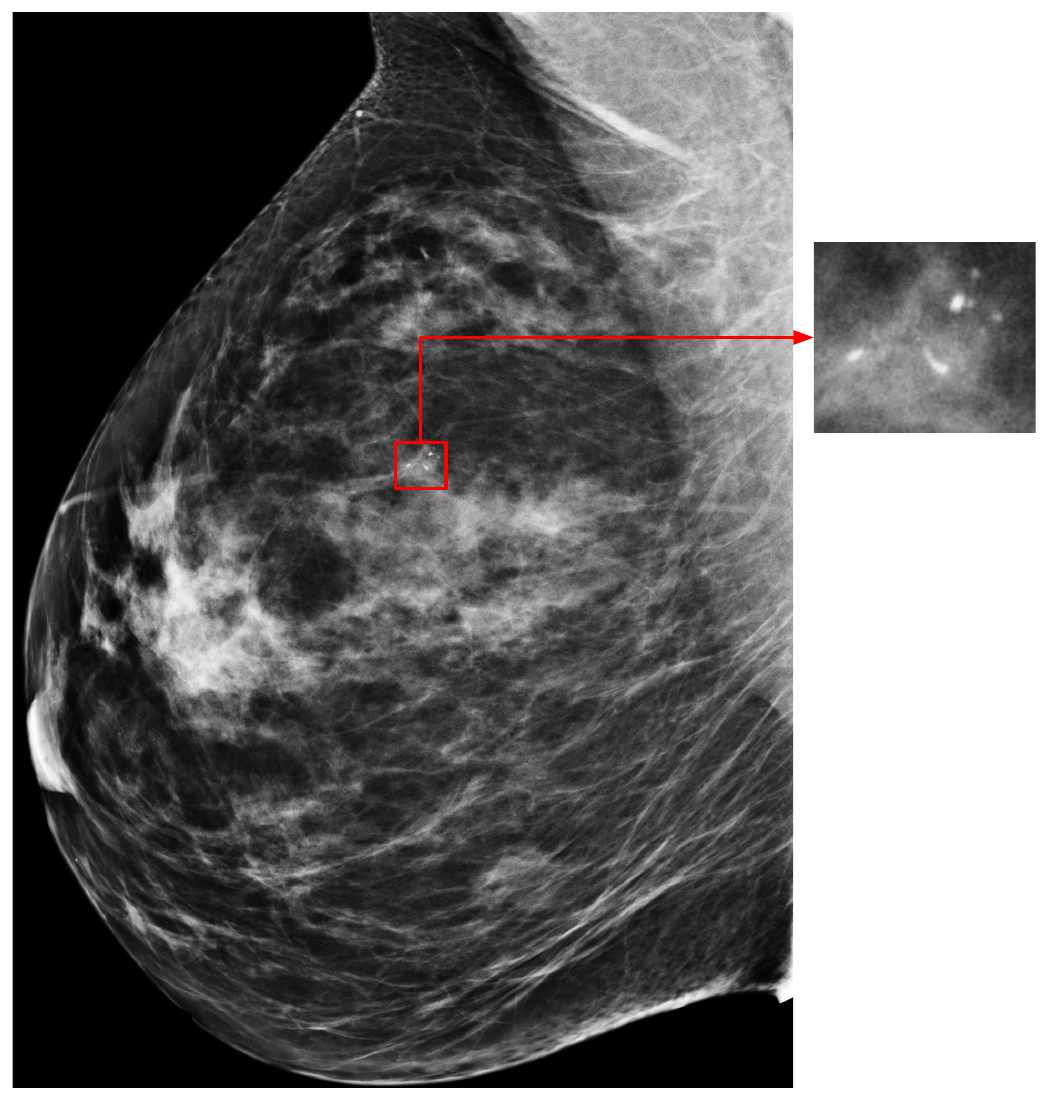}
        \caption{R-MLO}
    \end{subfigure}%
    \begin{subfigure}[b]{0.21\textwidth}
        \includegraphics[scale=0.21]{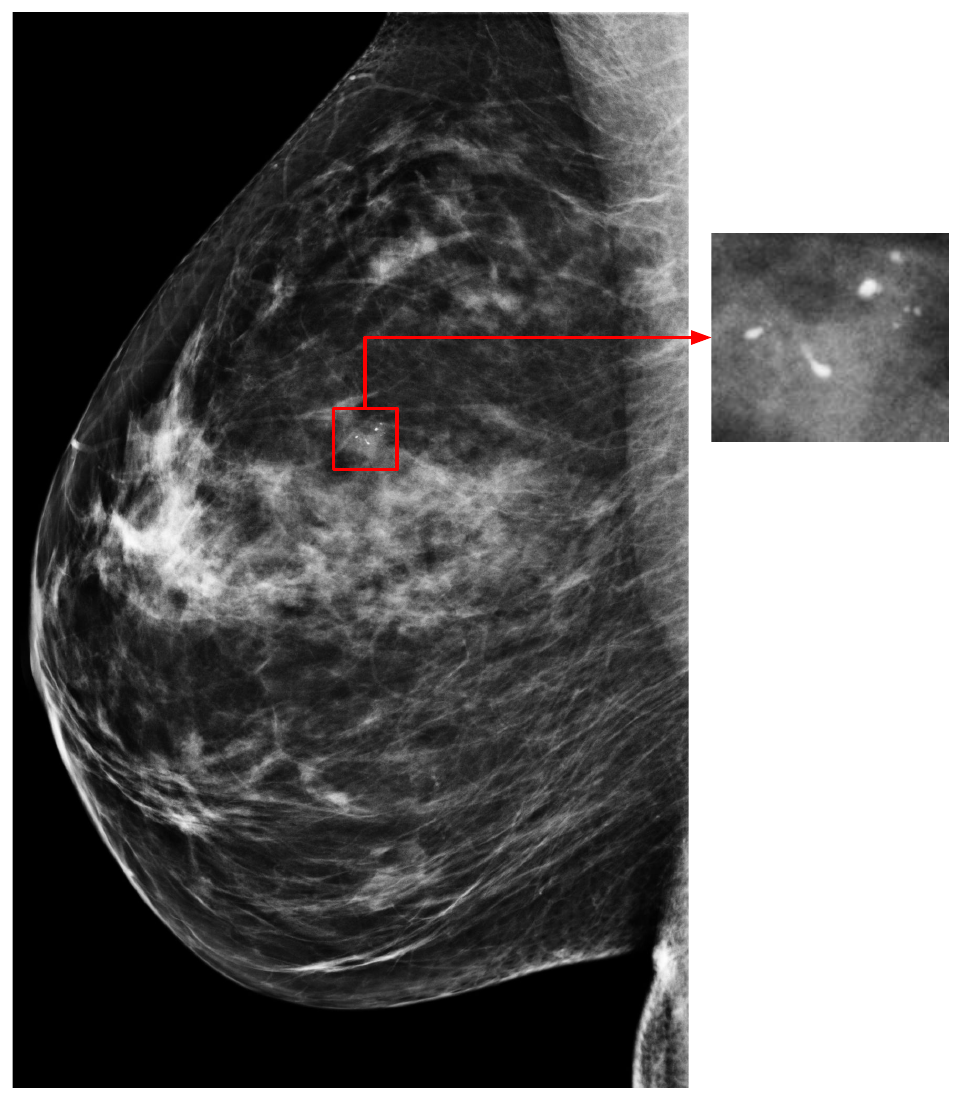}
        \caption{R-LM}
    \end{subfigure}%
    \begin{subfigure}[b]{0.17\textwidth}
        \includegraphics[scale=0.21]{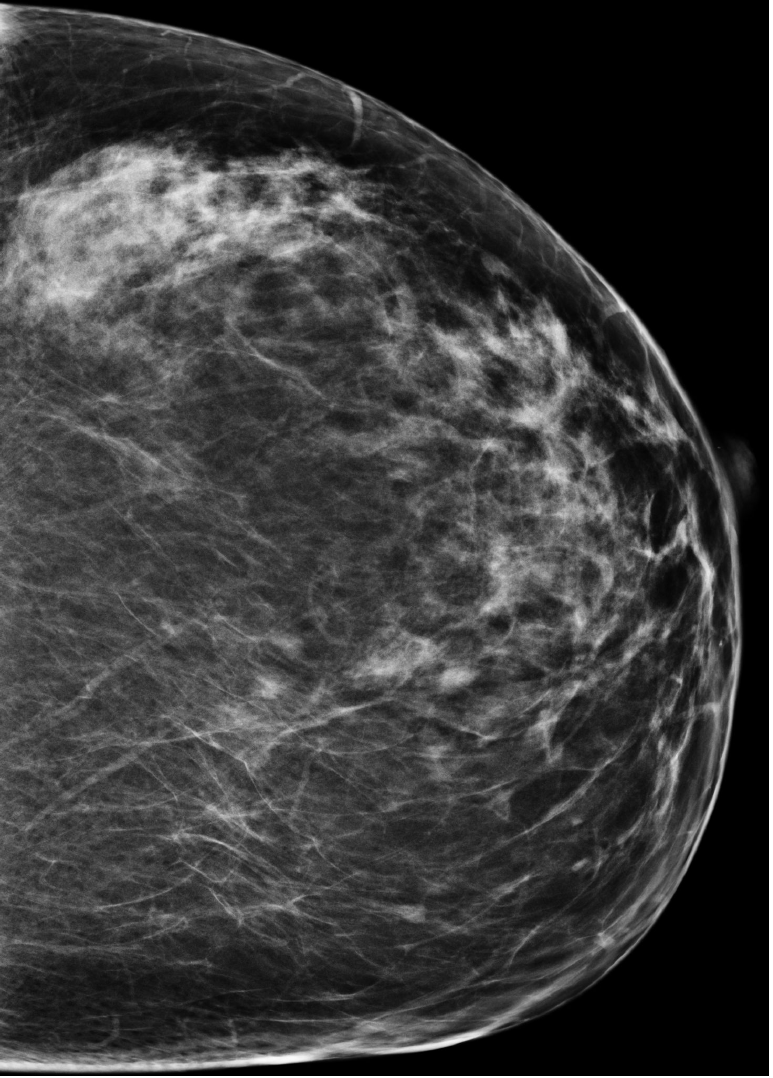}
        \caption{L-CC}
    \end{subfigure}%
    \begin{subfigure}[b]{0.18\textwidth}
        \includegraphics[scale=0.21]{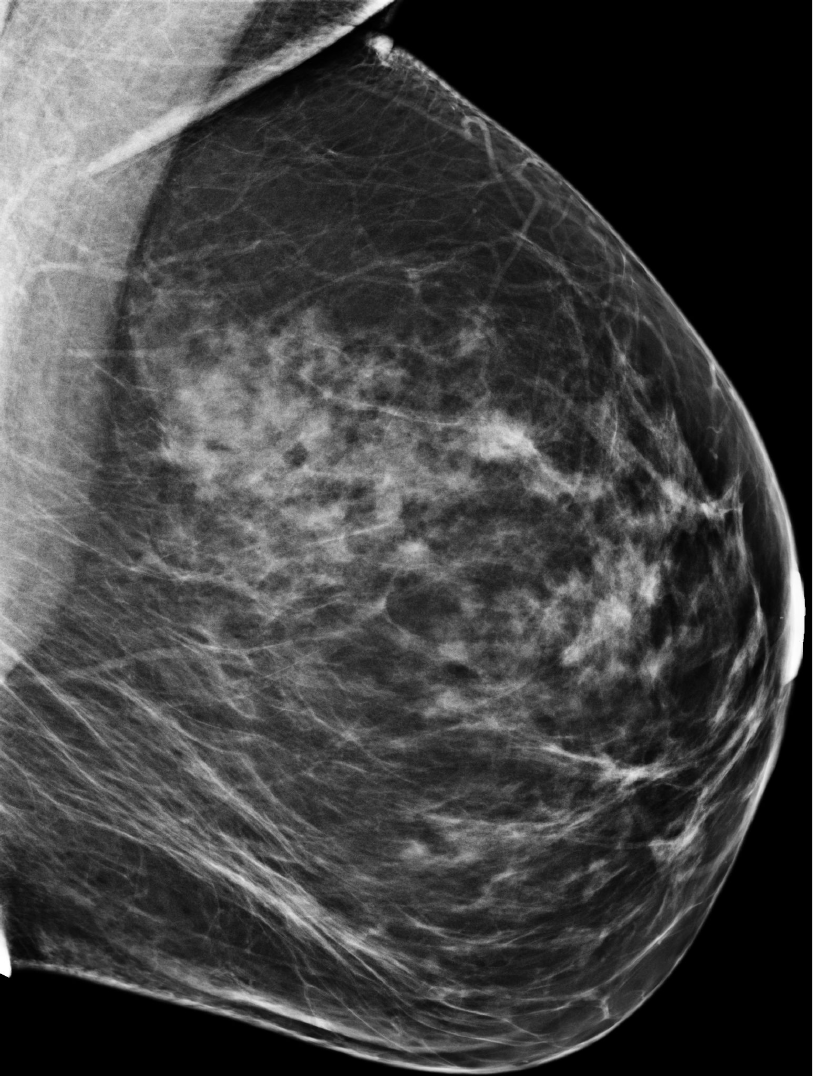}
        \caption{L-MLO}
    \end{subfigure}
\caption{\textit{Why is this case of interest? Variable-image case with $>4$ images and abnormality in one breast}. A benign case from the \mgm dataset showing standard craniocaudal (CC) and mediolateral oblique (MLO) views of the left (L-) and right (R-) breast, along with an additional lateromedial (LM) view of the right breast. A pathologically proven benign calcification of amorphous morphology and grouped distribution is visible in all views of the right breast and no abnormality visible in the left breast. The case is labeled benign in the hospital system due to the presence of benign abnormalities in the right breast.}
\label{sup:fig:case-5image-mgm-example}
\end{figure}


\begin{figure}[H]
\centering
\includegraphics[scale=0.27]{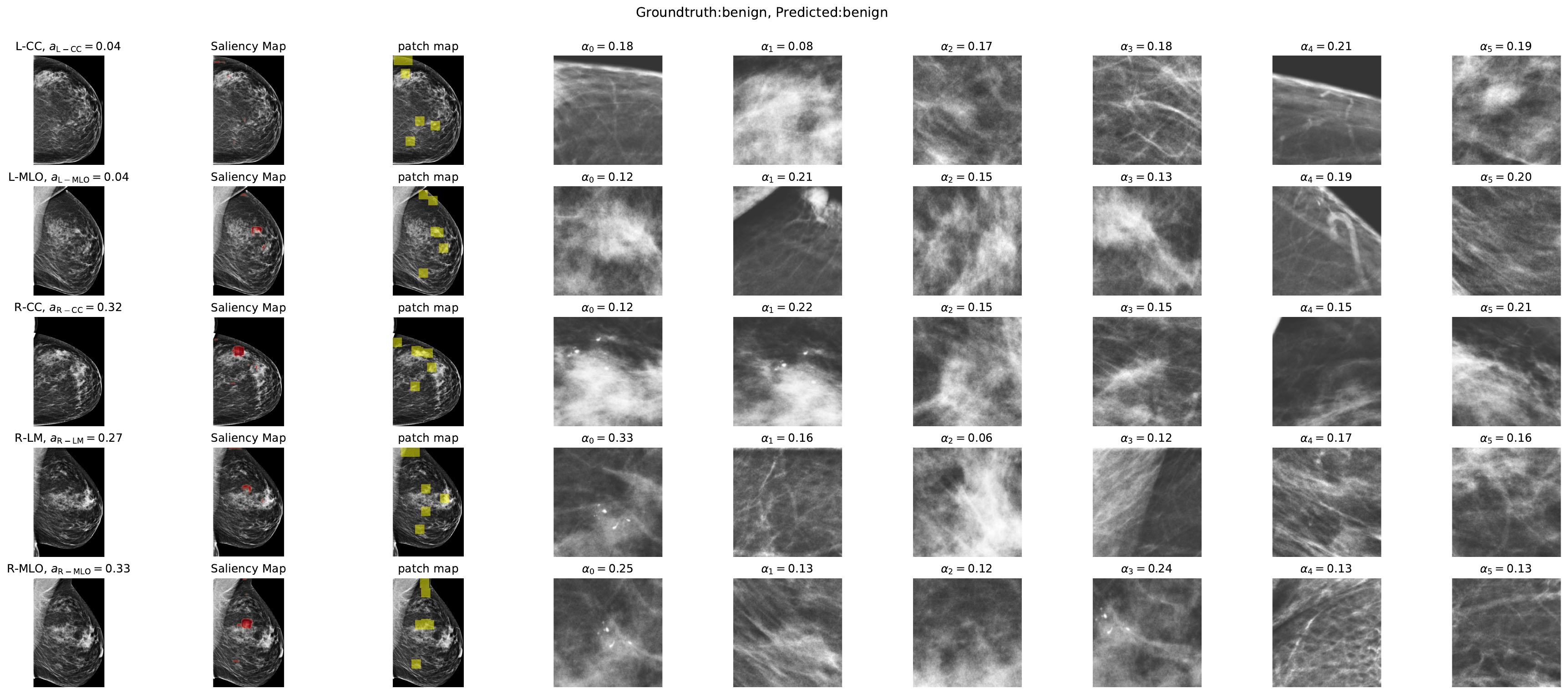}
\caption{\ESAttSide correctly predicted this benign case as benign. Calcifications present in the right breast was extracted by the model from all the views of the right breast. Though the model has not seen LM view during training (the model above is trained on \mgmfv which does not contain the additional views), the model extracted the correct ROI from that view. Our proposed model \ESAttSide has assigned high attention weights to all the views of the right breast and low attention weights to all the views of the left breast.}
\label{sup:fig:case-5image-tn}
\end{figure}

\subsection{Case 3: Malignant abnormality seen in one view of the left breast and benign abnormality in both views of the right breast; False negative prediction (malignant predicted as benign)}

\begin{figure}[H]
\centering
    \begin{subfigure}[b]{0.22\textwidth}
    \centering
        \includegraphics[scale=0.34]{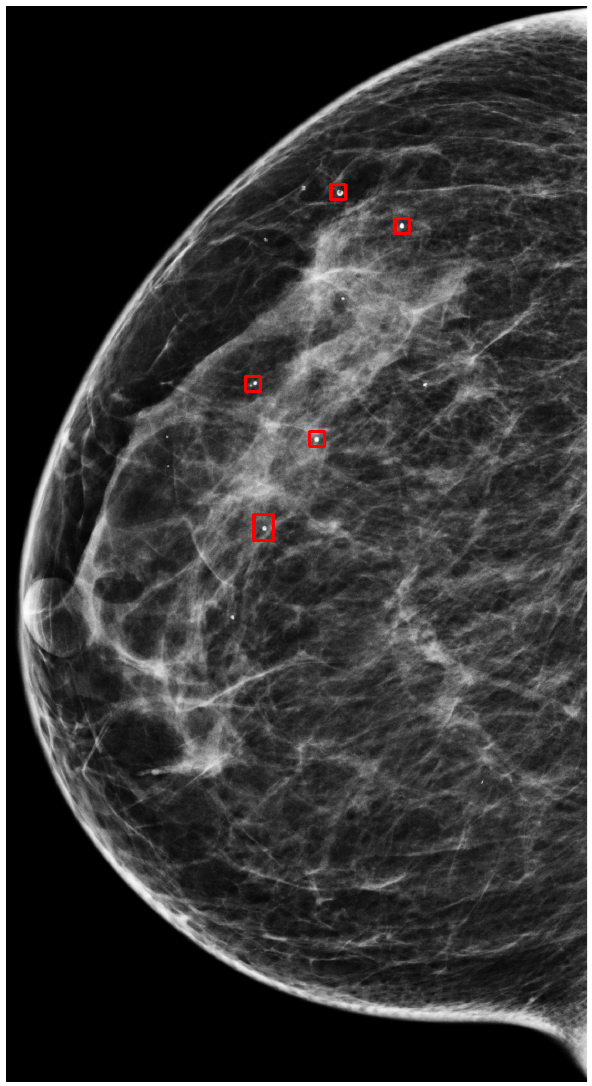}
        \caption{R-CC}
    \end{subfigure}%
    \begin{subfigure}[b]{0.22\textwidth}
    \centering
        \includegraphics[scale=0.336]{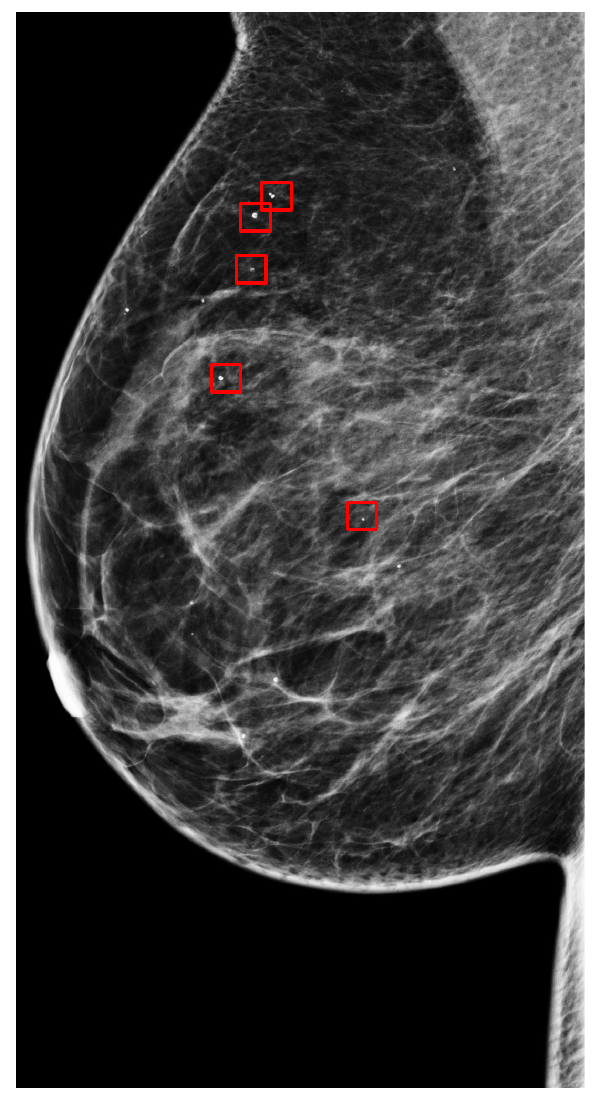}
        \caption{R-MLO}
    \end{subfigure}%
    \begin{subfigure}[b]{0.22\textwidth}
    \centering
        \includegraphics[scale=0.34]{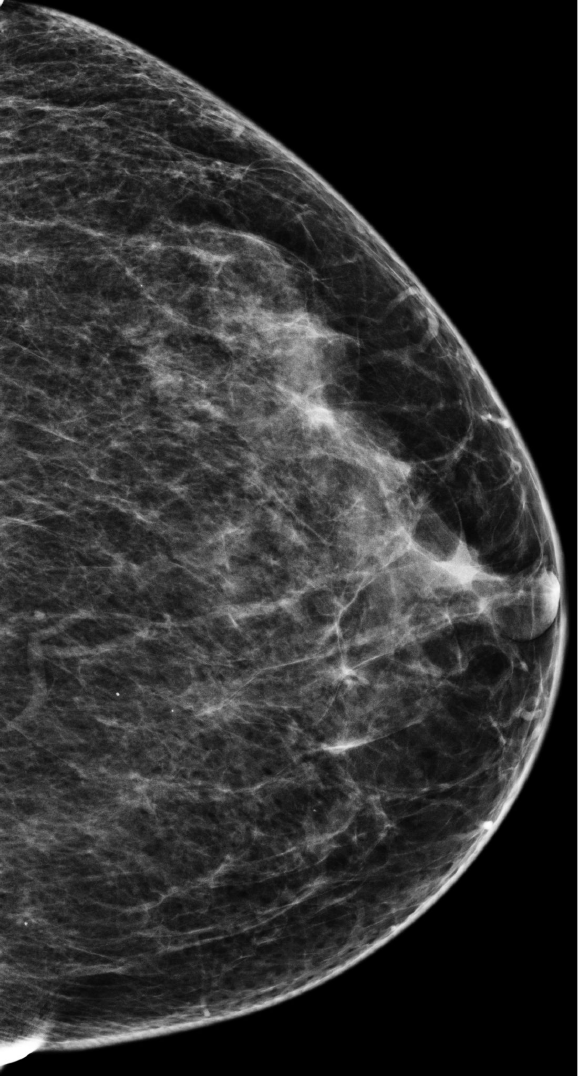}
        \caption{L-CC}
    \end{subfigure}%
    \begin{subfigure}[b]{0.33\textwidth}
    \centering
        \includegraphics[scale=0.341]{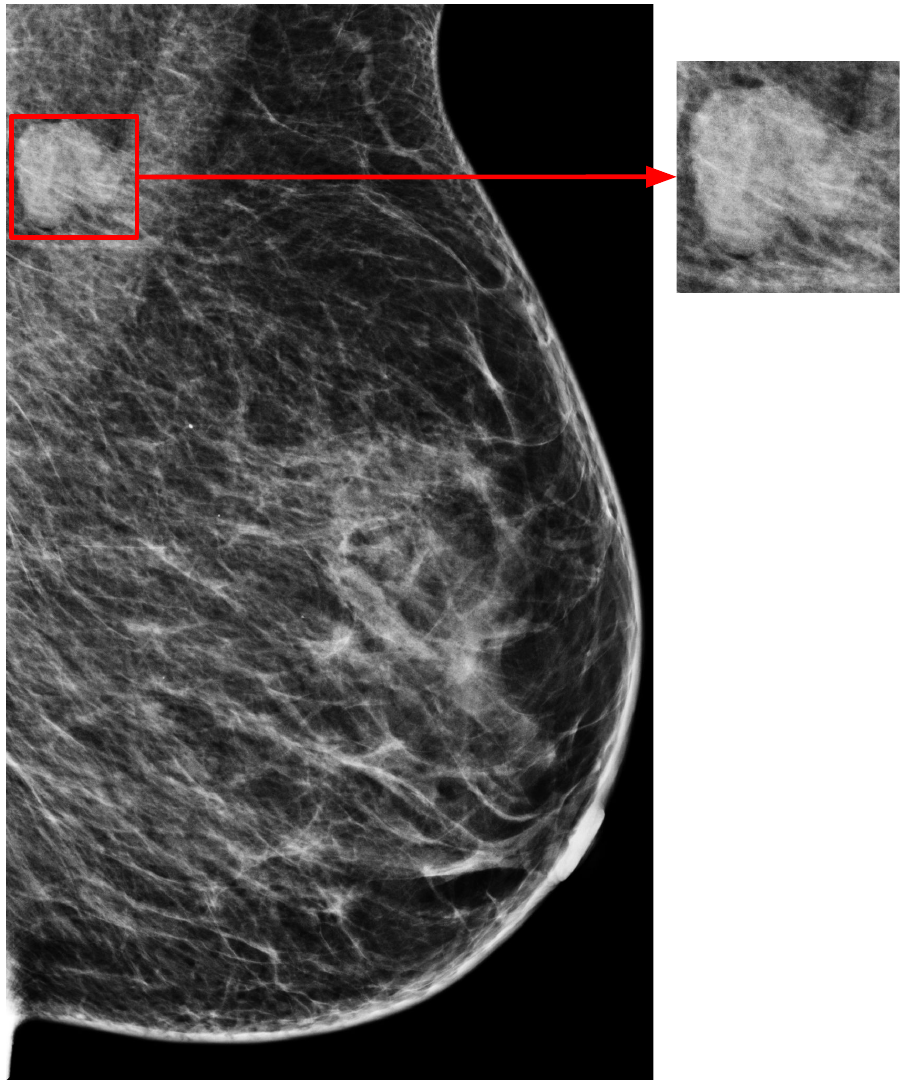}
        \caption{L-MLO}
    \end{subfigure}%
\caption{\textit{Why is this case of interest? Abnormality visible in one view of the left breast and both views of the right breast.} A malignant case from the \mgm dataset showing standard views from left (L-) and right (R-) breast. A pathologically proven malignant mass of oval shape and circumscribed margin can be seen projected over the pectoral muscle in L-MLO view and it is not possible to see this in L-CC view. Benign round/punctate calcification of diffuse distribution can be seen in both views of the left breast.}
\label{sup:fig:intro:case-mgm-roi-1viewR}
\end{figure}


\begin{figure}[H]
\centering
\includegraphics[scale=0.275]{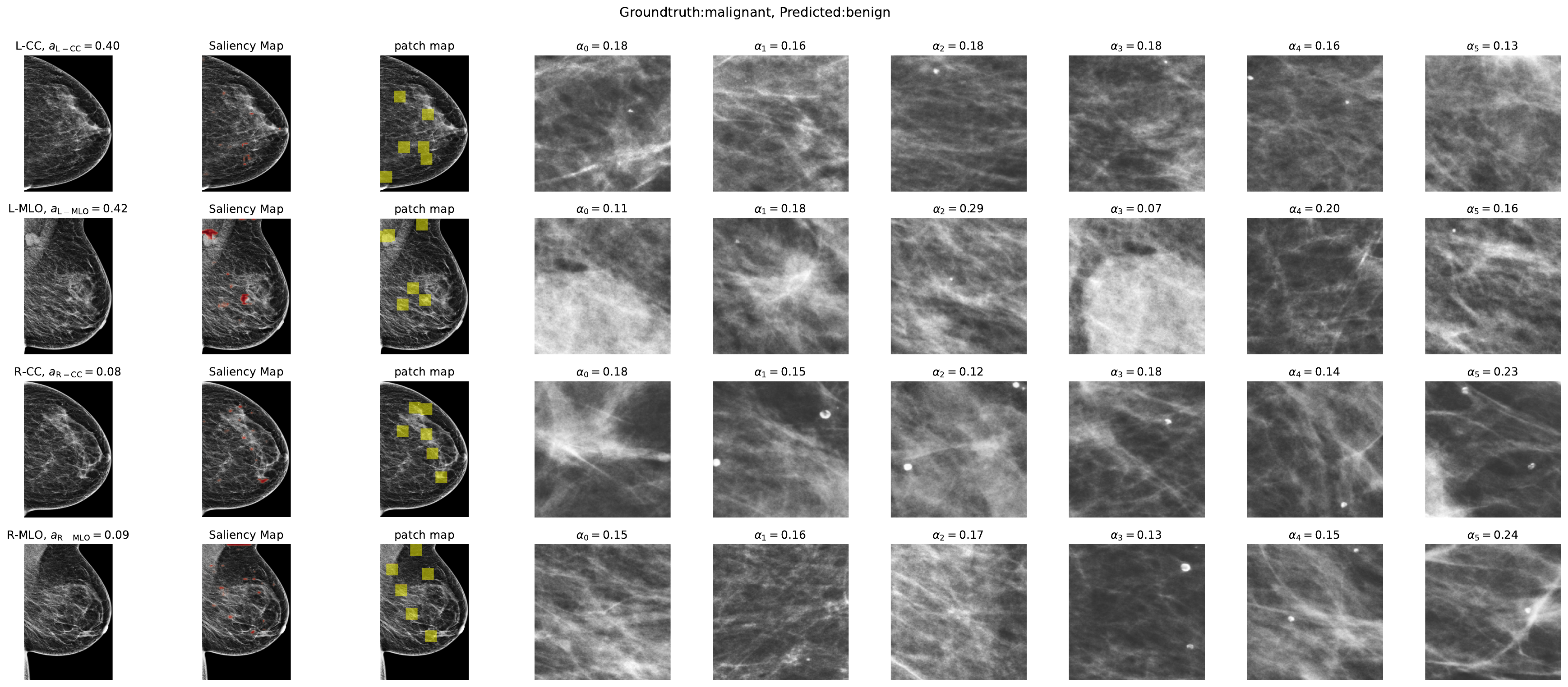}
\caption{This malignant mammogram case is predicted as benign by the model. The model extracted the mass ROI from L-MLO image (1st and 4th patches in \ESAttSide) and calcification ROIs from R-CC and R-MLO images. \ESAttSide assigned the highest image-level attention weight to L-MLO.}
\label{sup:fig:abnormality-oneview-fn}
\end{figure}

\subsection{Case 4: Benign abnormality seen in all views of the right breast and no abnormality seen in the left breast; False positive prediction (benign predicted as malignant)}

\begin{figure}[H]
\centering
    \begin{subfigure}[b]{0.22\textwidth}
    \centering
        \includegraphics[scale=0.26]{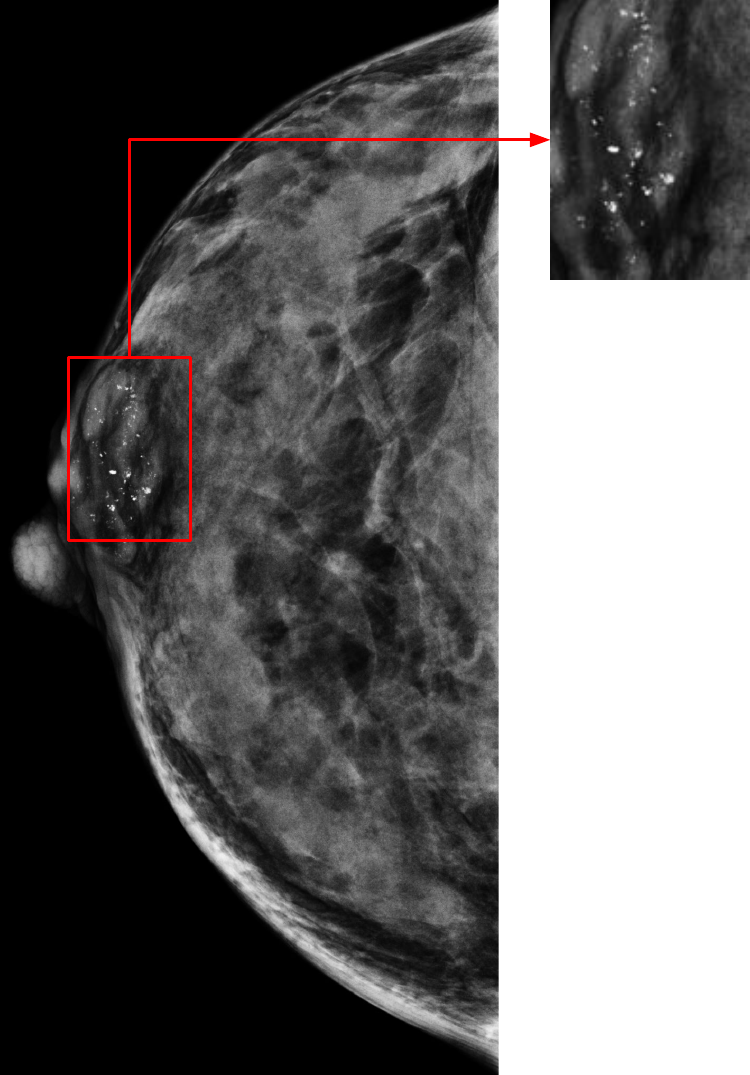}
        \caption{R-CC}
    \end{subfigure}%
    \begin{subfigure}[b]{0.19\textwidth}
    \centering
        \includegraphics[scale=0.26]{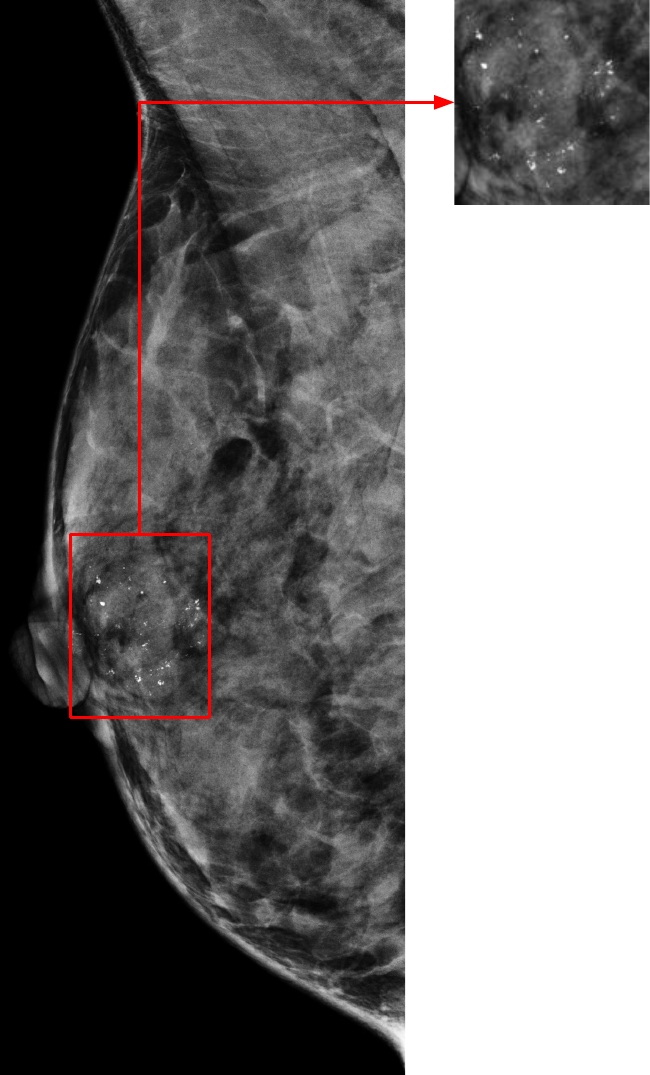}
        \caption{R-MLO}
    \end{subfigure}%
    \begin{subfigure}[b]{0.19\textwidth}
    \centering
        \includegraphics[scale=0.26]{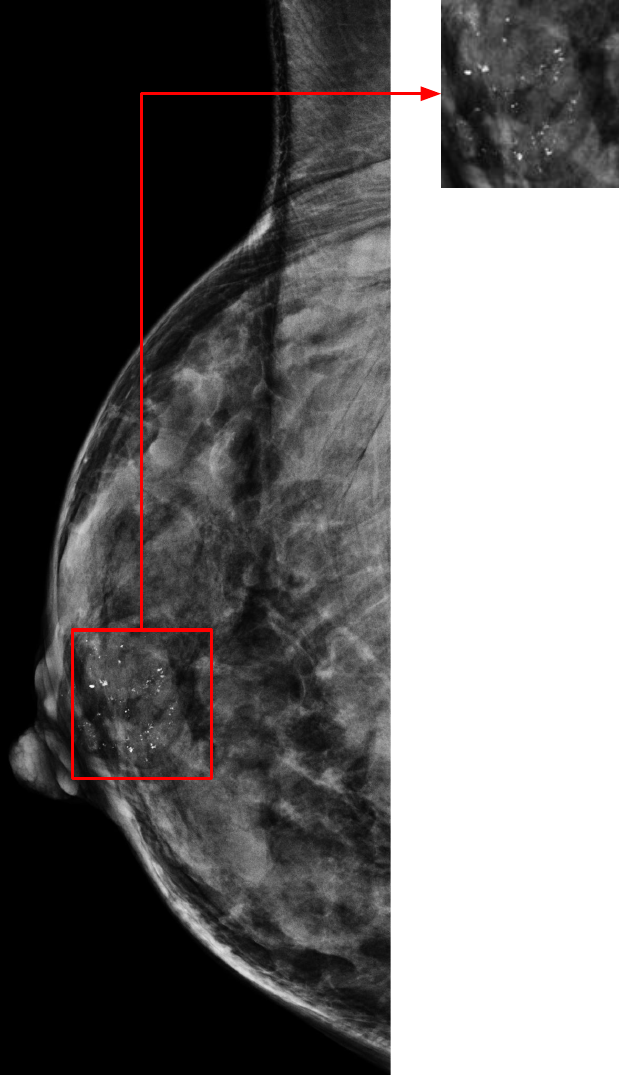}
        \caption{R-XCCL}
    \end{subfigure}%
    \begin{subfigure}[b]{0.13\textwidth}
    \centering
        \includegraphics[scale=0.26]{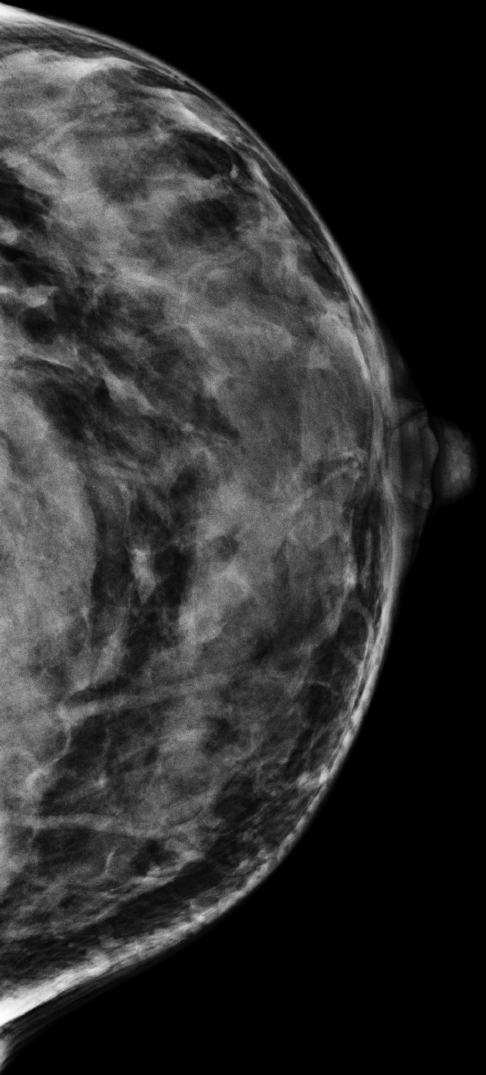}
        \caption{L-CC}
    \end{subfigure}%
    \begin{subfigure}[b]{0.13\textwidth}
    \centering
        \includegraphics[scale=0.26]{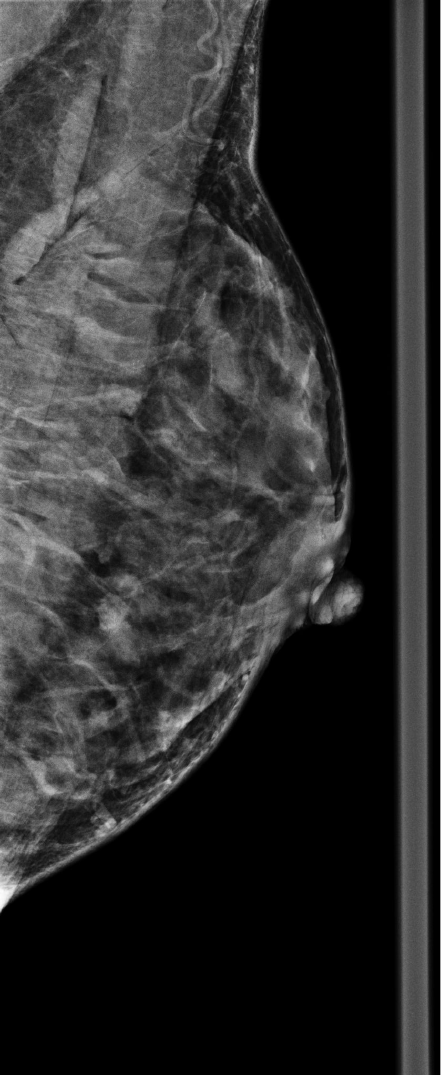}
        \caption{L-MLO}
    \end{subfigure}%
    \begin{subfigure}[b]{0.13\textwidth}
    \centering
        \includegraphics[scale=0.26]{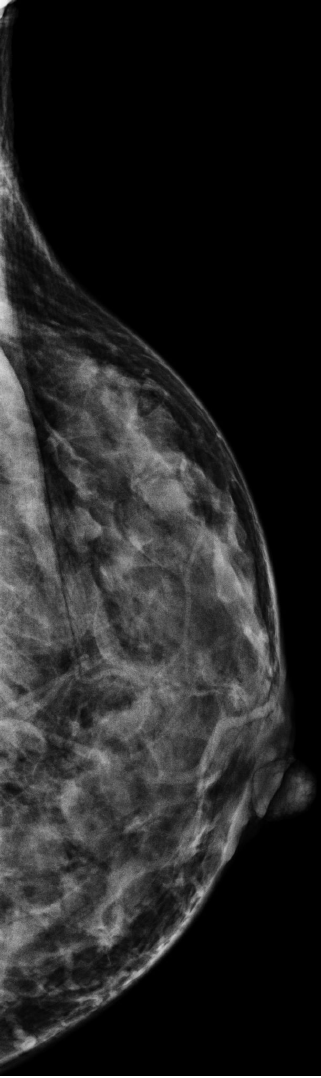}
        \caption{L-XCCL}
    \end{subfigure}   
\caption{\textit{Why is this case of interest? Variable-image case with $>4$ images and abnormality visible in all views of one breast.} A benign case from the \mgm dataset showing standard views, CC and MLO and an additional view, XCCL from left (L-) and right (R-) breast. A benign mass and calcification is visible in all views of the right breast and no abnormality visible in the left breast.}
\label{sup:fig:intro:case-mgm-roi-6images}
\end{figure}


\begin{figure}[H]
\centering
    \begin{subfigure}[b]{\textwidth}
       \includegraphics[scale=0.28]{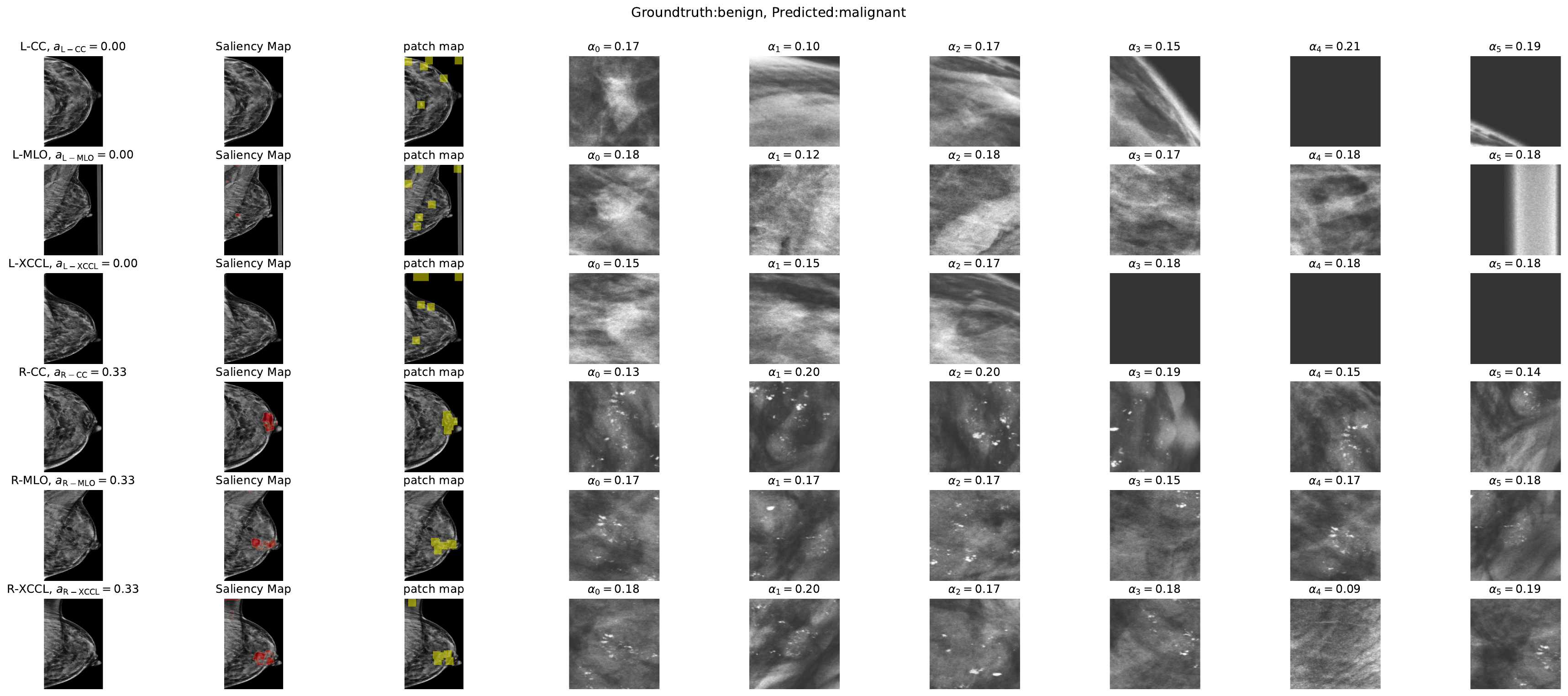}
        \caption{\ESAttSide trained on \mgmfv}
    \end{subfigure}
\caption{\ESAttSide wrongly predicted the case as malignant. The model can extract the abnormality correctly. \ESAttSide assign high image-level attention scores to all views in the right breast.}
\label{sup:fig:abnormality-6imagecase-fp}
\end{figure}


\section{Quality of Unsupervised ROI extraction}

\begin{table}[t!hbp]
\centering
\caption{IoU and DSC score averaged over all ROIs in an image for all SIL and MIL models on \cbiscus, \vindr, \mgmannot; n.a.: image labels are unavailable. \ESAttSide outperforms other MIL models for \cbiscus and \vindr and has similar performance to SIL models.}
\label{sup:tab:results:mean-roi-extraction}
\begin{small}
\begin{tabular}{+lllllll}
\toprule \tabhead
  & \multicolumn{2}{c}{\textbf{\cbiscus}} & \multicolumn{2}{c}{\textbf{VinDr}} & \multicolumn{2}{c}{\textbf{\mgmannot}} \\ 
\textbf{Model}  & \textbf{IoU} & \textbf{DSC} & \textbf{IoU} & \textbf{DSC} &  \textbf{IoU} & \textbf{DSC} \\ \otoprule
\SILil & $\textbf{0.07} \pm 0.03$ & $\textbf{0.10} \pm 0.03$ & $0.27 \pm 0.02$ & $0.38 \pm 0.03$ & n.a. & n.a.\\
\SILcl & $0.06 \pm 0.06$ & $0.09 \pm 0.09$ & $0.22 \pm 0.01$ & $0.32 \pm 0.02$ & $0.16$ & $0.25$\\
\ISMeanImg & $0.04 \pm 0.04$ & $0.07 \pm 0.06$ & $0.20 \pm 0.03$ & $0.30 \pm 0.04$ & $0.13$ & $0.20$\\  
\ISMaxImg & $0.03 \pm 0.04$ & $0.05 \pm 0.06$ & $0.22 \pm 0.03$ & $0.32 \pm 0.03$ & $0.00$ & $0.00$\\ 
\ISAttImg & $0.02 \pm 0.03$ & $0.03 \pm 0.03$ & $0.19 \pm 0.03$ & $0.27 \pm 0.05$ & $0.15$ & $0.22$ \\ 
\ISGattImg & $0.04 \pm 0.05$ & $0.06 \pm 0.07$ & $0.21 \pm 0.07$ & $0.33 \pm 0.08$ & $0.15$ & $0.23$\\ 
\ISAttSide & $0.03 \pm 0.05$ & $0.05 \pm 0.07$ & $0.26 \pm 0.03$ & $0.37 \pm 0.03$ & $\textbf{0.18}$ & $\textbf{0.27}$\\ 
\ESMeanImg & $0.04 \pm 0.04$ & $0.06 \pm 0.06$ & $0.21 \pm 0.02$ & $0.30 \pm 0.03$ & $0.17$ & $0.25$\\ 
\ESMaxImg & $0.00 \pm 0.00$ & $0.01 \pm 0.01$ & $0.26 \pm 0.04$ & $0.36 \pm 0.05$ & $0.17$ & $0.25$\\ 
\ESAttImg & $0.02 \pm 0.02$ & $0.02 \pm 0.03$ & $0.23 \pm 0.05$ & $0.33 \pm 0.07$ & $0.17$ & $0.25$\\ 
\ESGattImg & $0.01 \pm 0.02$ & $0.02 \pm 0.03$ & $0.26 \pm 0.02$ & $0.37 \pm 0.03$ & $0.17$ & $0.26$\\ 
\ESAttSide & $\textbf{0.07} \pm 0.04$ & $\textbf{0.10} \pm 0.06$ & $\textbf{0.28} \pm 0.04$ & $\textbf{0.39} \pm 0.05$ & $0.17$ & $0.25$\\  \bottomrule
\end{tabular}
\end{small}
\end{table}

\begin{table}[th!bp]
\centering
\caption{IoU and DSC score for the ROI candidate with the highest patch-level attention score for all SIL and MIL models on \cbiscus, \vindr, \mgmannot ; n.a.: image labels are unavailable.}
\label{sup:tab:results:maxattscore-roi-extraction}
\begin{small}
\begin{tabular}{+lllllll}
\toprule \tabhead
  & \multicolumn{2}{c}{\textbf{\cbiscus}} & \multicolumn{2}{c}{\textbf{VinDr}} & \multicolumn{2}{c}{\textbf{\mgmannot}} \\ 
\textbf{Model}  & \textbf{IoU} & \textbf{DSC} & \textbf{IoU} & \textbf{DSC} &  \textbf{IoU} & \textbf{DSC} \\ \otoprule
\SILil & $0.01 \pm 0.00$ & $0.01 \pm 0.00$ & $0.19 \pm 0.02$ & $0.27 \pm 0.03$ & n.a. & n.a.\\
\SILcl & $0.01 \pm 0.01$ & $0.01 \pm 0.02$ & $0.17 \pm 0.01$ & $0.27 \pm 0.03$ & $0.08$ & $0.12$\\
\ISMeanImg & $\textbf{0.02} \pm 0.01$ & $\textbf{0.02} \pm 0.02$ & $0.16 \pm 0.01$ & $0.24 \pm 0.02$ & $0.07$ & $0.10$\\  
\ISMaxImg & $0.00 \pm 0.01$ & $0.00 \pm 0.01$ & $0.17 \pm 0.03$ & $0.24 \pm 0.04$ & $0.00$ & $0.00$\\ 
\ISAttImg & $0.00 \pm 0.00$ & $0.00 \pm 0.01$ & $0.17 \pm 0.01$ & $0.23 \pm 0.02$ & $0.07$ & $0.11$ \\ 
\ISGattImg & $0.01 \pm 0.01$ & $0.01 \pm 0.01$ & $0.17 \pm 0.01$ & $0.27 \pm 0.04$ & $\textbf{0.10}$ & $\textbf{0.16}$\\ 
\ISAttSide & $0.00 \pm 0.01$ & $0.01 \pm 0.01$ & $0.18 \pm 0.03$ & $\textbf{0.29} \pm 0.03$ & $0.08$ & $0.12$\\ 
\ESMeanImg & $0.01 \pm 0.01$ & $0.01 \pm 0.02$ & $\textbf{0.20} \pm 0.02$ & $0.23 \pm 0.02$ & $0.08$ & $0.13$\\ 
\ESMaxImg & $0.00 \pm 0.00$ & $0.00 \pm 0.00$ & $0.16 \pm 0.02$ & $0.26 \pm 0.03$ & $0.07$ & $0.10$\\ 
\ESAttImg & $0.00 \pm 0.01$ & $0.00 \pm 0.01$ & $0.18 \pm 0.02$ & $0.26 \pm 0.03$ & $0.07$ & $0.10$\\ 
\ESGattImg & $0.01 \pm 0.01$ & $0.01 \pm 0.02$ & $\textbf{0.20} \pm 0.02$ & $0.28 \pm 0.03$ & $0.09$ & $0.13$\\ 
\ESAttSide & $0.01 \pm 0.01$ & $\textbf{0.02} \pm 0.01$ & $\textbf{0.20} \pm 0.02$ & $0.28 \pm 0.03$ & $0.07$ & $0.11$\\  \bottomrule
\end{tabular}
\end{small}
\end{table}

\newpage

\section{Training with and without Gradient Accumulation}

\begin{table}[th!bp]
\centering
\caption{Comparison of model training with gradient accumulation with accumulation step 2 for batch size=5 vs training without gradient accumulation for batch size=5 for \ESAttSide on \mgmfv.}
\label{sup:tab:results:grad-accum}
\begin{small}
\begin{tabular}{+l^c^c}
\toprule \tabhead
\textbf{Training} & \textbf{F1} & \textbf{AUC} \\ \otoprule 
Gradient accumulation  & $0.58 \pm 0.01$ & $0.83 \pm 0.01$ \\
No accumulation  & $\textbf{0.59} \pm 0.00$ & $\textbf{0.85} \pm 0.01$ \\
\bottomrule
\end{tabular}
\end{small}
\end{table}

\end{document}